# Uncertainty Propagation Networks for Neural Ordinary Differential Equations

Hadi Jahanshahi, Zheng H. Zhu*

Department of Mechanical Engineering, York University, 4700 Keele Street, Toronto, Ontario, M3J 1P3, Canada

## Abstract

This paper introduces Uncertainty Propagation Network (UPN), a novel family of neural differential equations that naturally incorporate uncertainty quantification into continuous-time modeling. Unlike existing neural ordinary differential equations (neural ODEs) that predict only state trajectories, UPN simultaneously models both state evolution and its associated uncertainty by parameterizing coupled differential equations for mean and covariance dynamics. The architecture is grounded in Gaussian moment closure approximation, which enables efficient analytical uncertainty propagation through nonlinear dynamics without requiring stochastic sampling or ensemble methods. UPN supports two operational modes: pure prediction from initial conditions, and adaptive filtering with sparse measurement updates when observations become available during the prediction horizon. The continuous-depth formulation provides principled uncertainty quantification in a single forward pass, handles irregularly-sampled observations naturally, and adapts evaluation strategy to each input's complexity. Experimental results demonstrate UPN's effectiveness across multiple domains: (1) four canonical non-chaotic dynamical systems achieve near-perfect 96.7% confidence interval coverage with single-point Markovian initialization; (2) chaotic Lorenz attractor modeling maintains 94.5% calibration while correctly capturing exponential uncertainty growth in a fully Markovian framework; (3) real-world CubeSat trajectory prediction achieves 89.6% error reduction through integrated measurement updates; and (4) time-series forecasting on the ETTh1 benchmark dataset demonstrates 14% improved accuracy and 6.6× faster inference compared to Neural Stochastic Differential Equations (Neural SDEs). These gains stem from UPN's analytical distribution evolution, which provides superior computational efficiency and calibration compared to sampling-based approaches.

---

* Corresponding Author, Email: gzhu@yorku.ca

dynamical systems; stochastic processes; Gaussian process; Kalman filtering.

## 1. Introduction

Imagine predicting the trajectory of a chaotic system like weather patterns or the spread of an epidemic. While reasonable predictions can be made for the near future, confidence should naturally decrease as predictions extend further ahead. Yet most neural differential equation models provide only point estimates, giving no indication of when their predictions become unreliable. This fundamental limitation becomes critical in safety-sensitive applications where understanding the bounds of predictability is as important as the prediction itself.

Consider the concrete example shown in Figure 1: predicting the future state of a damped oscillator with measurement noise. A standard Neural ODE provides a single trajectory prediction (blue line) but gives no indication of uncertainty. In contrast, our proposed UPN naturally quantifies how uncertainty evolves over time (shaded region), providing both the expected trajectory and confidence bounds that widen appropriately as prediction difficulty increases.

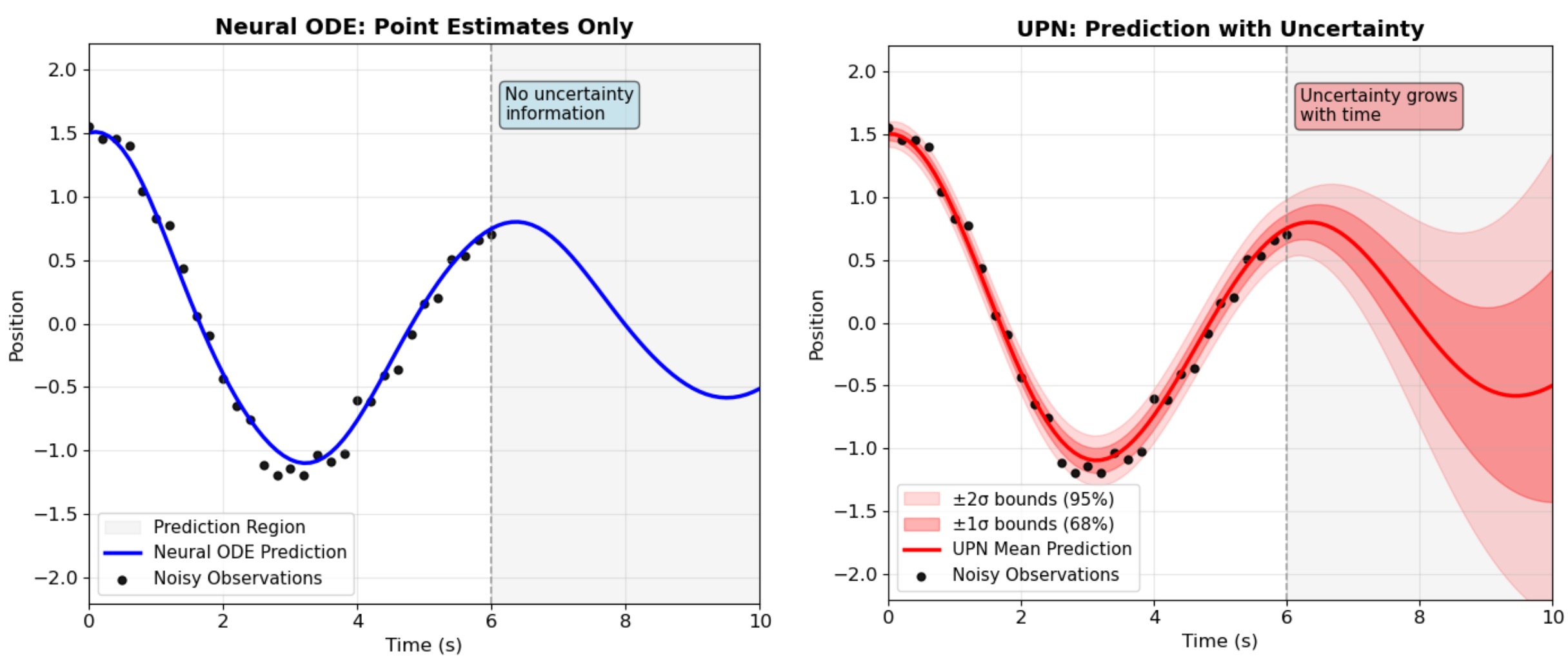

(a) (b)

Figure 1. Sample damped oscillator prediction. (a) Standard Neural ODE provides only point estimates with no uncertainty information. (b) UPN simultaneously predicts mean trajectory (red line) and quantifies uncertainty (shaded bands) that appropriately widens in the prediction region (gray background, t > 6s). Black dots represent noisy training observations.

This example illustrates a fundamental challenge in continuous-time modeling: the need for principled uncertainty quantification. Neural ODEs [1] have emerged as a powerful framework for continuous-time modeling, offering memory efficiency, adaptive computation, and elegant theoretical properties. However, they suffer from a critical limitation: the inability to represent and propagate uncertainty through their dynamics.

Real-world systems are inherently uncertain due to measurement noise in observations, process uncertainty in the underlying dynamics, model inadequacy when the true system is more complex than our representation, and initial condition uncertainty from imperfect state estimation. Traditional approaches to uncertainty quantification in neural networks, including ensemble methods [2,3], Bayesian neural networks [4–7], and Neural SDE [8,9], face fundamental trade-offs between computational efficiency, calibration quality, and theoretical soundness.

This work introduces UPN, a novel family of neural differential equations that naturally incorporate uncertainty quantification into continuous-time modeling. UPN simultaneously models both state evolution and uncertainty by parameterizing coupled differential equations for mean and covariance dynamics, as illustrated in Figure 2. Our approach is grounded in Gaussian moment closure approximation of continuous-time stochastic processes, which enables efficient analytical uncertainty propagation without discretization artifacts or sampling overhead.

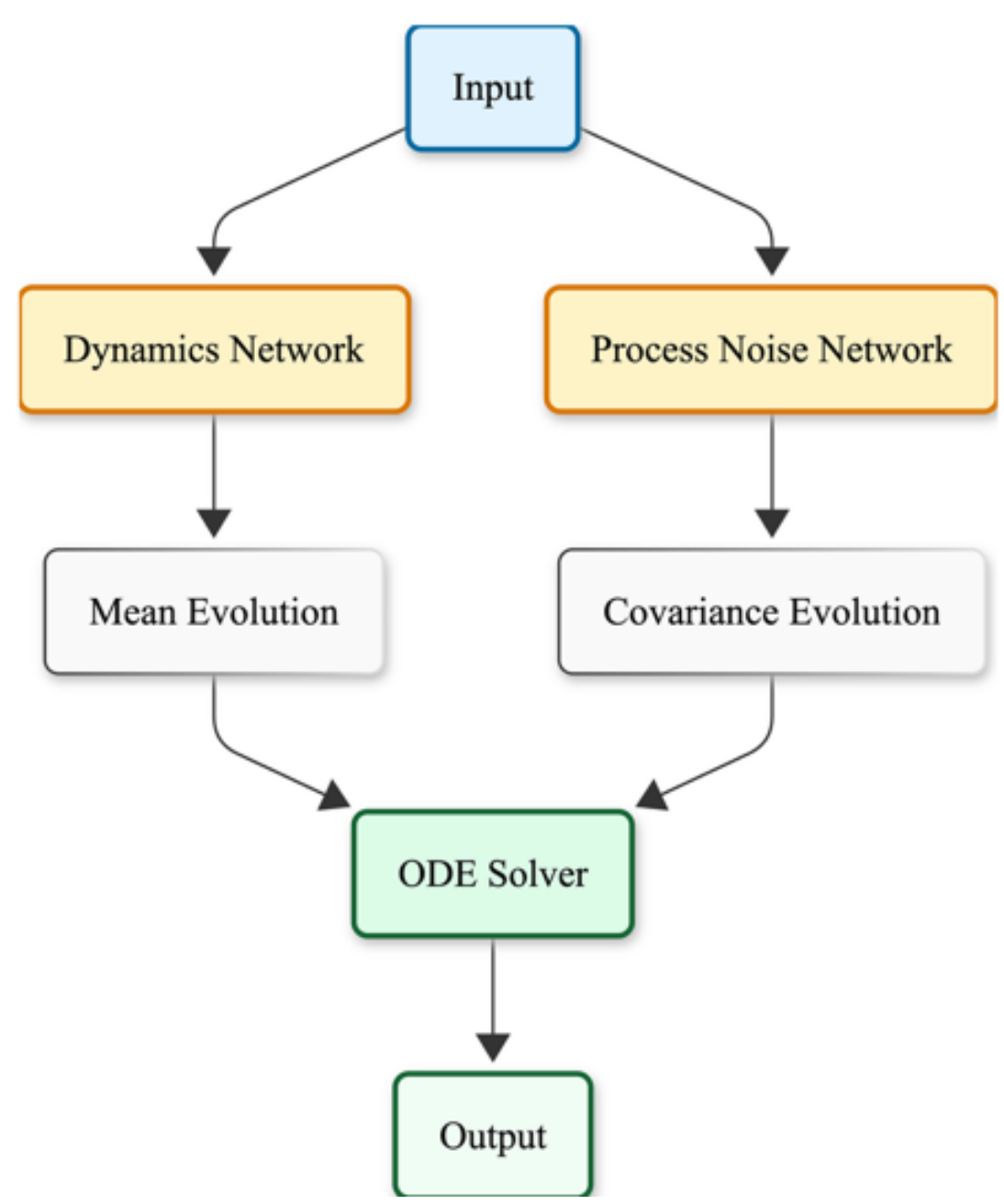

Figure 2. UPN Architecture Overview. Two neural networks parameterize the mean dynamics and process noise, with coupled ODEs solved simultaneously to track both state and uncertainty evolution.

UPN integrates ideas from Neural ODEs [1], continuous-time Gaussian processes [10],

Kalman filtering [11–14], and physics-informed modeling [15–18] into a unified framework. By tracking mean and covariance evolution through coupled ODEs, UPN provides: (1) theoretical soundness via Gaussian process foundations and Fokker-Planck equation derivation, (2) practical efficiency with single-pass uncertainty propagation that is 6–14× faster than sampling-based alternatives, and (3) robustness to irregular sampling, missing data, and chaotic dynamics. This positions UPN as a general-purpose tool for uncertainty-aware modeling in dynamic, data-driven systems where both prediction accuracy and calibration quality are essential.

A distinguishing feature of UPN is its support for two operational scenarios depending on data availability. In pure prediction mode, the model forecasts future states from initial conditions without receiving intermediate observations, appropriate for systems with infrequent measurements. In adaptive filtering mode, UPN incorporates sparse measurements through an integrated Kalman update mechanism, enabling dynamic correction of both state estimates and uncertainty quantification when observations arrive. This dual-mode capability is architecturally inherent to UPN's probabilistic formulation, requiring no model modification between scenarios, whereas deterministic methods would need substantial redesign to incorporate measurements during inference.

Our approach makes several key contributions:

1. Novel Framework: The study introduces the first continuous-time neural architecture that analytically evolves both the mean and covariance of the state distribution through coupled differential equations, grounded in the Gaussian moment closure approximation of stochastic differential equations.
2. Superior Uncertainty Calibration: UPN achieves near-perfect uncertainty quantification across diverse domains, significantly outperforming sampling-based approaches while maintaining competitive or superior point prediction accuracy.
3. Computational Efficiency: By propagating uncertainty analytically rather than through Monte Carlo sampling, UPN achieves 6–14× faster inference than Neural SDEs and ensemble methods while providing better-calibrated confidence intervals.
4. Real-World Validation: Comprehensive experiments demonstrate effectiveness on: (1) four canonical non-chaotic dynamical systems with single-point initialization, (2) the chaotic Lorenz attractor under full Markovian observation, (3) real-world CubeSat trajectory data with integrated measurement updates achieving 89.6% error reduction, and (4) industrial time-series forecasting on the ETTh1 benchmark with 14% improved accuracy over Neural SDEs.

The remainder of this paper is organized as follows: Section 2 positions UPN within the broader landscape of related work, emphasizing the distinction between sampling-based and analytical approaches to uncertainty quantification. Section 3 details the mathematical formulation, including the Gaussian moment closure approximation and optional measurement update mechanism. Section 4 describes the computational implementation, gradient computation via the adjoint method, and complexity analysis demonstrating $O(n^3)$ per-step cost with practical scalability. Section 5 presents comprehensive experimental validation across dynamical systems (non-chaotic and chaotic), real-world CubeSat state estimation, and time-series forecasting on industrial data. Sections 6 and 7 provide discussion and conclusions, addressing the trade-offs between point accuracy and uncertainty calibration, computational considerations, and future research directions.

## 2. Related Work

Neural differential equations can be broadly categorized based on whether they incorporate uncertainty quantification. As shown in Figure 3, this leads to a high-level taxonomy: deterministic models, which ignore uncertainty, and uncertainty-aware models, which employ different computational strategies, including sampling-based methods that require multiple evaluations (ensembles or trajectory sampling), and analytical distribution evolution that directly propagates uncertainty through moment dynamics.

Well-known deterministic approaches include Standard Neural ODEs [1], Augmented Neural ODEs [19], Hamiltonian Neural Networks [17], Neural Controlled Differential Equations (Neural CDE) [20], and Second-order Neural ODEs [21]. While these frameworks provide powerful tools for continuous-time modeling, they fundamentally lack mechanisms for uncertainty quantification. Our proposed UPN addresses this limitation by providing an uncertainty-aware alternative that explicitly models uncertainty dynamics through coupled differential equations, positioning it within the distribution evolution branch of uncertainty-aware approaches.

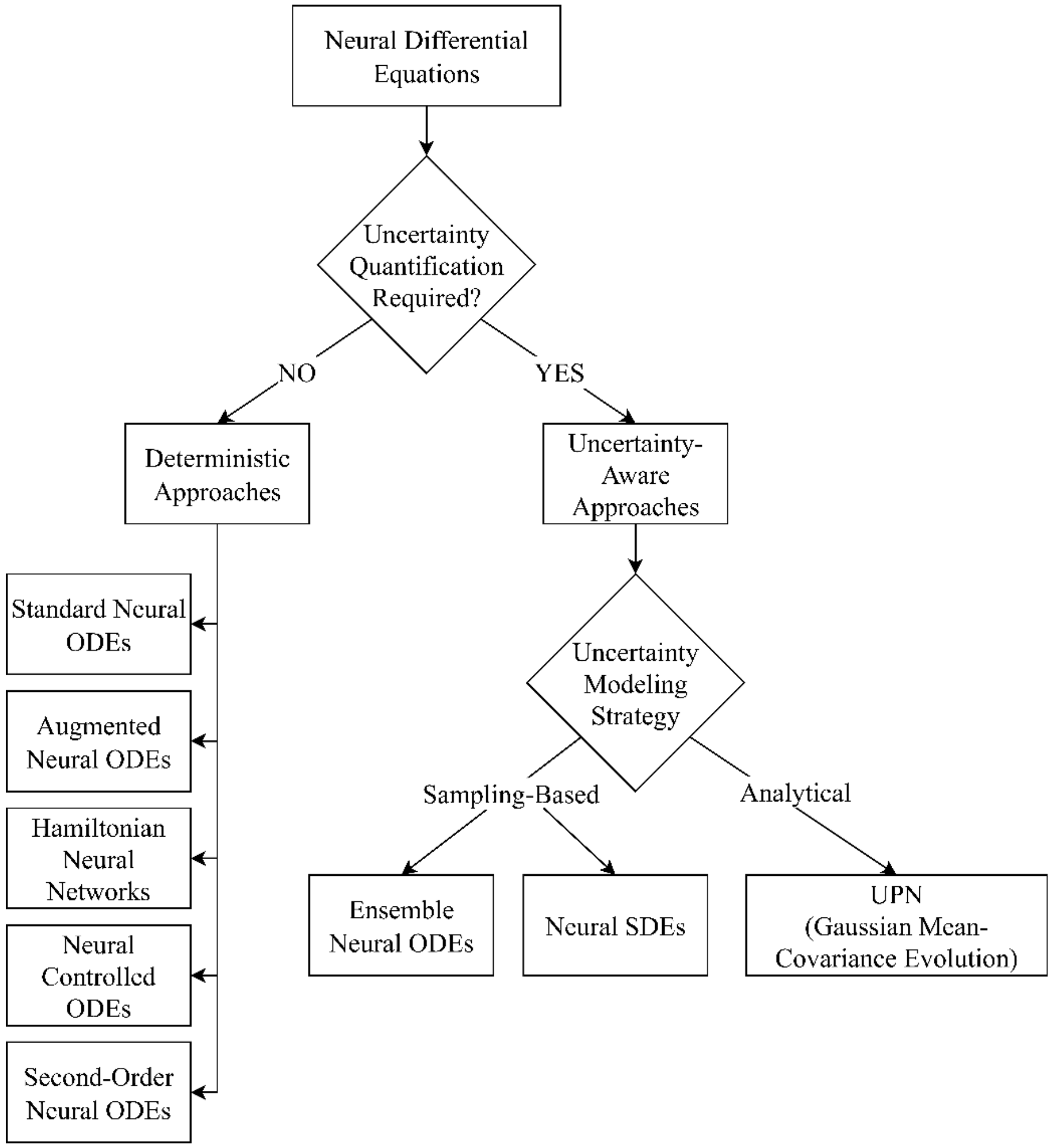

Figure 3. Taxonomy of neural differential equation approaches. UPN provides a computationally efficient alternative to sampling-based methods by analytically evolving the mean and covariance of Gaussian state distributions through coupled differential equations.

### *2.1. Neural Differential Equations*

Neural ODEs [1] originated from the insight that residual networks approximate continuous-time systems [22–24]. These models define state evolution via neural networks [25] and have seen success in time-series modeling and normalizing flows [26,27]. The key innovation was recognizing that the adjoint sensitivity method enables memory-efficient backpropagation through ODE solvers, making continuous-depth networks practical.

Extended variants have been developed to address specific limitations: Augmented Neural

ODEs [19] increase expressivity by expanding the state space, Second-order ODEs [21] capture acceleration-based dynamics for physical systems, and Neural CDE [20] handle irregular time series through controlled dynamics. However, all these variants remain fundamentally deterministic, lacking mechanisms for uncertainty quantification beyond post-hoc confidence estimation.

### *2.2. Neural Stochastic Differential Equations*

The closest related work to our approach comes from Neural SDEs, which extend neural ODEs to incorporate stochastic dynamics [8,9]. Neural SDEs are formulated as [9]:

$$d\mathrm{h}(t) = f_\theta(\mathrm{h}(t), t)dt + g_\phi(\mathrm{h}(t), t)d\mathrm{W}(t) \quad (1)$$

where $f_\theta$ is the drift function, $g_\phi$ is the diffusion function, and $\mathrm{W}(t)$ represents Brownian motion.

While Neural SDEs can model stochastic behavior and have demonstrated success in applications requiring trajectory-level stochasticity, they differ fundamentally from our approach in several key aspects:

- Trajectory-based vs. Distribution-based: Neural SDEs model individual stochastic trajectories, requiring Monte Carlo sampling for uncertainty quantification. UPN directly models the evolution of probability distributions through their sufficient statistics (mean and covariance), providing uncertainty estimates in a single forward pass.
- Computational Efficiency: Neural SDEs require multiple trajectory samples to estimate predictive distributions, with inference time scaling linearly with sample count. Our experiments (Section 5.4) demonstrate that achieving reasonable calibration with Neural SDEs requires 100 samples, resulting in 6.6× higher inference cost compared to UPN while still achieving 14% worse prediction accuracy.
- Theoretical Foundation: Neural SDEs rely on stochastic calculus and Itô integration, modeling trajectories of the underlying SDE. UPN is grounded in the deterministic evolution of Gaussian distribution parameters via moment closure approximation, enabling more stable and interpretable uncertainty propagation through coupled ODEs that directly evolve $\mu(t)$ and $\Sigma(t)$.
- Numerical Stability: Stochastic integration requires specialized solvers and can introduce numerical challenges [28,29], particularly for stiff systems. UPN uses standard ODE solvers (e.g., Dormand-Prince) for the coupled mean-covariance system, benefiting from decades of numerical analysis research on deterministic ODEs.

### *2.3. Uncertainty Quantification in Neural Networks*

Traditional approaches to uncertainty quantification span three main paradigms. Bayesian neural networks [4–7] place distributions over network weights, enabling principled uncertainty estimates through either sampling (MCMC) or variational approximations. However, these methods face computational challenges and often struggle with continuous-time dynamics. Ensemble models [2,3] aggregate predictions from multiple independently trained networks, representing current best practice for uncertainty estimation in deep learning. Yet our experiments reveal that ensembles systematically underestimate uncertainty in dynamical systems: for non-chaotic systems, 10-model ensembles achieve only 26% confidence interval coverage compared to UPN’s 96.7%, while even 30-model ensembles on the chaotic Lorenz attractor reach only 66.8% coverage versus UPN's 94.5%. This failure occurs because ensemble variance captures only epistemic uncertainty (model disagreement due to initialization) but misses aleatoric uncertainty (observation noise and process stochasticity) that UPN explicitly models through learned process noise covariance $Q_{\phi}(\mu, t)$. Deep Gaussian Processes [30,31] and Neural Processes [32] provide probabilistic approximators with theoretical guarantees but struggle with scalability and continuous-time dynamics.

### *2.4. Physics-Informed Neural Networks*

Physics-Informed Neural Networks (PINNs) [15,33,34] enforce physical laws through loss function constraints, while Hamiltonian Neural Networks [16,17] encode energy-conserving structures for mechanical systems. These approaches improve physical plausibility and extrapolation but typically lack native uncertainty modeling. Recent extensions have explored uncertainty quantification in PINNs through ensemble methods or Bayesian frameworks, but face similar computational trade-offs as general ensemble approaches.

UPN draws inspiration from physics-informed modeling by incorporating learnable, state-dependent process noise $Q_{\phi}(\mu, t)$ grounded in Gaussian process theory, enabling the model to adapt uncertainty to local dynamical properties without requiring explicit physical laws.

### *2.5. Classical Kalman Filtering and Extensions*

Kalman filters [11–14] and their nonlinear extensions (Extended Kalman Filter, Unscented

Kalman Filter) provide optimal uncertainty-aware state estimation for linear-Gaussian systems. The classical Kalman framework elegantly combines prediction and measurement update steps, propagating both state estimates and uncertainty through time. However, traditional Kalman filtering approaches face fundamental limitations: they operate in discrete time, require known or partially-known system models, and struggle with highly nonlinear dynamics where linearization approximations break down.

Recent work on neural network-aided Kalman filters [35,36] and meta-learned Kalman networks [37] has shown promise in learning components of the filtering process. These hybrid approaches typically use neural networks to learn measurement or process models while maintaining the Kalman update structure. However, they generally operate in discrete time and require architecture-specific modifications for different applications.

UPN generalizes Kalman filtering to continuous-time, end-to-end learnable systems that simultaneously evolve state and uncertainty through coupled ODEs. The measurement update mechanism is architecturally integrated: UPN operates on distributional parameters $(\mu, \Sigma)$ that can be naturally updated via Kalman equations when observations arrive, without requiring model redesign. This enables seamless switching between pure prediction and adaptive filtering modes, effectively bridging model-based filtering and neural dynamics learning.

### *2.6. Neural State Space Models and Koopman Approaches*

Recent advances have combined Koopman operator theory with conditional Gaussian structures for simultaneous forecasting and data assimilation. The Conditional Gaussian Koopman Network (CGKN) [38,39] represents an important parallel development in uncertainty-aware neural dynamics modeling. CGKN transforms general nonlinear systems into neural differential equations with conditional Gaussian structures, where dynamics become conditionally linear in lifted coordinates. By discovering appropriate Koopman embeddings, CGKN enables systems to be represented such that the dynamics of latent states are linear conditioned on observed states, enabling analytical Kalman filtering without ensemble sampling.

The discrete-time extension [39] focuses on partially observed systems governed by PDEs, achieving competitive forecasting performance on Burgers', Kuramoto-Sivashinsky, and Navier-Stokes equations while providing efficient data assimilation through analytical posterior computation. This work demonstrates that Gaussian structures provide computationally efficient alternatives to sampling-based uncertainty quantification, a principle shared with our approach.

While CGKN and UPN both exploit Gaussian structures for analytical uncertainty propagation, they differ in key aspects. CGKN achieves conditional linearity through Koopman lifting and emphasizes continuous data assimilation with frequent observations, whereas UPN maintains fully nonlinear mean-covariance dynamics without dimension augmentation and supports both pure prediction and sparse measurement updates. CGKN's conditional linearity guarantees enable exact Kalman updates in the lifted space, while UPN uses Extended Kalman Filter linearization around the current mean. CGKN is particularly well-suited for PDE systems with rich spatial structure, while UPN targets general dynamical systems with arbitrary nonlinearity.

Both approaches validate that Gaussian frameworks can achieve superior computational efficiency and calibration compared to sampling-based methods, though through different architectural choices suited to different application domains.

### *2.7. Comparison with Related Uncertainty Quantification Approaches*

Table 1 provides a systematic comparison of UPN against major uncertainty quantification approaches across key dimensions. The comparison reveals UPN's unique position: it combines the computational efficiency of analytical methods with the expressivity of learned dynamics, while supporting both pure prediction and measurement integration within a unified framework.

Table 1. Comparison of uncertainty quantification approaches in continuous-time modeling.

| Method | Uncertainty Type | Computational Cost | Calibration Quality | Measurement Integration | Continuous-Time Native |
|---|---|---|---|---|---|
| Ensemble Neural ODE | Epistemic only | $\mathcal{O}(K_{\text{models}})$ | Poor | Requires redesign | ✓ |
| Neural SDE | Aleatoric + Epistemic | $\mathcal{O}(N_{\text{samples}})$ | Moderate | Particle filtering | ✓ |
| Bayesian Neural Nets | Epistemic | High (MCMC/VI) | Variable | Sequential update | ✗ |
| Deep Gaussian Process | Aleatoric + Epistemic | $\mathcal{O}(n^3)$ or approx. | Good | GP conditioning | ✗ |
| UPN | Aleatoric + Epistemic | $\mathcal{O}(1)$ forward pass and $\mathcal{O}(n^3)$ per-step | Excellent | Native Kalman | ✓ |

The key advantages of UPN emerge from its analytical distribution evolution strategy: (1) it captures both aleatoric uncertainty (via learned process noise $Q_\phi$) and epistemic uncertainty (via model capacity) in a single framework, (2) it achieves superior calibration (81-97% coverage) across diverse dynamical regimes, (3) it provides uncertainty estimates in a single forward pass with $O(n^3)$ per-step complexity rather than requiring multiple samples or model evaluations, and (4) it naturally integrates measurements through Kalman updates without architectural modification.

## 3. Uncertainty Propagation Network

### *3.1. Formulation*

A continuous-time dynamical system with uncertainty is defined by modeling the evolution of both the state mean $\mu(\mathrm{t}) \in \mathbb{R}^{\mathrm{n}}$ and covariance $\Sigma(\mathrm{t}) \in \mathbb{R}^{\mathrm{nxn}}$ using coupled differential equations:

$$\frac{d\mu(t)}{dt} = f_\theta(\mu(t), t) \tag{2}$$

$$\frac{d\Sigma(t)}{dt} = J_f(\mu(t), t)\Sigma(t) + \Sigma(t)J_f(\mu(t), t)^T + Q_\phi(\mu(t), t) \tag{3}$$

Here, $J_f(\mu(t), t) = \frac{\partial f_\theta}{\partial \mu}\Big|_{\mu(t)}$ denotes the Jacobian of the dynamics function, and $Q_\phi(\mu(t), t)$ is a positive semi-definite, state-dependent process noise covariance. These equations provide an approximation to the moment dynamics of the stochastic differential equation:

$$dX(t) = f_\theta(X(t), t)dt + g_\phi(X(t), t)dW(t) \tag{4}$$

where $\mathrm{W(t)}$ is a standard Wiener process and $g_\phi g_\phi^T = Q_\phi$.

For general nonlinear dynamics, the exact evolution of the probability distribution $p(X, t)$ is governed by the Fokker-Planck equation. However, directly solving this partial differential equation is computationally intractable in high dimensions. Our approach employs a Gaussian moment closure approximation: we assume the state distribution remains approximately Gaussian and evolve its first two moments using Equations (2) and (3). This approximation becomes exact when $f_\theta$ is linear in $X$, and provides an accurate local linearization for mildly nonlinear systems. The Jacobian-based covariance evolution in Equation (3) captures how uncertainty propagates through the linearized dynamics around the mean trajectory.

Unlike trajectory-based stochastic modeling that requires sampling, this approach evolves the distribution's sufficient statistics directly under the Gaussian assumption, enabling efficient and differentiable uncertainty propagation through a single forward pass.

### *3.2. Neural Network Parameterization*

The dynamics function $f$ and process noise covariance $Q$ in Equations (2) and (3) are parameterized using neural networks that directly define the right-hand sides of the coupled differential equations. These networks constitute the core learnable components of UPN, enabling data-driven discovery of both state evolution dynamics and uncertainty propagation mechanisms.

A neural network $f_\theta: \mathbb{R}^n \times \mathbb{R} \to \mathbb{R}^n$ parameterizes the mean evolution function in Equation (2). This network maps the current state mean $\mu(t)$ and time $t$ to the time derivative of the mean, $\frac{d\mu(t)}{dt} = f_\theta(\mu(t), t)$, thereby governing the expected trajectory through the state space. The network learns to approximate the underlying deterministic dynamics from observed data, capturing how the system's mean state evolves over time.

A separate neural network $Q_\phi: \mathbb{R}^n \times \mathbb{R} \to \mathbb{R}^{n\times n}$ parameterizes the state-dependent process noise covariance appearing in Equation (3). Rather than directly outputting the covariance matrix, this network produces a lower-triangular matrix $L_\phi(\mu, t) \in \mathbb{R}^{n\times n}$, which is used to construct the positive semi-definite process noise covariance:

$$Q_\phi(\mu, t) = L_\phi(\mu, t) L_\phi(\mu, t)^T + \epsilon I \tag{5}$$

where $\epsilon > 0$ is a small constant added for numerical stability. This Cholesky-like parameterization guarantees positive semi-definiteness, ensuring that the learned noise covariance remains physically meaningful throughout training. The network learns to adapt uncertainty growth rates based on the current state, capturing how predictability varies across different regions of the state space.

Both $f_\theta$ and $Q_\phi$ receive the same inputs, the current state mean $\mu(t)$ and time $t$, but perform distinct roles: $f_\theta$ governs the expected trajectory through deterministic dynamics, while $Q_\phi$ models the state-dependent uncertainty growth rate. These functions can be implemented using various differentiable neural network architectures depending on the application domain:

- Multi-layer Perceptrons (MLPs): Suitable for general-purpose modeling with fully-connected layers. In our experiments (Sections 5.1-5.4), we primarily use 2-layer MLPs with 32-64 hidden units and tanh activations.

- Convolutional Neural Networks (CNNs): Applicable when the state space exhibits spatial structure (e.g., image-based observations, spatially-distributed systems), enabling parameter sharing across spatial dimensions.
- Recurrent Neural Networks (RNNs): Useful when learning from observation histories, as in the time-series experiments where an encoder RNN maps historical observations to initial distributional parameters.

The key requirement is that both networks must be fully differentiable to enable gradient-based training via the adjoint method.

The outputs of both networks define the right-hand sides of the coupled differential equations (2) and (3). During the forward pass, a continuous-time ODE solver (e.g., Dormand-Prince method) simultaneously integrates both equations from initial conditions $(\mu(t_0), \Sigma(t_0))$ to produce trajectories of the mean $\mu(t)$and covariance $\Sigma(t)$ over the desired time interval. This joint integration tracks the complete state distribution evolution through time. Figure 4 provides a schematic overview of this parameterization and integration process.

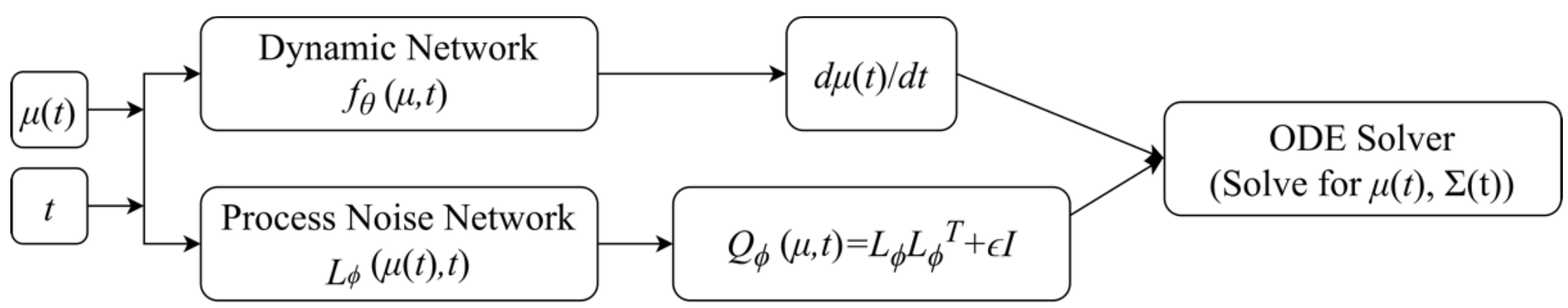

Figure 4. Neural network parameterization in UPN. The dynamics network predicts the evolution of the mean, while the process noise network produces a structured prediction of the process noise covariance. Both outputs feed into the continuous-time ODE solver that tracks the state distribution over time by integrating the coupled mean-covariance dynamics.

### *3.3. Numerical Solution*

The coupled differential equations for $\mu(t)$ and $\Sigma(t)$ in Equations (2) and (3) define an initial value problem solvable with standard ODE solvers such as the Dormand-Prince method (dopri5) or Runge-Kutta schemes. However, direct integration of the full covariance matrix $\Sigma(t) \in \mathbb{R}^{n\times n}$ would require evolving $n^2$ elements, which is computationally inefficient since $\Sigma$ is symmetric and thus contains redundant information.

To reduce computational overhead, the symmetric covariance matrix $\Sigma(t)$ is compressed using the half-vectorization operator, which extracts only the lower-triangular elements (including

the diagonal):

$$\text{vech}(\Sigma) = [\Sigma_{11}, \Sigma_{21}, \Sigma_{22}, \Sigma_{31}, \Sigma_{32}, \Sigma_{33}, \dots, \Sigma_{n1}, \Sigma_{n2}, \dots, \Sigma_{nn}]^T \tag{6}$$

This operator stacks the columns of the lower triangle in order: first all elements of column 1 from row 1 to $n$, then column 2 from row 2 to $n$, and so forth. For a concrete example, consider a $3 \times 3$ symmetric covariance matrix:

$$\Sigma = \begin{bmatrix} \Sigma_{11} & \Sigma_{21} & \Sigma_{31} \\ \Sigma_{21} & \Sigma_{22} & \Sigma_{32} \\ \Sigma_{31} & \Sigma_{32} & \Sigma_{33} \end{bmatrix}$$

The half-vectorization produces:

$$\text{vech}(\Sigma) = \begin{bmatrix} \Sigma_{11} \\ \Sigma_{21} \\ \Sigma_{22} \\ \Sigma_{31} \\ \Sigma_{32} \\ \Sigma_{33} \end{bmatrix} \in \mathbb{R}^6$$

This compression reduces dimensionality from $n^2 = 9$ to $\frac{n(n+1)}{2} = 6$ by exploiting symmetry ($\Sigma_{ij} = \Sigma_{ji}$), storing each unique element exactly once. For the general $n$-dimensional case, storage requirements decrease from $n^2$ to $\frac{n(n+1)}{2}$ elements.

To evolve the compressed representation through time, the vech operator is applied to the covariance dynamics in Equation (3). The time derivative of the vectorized covariance is given by:

$$\frac{d}{dt}\text{vech}\,(\Sigma(t)) = D^{+}\text{vec}\left(J_f\Sigma + \Sigma J_f^T + Q_\phi\right) \tag{7}$$

where $D^{+} \in \mathbb{R}^{\frac{n(n+1)}{2} \times n^2}$ is the Moore-Penrose pseudoinverse of the duplication matrix $D$, which maps half-vectorized form back to full vectorized form. The duplication matrix relationship is $\text{vec}(\Sigma) = D \cdot \text{vech}(\Sigma)$, and its pseudoinverse $D^{+}$performs the reverse operation while accounting for symmetry constraints. The vec operator (full vectorization) stacks all matrix columns, while vech extracts only unique elements; $D^{+}$bridges these representations. The complete UPN dynamics evolve both mean and compressed covariance simultaneously as an augmented state vector:

$$\frac{d}{dt}\begin{bmatrix} \mu(t) \\ \text{vech}\,(\Sigma(t)) \end{bmatrix} = \begin{bmatrix} f_\theta(\mu(t), t) \\ D^{+}\text{vec}\left(J_f\Sigma + \Sigma J_f^T + Q_\phi\right) \end{bmatrix} \tag{8}$$

This formulation enables efficient and simultaneous computation of the state and uncertainty evolution. The augmented state has dimension $n + \frac{n(n+1)}{2}$, which for $n = 3$ yields 9 total elements (3 mean + 6 covariance) compared to 12 for uncompressed representation (3 mean + 9 full covariance). For higher dimensions, the savings become more substantial: at $n = 10$, compression reduces from 100 to 55 covariance elements. Figure 5 illustrates the procedure.

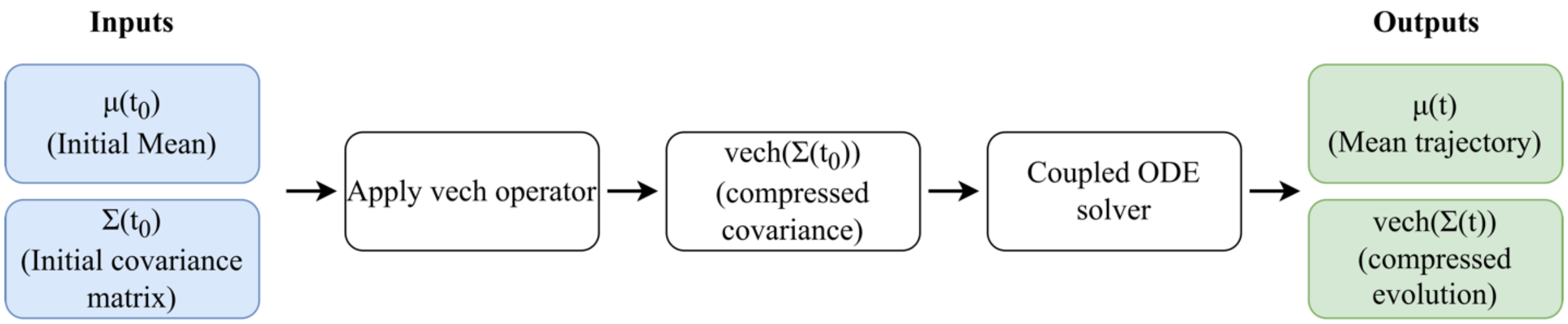

Figure 5. Efficient numerical solution procedure in UPN. The initial mean $\mu(t_0)$and symmetric covariance matrix $\Sigma(t_0)$ are processed by applying the vech operator to compress the covariance from $n^2$ to $\frac{n(n+1)}{2}$ elements by exploiting symmetry. The compressed system then evolves through a coupled ODE solver, yielding continuous-time trajectories for both the mean and the compressed covariance evolution. The covariance can be reconstructed at any time by inverting the vech operation.

### *3.4. Observations and Measurement Updates*

In many real-world scenarios, the underlying system evolves continuously over time according to stochastic dynamics, but discrete, noisy observations of the system are available only at specific time points. This section describes how UPN can incorporate these observations through a differentiable measurement update mechanism when they become available during prediction.

#### *3.4.1. State-Space Model Formulation*

Following standard state-space modeling [10], it is assumed that the latent state $x(t_i)$ evolves according to:

$$x(t_i) \sim \mathcal{N}\big(\mu(t_i), \Sigma(t_i)\big) \tag{9}$$

where $\mu(t_i)$ and $\Sigma(t_i)$ represent the predicted mean and covariance of the latent state distribution at time $t_i$, which are propagated forward from the initial condition via the coupled ODEs (Equations 2-3). The observation $y(t_i)$ received at time $t_i$ relates to the latent state through the observation model:

$$y(t_i) = Hx(t_i) + v_i, v_i \sim \mathcal{N}(0, R) \quad (10)$$

In this model, $H \in \mathbb{R}^{m \times n}$ is the observation matrix that maps the $n$-dimensional latent state to the $m$ dimensional observation space, and $v_i$ represents measurement noise with covariance $R$. In this formulation $x(t_i)$ is the true latent state (unobserved), $\mu(t_i), \Sigma(t_i)$ are the predicted distribution over $x(t_i)$ from UPN's dynamics, and $y(t_i)$ is the actual noisy observation received at time $t_i$.

*3.4.2. When Measurement Updates Are Used*

UPN supports two operational scenarios depending on whether observations arrive during the prediction horizon.

Scenario 1 is prediction without intermediate observations. When forecasting into the future, where no observations are available during the prediction period, UPN operates in pure prediction mode. The model learns dynamics by minimizing negative log-likelihood (NLL) on complete observed trajectories during training, then makes predictions by integrating the learned ODEs forward in time without any corrections. This is demonstrated in the ETTh1 time series experiments, where the model predicts 12 hours ahead without receiving any intermediate measurements.

Scenario 2 is prediction with intermediate observations. When observations become available sporadically during the prediction period (e.g., spacecraft with intermittent sensor readings), UPN can incorporate this information through Kalman updates. This filtering mode uses the measurement update equations described below.

*3.4.3. Measurement Update Procedure*

Upon receiving an observation $y(t_i)$, UPN corrects its state estimate using the Kalman update equations. For the linear observation model (Equation 10), the update is performed as:

$$K = \Sigma^-(t_i)H^T(H\Sigma^-(t_i)H^T + R)^{-1} \quad (11)$$

$$\mu^+(t_i) = \mu^-(t_i) + K\big(y(t_i) - H\mu^-(t_i)\big) \quad (12)$$

$$\Sigma^+(t_i) = (I - KH)\Sigma^-(t_i) \quad (13)$$

where $\mu^-(t_i)$ and $\Sigma^-(t_i)$ denote the predicted mean and covariance before the update (obtained by integrating Equations 2-3 forward to time $t_i$), and $\mu^+(t_i)$ and $\Sigma^+(t_i)$ denote the corrected posterior values after incorporating the observation. The Kalman gain $K$ weights the innovation $y(t_i) - H\mu^-(t_i)$ based on the relative uncertainties in the prediction and measurement.

*3.4.4. Nonlinear Observation Models*

For nonlinear observation models of the form:

$$y(t_i) = h\big(x(t_i)\big) + v_i \tag{14}$$

the method linearizes the observation function $h$ around the predicted mean:

$$H = \frac{\partial h}{\partial x}\bigg|_{\mu^-(t_i)} \tag{15}$$

and applies the same update formulas (Equations 11-13) using this locally linearized model. This procedure is analogous to the Extended Kalman Filter.

*3.4.5. Differentiability and End-to-End Training*

All operations involved in the measurement update including matrix multiplications, inversions, and residual computations, are fully differentiable with respect to the network parameters $\theta$ and $\phi$. This allows gradients to flow through both the continuous ODE dynamics (Equations 2-3) via the adjoint method (Section 4.1) and the discrete measurement updates (Equations 11-13), enabling end-to-end training.

The key architectural advantage is that measurement updates operate on the same $(\mu, \Sigma)$ representation that UPN maintains for prediction. This means:

- Models trained without updates (Scenario 1) can be deployed with updates if observations become available
- No architectural modification is required as the filtering capability is inherent to the probabilistic formulation
- In contrast, deterministic methods (e.g., standard Neural ODEs) would require substantial redesign to incorporate observations during inference

Section 5.3 demonstrates that incorporating measurements every 5 timesteps reduces prediction error by 89.6% in CubeSat experiments, while Section 5.4 shows that UPN achieves competitive performance (mean squared error (MSE) 0.589, 90.3% coverage) even without any updates during 12-step time-series forecasting.

Figure 6 depicts the predict-observe-correct cycle used in Scenario 2. The system predicts the evolution of the state distribution by solving the coupled ODEs, incorporates new information via a Kalman update when observations arrive, and then continues solving the dynamics from the corrected state.

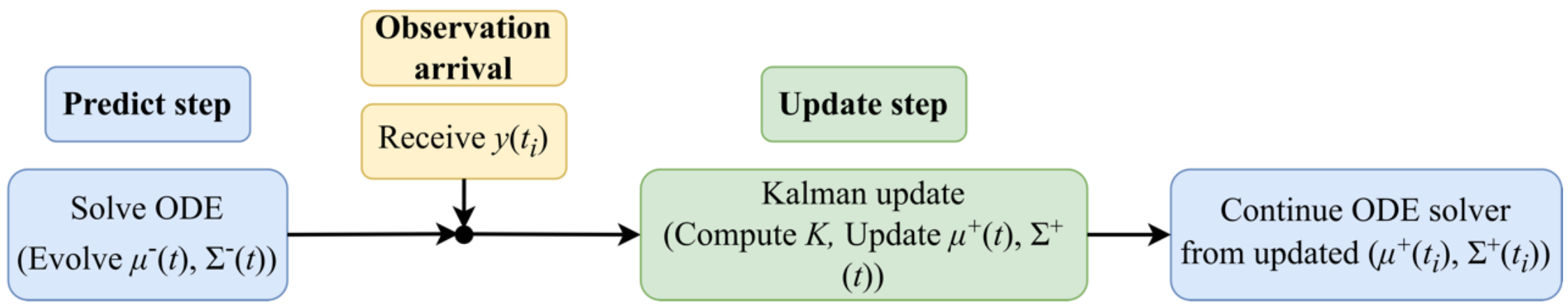

Figure 6. Predict-observe-correct flow for handling discrete observations in UPN (Scenario 2). The system predicts the evolution of the state distribution, incorporates new information via a Kalman update when observations arrive, and then continues solving the dynamics from the corrected state. This filtering mode is optional and used only when intermediate observations are available during prediction.

## 4. Computing Gradients

Training the UPN involves backpropagating through the coupled ODEs that govern both the state mean and covariance. To achieve this efficiently, the adjoint sensitivity method is extended to support joint dynamics.

### *4.1. Adjoint Sensitivity Method for Coupled ODEs*

The adjoint method enables reverse-mode differentiation by solving an auxiliary ODE backward in time. For UPN, the state combines the mean $\mu(t)$ and the half-vectorized covariance vech $(\Sigma(t))$:

$$z(t) = \begin{bmatrix} \mu(t) \\ \text{vech}\ (\Sigma(t)) \end{bmatrix} \tag{16}$$

Given a scalar loss $L$ dependent on the terminal state $z(T)$, the adjoint variable $a(t)$ is defined as:

$$a(t) = \frac{\partial L}{\partial z(t)} \tag{17}$$

Its evolution is governed by:

$$\frac{da(t)}{dt} = -a(t)^T \frac{\partial f_z(z(t), t, \theta, \phi)}{\partial z} \tag{18}$$

where $f_z$ denotes the right-hand side of the coupled ODE system. Gradients with respect to model parameters $\theta$ (dynamics) and $\phi$ (noise) are computed via:

$$\frac{dL}{d\theta} = -\int_{t_1}^{t_0} a(t)^T \frac{\partial f_z(z(t), t, \theta, \phi)}{\partial \theta} dt \tag{19}$$

$$\frac{dL}{d\phi} = -\int_{t_1}^{t_0} a(t)^T \frac{\partial f_z(z(t), t, \theta, \phi)}{\partial \phi} dt \tag{20}$$

This approach enables memory-efficient gradient computation by avoiding the storage of intermediate trajectories.

### *4.2. Efficient Jacobian-Vector Products*

Direct computation of Jacobians, especially for the covariance dynamics, involves large third-order tensors, which are computationally expensive in high dimensions. To mitigate this, automatic differentiation (AD) is used to compute vector-Jacobian products (VJPs) without explicit Jacobian construction.

- For the mean dynamics, VJPs are computed directly using reverse-mode AD.
- For terms such as $J_f\Sigma + \Sigma J_f^{\top}$, the identity

  $$\text{vec}\,(AXB) = (B^T \otimes A)\text{vec}(X) \tag{21}$$

  is applied using Kronecker products.
- Covariance-related derivatives are computed using efficient directional derivative approximations.

These strategies ensure linear scalability of time and memory complexity, especially when using diagonal or low-rank covariance structures.

### *4.3. Backpropagation Through Measurement Updates*

When discrete-time observations are present, gradients must also propagate through the measurement update mechanism described in Section 3.4. This requires differentiating through matrix operations, including inversions and multiplications.

Standard automatic differentiation tools are employed to handle these computations, with additional steps taken to ensure numerical stability. Techniques such as singular value decomposition (SVD) and regularization terms are applied during inversion to prevent instability.

The overall training process combines:

- Adjoint-based differentiation for continuous-time dynamics, and
- Differentiable updates for observation corrections.

This enables end-to-end optimization of all network parameters. Figure 7 illustrates the full training loop, where the forward pass computes the terminal state and loss, and the backward pass solves the adjoint system. Algorithm 1 outlines the training procedure using the adjoint method.

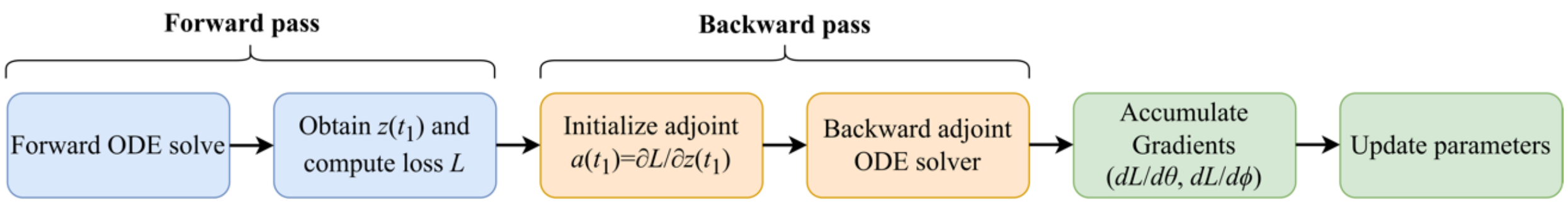

Figure 7. Training flow for UPN using adjoint-based gradient computation. The forward pass solves the continuous-time ODE to produce the final state and loss. The backward pass solves an adjoint ODE in reverse time to accumulate gradients with respect to model parameters, enabling efficient memory usage and scalable optimization.

**Algorithm 1: UPN Training with Adjoint Method**

Input: Training data D, initial parameters θ (dynamics), ϕ (noise), learning rate η
Output: Optimized parameters θ*, ϕ*

procedure TRAIN_UPN(D, θ, ϕ, η)
    for each batch B in D do
        // Extract batch data
        $\{x_1:T^{(i)}, t_1:T^{(i)}\}_{i=1}^{N} \leftarrow B$ ▷ Observed trajectories and time points

        // Initialize from observations
        $\mu_0 \leftarrow$ encode_initial_mean($x_{1:h}$) ▷ Extract initial mean from history
        $\Sigma_0 \leftarrow$ initialize_covariance() ▷ Initial covariance (learned or fixed)
        $z_0 \leftarrow [\mu_0, \mathrm{vech}(\Sigma_0)]$ ▷ Concatenate for joint ODE

        // Forward Pass ($t_0 \rightarrow T$)
        z_traj ← ODEsolve(UPN_DYNAMICS, $z_0$, $[t_0, t_1, ..., t_t]$, θ, ϕ)
        $\{\mu(t_i), \Sigma(t_i)\}_{i=1}^{T}$ ← extract_from_trajectory(z_traj)

        // Compute Loss
        $\mathscr{L} \leftarrow$ NLL($\{\mu(t_i), \Sigma(t_i)\}_{i=1}^{T}, \{x(t_i)\}_{i=1}^{T}$) ▷ Negative log-likelihood

        // Backward Pass via Adjoint Method ($T \rightarrow t_0$)
        $\partial\mathscr{L}/\partial\theta, \partial\mathscr{L}/\partial\phi$ ← compute_gradients_adjoint($\mathscr{L}$, z_traj, θ, ϕ)

```
    // Parameter Update
    θ ← θ - η · ∂ℒ/∂θ                ▷ Update dynamics parameters
    ϕ ← ϕ - η · ∂ℒ/∂ϕ                ▷ Update noise parameters
  end for
  return θ, ϕ
end procedure

function UPN_DYNAMICS(t, z, θ, ϕ)
  μ ← z[1:n]                         ▷ Extract mean (first n elements)
  σ_vec ← z[n+1:end]                 ▷ Extract vech(Σ)
  Σ ← unvech(σ_vec, n)               ▷ Reconstruct covariance matrix

  dμ/dt ← f_θ(μ, t)                  ▷ Equation (2): Mean dynamics
  J ← ∂f_θ/∂μ |_(μ,t)                ▷ Jacobian via automatic differentiation
  Q ← Q_φ(μ, t)                      ▷ Equation (5): Process noise

  dΣ/dt ← JΣ + ΣJᵀ + Q               ▷ Equation (3): Covariance dynamics

  dz/dt ← [dμ/dt, vech(dΣ/dt)]       ▷ Concatenate derivatives
  return dz/dt
end function

function COMPUTE_GRADIENTS_ADJOINT(ℒ, z_traj, θ, ϕ)
  // Uses automatic differentiation through ODE solver
  // Implementation provided by torchdiffeq.odeint_adjoint
  ∂ℒ/∂θ ← autograd(ℒ, θ)             ▷ Via adjoint sensitivity method
  ∂ℒ/∂ϕ ← autograd(ℒ,ϕ)              ▷ Via adjoint sensitivity method
  return ∂ℒ/∂θ, ∂ℒ/∂ϕ
end function
```

### *4.4. Computational Complexity Analysis*

The computational complexity of UPN warrants detailed examination, as the coupled mean-covariance dynamics involve operations that scale with the latent space dimension $n$. The half-vectorization operator (Equation 6) reduces covariance storage from $n^2$ to $\frac{n(n+1)}{2}$ elements by exploiting symmetry. This results in $\mathcal{O}(n^2)$ storage complexity for the augmented state vector $z(t) = [\mu(t), \text{vech}(\Sigma(t))]$, where $\mu \in \mathbb{R}^n$ and $\text{vech}(\Sigma) \in \mathbb{R}^{n(n+1)/2}$.

Each evaluation of the coupled ODE system (Equations 2-3) requires the following computations:

1. Mean Dynamics (Equation 2)

The neural network forward pass for $f_\theta(\mu(t), t)$ has complexity $\mathcal{O}(n \times H + H^2)$, where $H$ is the hidden dimension. For $H = 4n$, which is used in our architecture, the cost becomes $\mathcal{O}(n^2)$.

2. Jacobian Computation

The Jacobian $J = \frac{\partial f_\theta}{\partial \mu}$ requires computing gradients of $n$ outputs with respect to $n$ inputs. Using automatic differentiation, this involves $n$ separate backward passes through the dynamics network. Each backward pass has $\mathcal{O}(n^2)$ complexity through network layers, leading to a total Jacobian cost of $\mathcal{O}(n^3)$.

3. Covariance Dynamics (Equation 3)

Computing $\frac{d\Sigma}{dt} = J\Sigma + \Sigma J^T + Q$ involves matrix multiplication $J\Sigma$: $\mathcal{O}(n^3)$ and $\Sigma J^T$: $\mathcal{O}(n^3)$, process noise $Q = L_\phi L_\phi^T + \epsilon I$ : $\mathcal{O}(n^3)$, and matrix additions: $\mathcal{O}(n^2)$. Thus, the subtotal for covariance dynamics is $\mathcal{O}(n^3)$.

4. Vectorization Operations

Applying the vech $(\cdot)$ operator requires $\mathcal{O}(n^2)$ computational cost. Similarly, reconstructing the covariance matrix using the unvech operation also incurs a computational cost of $\mathcal{O}(n^2)$.

Per-step total complexity is $\mathcal{O}(n^3)$, dominated by the Jacobian computation and matrix multiplications in the covariance dynamics.

To validate this theoretical complexity, systematic benchmarks were conducted on a desktop workstation with Intel Core i7-14700F processor (2.10 GHz), 32 GB RAM (4800 MT/s), and NVIDIA GeForce RTX 5060 GPU (8 GB), using batch size of 32 and observation dimension of 3. All experiments throughout this paper were performed on this hardware configuration. Table 2 presents timing measurements across different latent dimensions, and Figure 8 shows a comprehensive scaling analysis.

Table 2. Empirical scaling of UPN on NVIDIA GeForce RTX 5060.

| $n$ | Parameters | Forward Pass (ms) | Training Step (ms) | Memory (MB) |
|---|---|---|---|---|
| 2 | 261 | 1.67 ± 0.32 | 4.52 ± 0.90 | 17.3 |
| 4 | 892 | 2.35 ± 0.41 | 7.49 ± 1.94 | 17.3 |
| 8 | 4,231 | 3.93 ± 0.47 | 12.59 ± 2.33 | 17.5 |
| 16 | 20,555 | 8.18 ± 1.78 | 28.30 ± 3.08 | 18.2 |

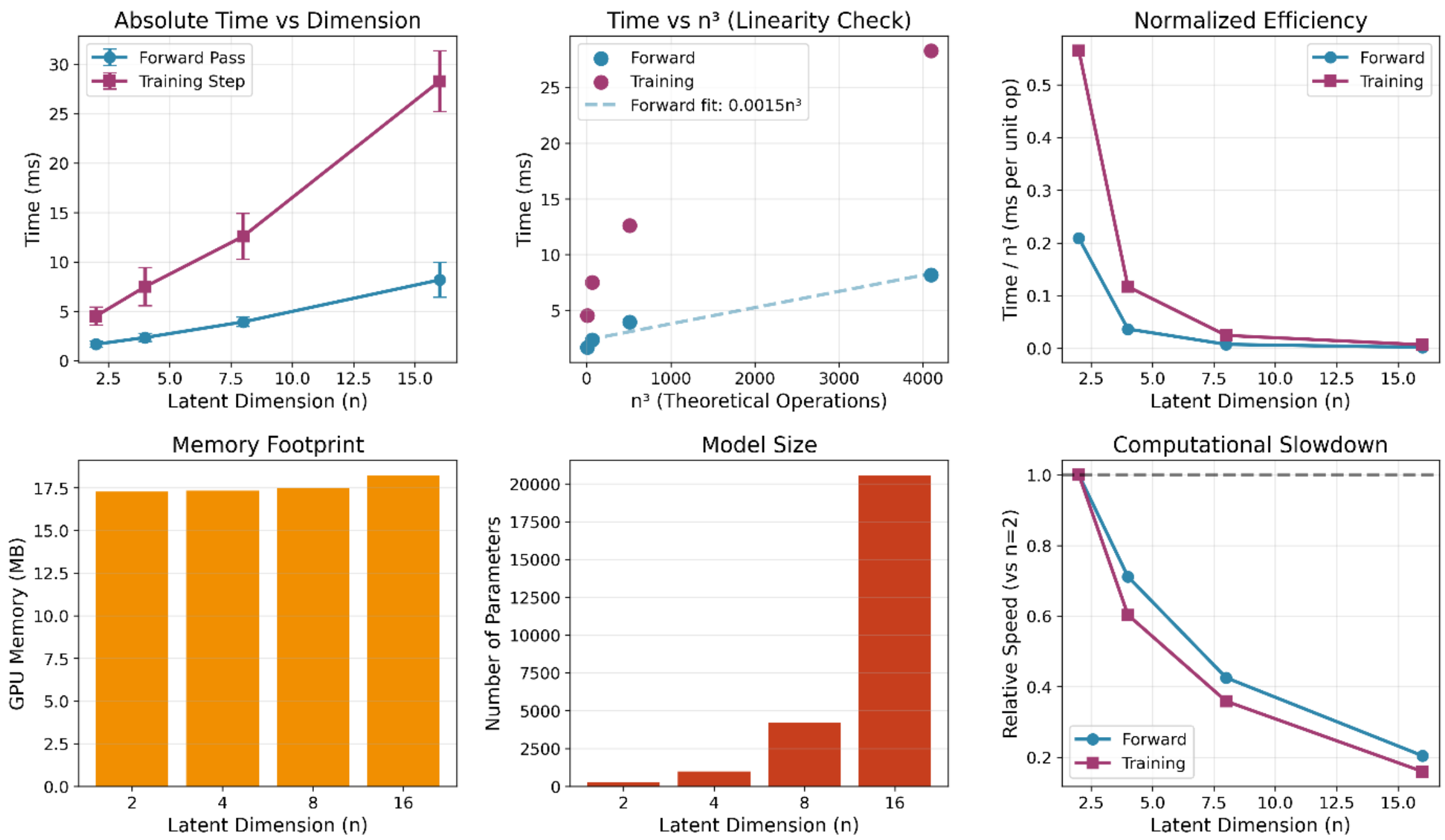

Figure 8. UPN computational scaling analysis on NVIDIA GeForce RTX 5060. (Top) Forward pass and training step times grow with dimension (left), exhibiting linear scaling with $n^3$ (center) and improving normalized efficiency at larger dimensions (right). (Bottom) Memory footprint remains nearly constant (left), model size grows quadratically with parameters (center), and relative computational speed decreases to 20% at $n$=16 compared to $n$=2 baseline (right).

The empirical measurements confirm the $\mathcal{O}(n^3)$ theoretical complexity. Figure 8 (center top panel) demonstrates near-linear scaling when plotting forward pass time against $n^3$, with a fitted slope of 0.0015 ms per unit $n^3$. This linear relationship validates that computational cost grows cubically with latent dimension. The normalized efficiency metric (Figure 8, top right) shows time per $n^3$ decreasing from 0.21 ms at $n = 2$ to 0.002 ms at $n = 16$, reflecting GPU parallelization benefits where larger problem sizes better utilize parallel computing resources through improved memory access patterns and increased arithmetic intensity. Several important insights can be drawn from the empirical analysis:

- Cubic scaling confirmed: Time scales linearly with $n^3$ (Figure 8, center top), with fitted slope 0.0015 ms/$\text{n}^3$, validating theoretical $\mathcal{O}(n^3)$ complexity
- GPU efficiency: Despite a 512-fold increase in theoretical operations ($n^3$: 8 → 4096 when $n$ increases from 2 to 16), wall-clock time increases only 4.9-fold (1.67 → 8.18 ms), demonstrating effective parallelization

- Training overhead: Training steps require 2.7-3.5 × more time than forward passes due to gradient computation through the adjoint system
- Memory efficiency: GPU memory remains nearly constant (17-18 MB) across tested dimensions, dominated by model parameters rather than intermediate computations
- Practical tractability: Even at $n = 16$ with 20,555 parameters, forward passes complete in under 10 ms, suitable for real-time applications

Despite $\mathcal{O}(n^3)$ cost per ODE step, UPN achieves analytical uncertainty quantification in a single forward pass. From our ETTh1 time series experiments (Section 5.4), Neural SDE with 100 stochastic samples requires 6.6 × more inference time than UPN while achieving 14% worse MSE (0.669 vs 0.589). This computational advantage stems from UPN's deterministic uncertainty propagation through coupled ODEs versus repeated stochastic trajectory integration.

Moreover, several approaches could reduce complexity for higher-dimensional systems.

- First, low-rank covariance approximation [40] represents $\Sigma \approx LL^T$, where $L \in \mathbb{R}^{n \times r}$ with $r \ll n$. This reduces storage to $\mathcal{O}(nr)$ and matrix operations to $\mathcal{O}(n^2 r)$. This approach is particularly effective when the intrinsic dimensionality of uncertainty is lower than the state dimension.
- Second, structured covariance constraints [41] such as diagonal or block-diagonal forms reduce computational cost to $\mathcal{O}(n^2)$ or $\mathcal{O}(b^2 k)$ for $k$ blocks of size $b$, when domain knowledge suggests sparse correlations between state components.
- Third, sparse Jacobian exploitation [42] can be applied when the network architecture induces sparsity (for example, in locally-connected layers). In such cases, complexity reduces to $\mathcal{O}(\text{nnz} \times n)$, where nnz is the number of non-zero Jacobian elements. This can potentially achieve order-of-magnitude speedups for high-dimensional systems.
- Fourth, mixed-precision computation [43] involves using FP16 for matrix operations with FP32 accumulation. This can yield a $2 - 6 \times$ speedup on modern GPUs with minimal accuracy loss, which is particularly beneficial for the covariance propagation component that dominates the computational cost.

## 5. Experimental Evaluation

### *5.1. Non-Chaotic Dynamical Systems*

To validate UPN's uncertainty quantification capabilities on dynamical systems, comprehensive experiments were conducted on four canonical non-chaotic systems that represent a range of different dynamical behaviors. A single-point initialization approach was used, where predictions are made from a single noisy observation. The initial uncertainty was set to match the observation noise variance ($\Sigma_0 = \sigma_{obs}^2 I$). This setup is appropriate for Markovian systems, where the current state holds all the necessary information to predict future behavior.

#### *5.1.1. Benchmark Systems*

Four different systems with distinct properties were chosen to test UPN's performance across various types of dynamics:

1. Damped Harmonic Oscillator: This is a fundamental linear system that describes oscillatory motion with damping. The governing equations are:

$$\frac{dx}{dt} = v \tag{22}$$

$$\frac{dv}{dt} = -\frac{k}{m}x - \frac{c}{m}v \tag{23}$$

   With spring constant $k = 1.0$, mass $m = 1.0$, and damping coefficient $c = 0.1$. This system exhibits exponentially decaying oscillations and serves as a baseline for linear dynamics.

2. Van der Pol Oscillator: A nonlinear system that exhibits limit cycle behavior:

$$\frac{dx}{dt} = y \tag{24}$$

$$\frac{dy}{dt} = \mu(1 - x^2)y - x \tag{25}$$

   where $\mu = 0.5$. The Van der Pol oscillator is a well-known example of self-sustained oscillations due to nonlinear damping, and is commonly studied in electrical circuits and biological systems.

3. Linear 2D System: A coupled linear system with complex eigenvalues, described by:

$$\frac{d\mathrm{x}}{dt} = A\mathrm{x}, A = \begin{bmatrix} -0.1 & 0.5 \\ -0.5 & -0.1 \end{bmatrix} \tag{26}$$

This system results in stable spiraling behavior and tests UPN's ability to model coupled dynamics across multiple dimensions.

4. Damped Pendulum: A nonlinear physical system governed by:

$$\frac{d\theta}{dt} = \omega \tag{27}$$

$$\frac{d\omega}{dt} = -\frac{g}{l}\sin(\theta) - \frac{c}{l}\omega \tag{28}$$

with gravitational acceleration $g = 9.81\ \mathrm{m/s^2}$, length $l = 1.0$ m, and damping coefficient $c = 0.1$. This system includes nonlinear restoring forces and represents real-world physical dynamics, commonly seen in robotics and control applications.

*5.1.2. Data Generation*

For each system, 50 trajectories were generated using the RK45 integration method with tight tolerances (rtol $= 10^{-8}$, atol $= 10^{-10}$) to ensure numerical accuracy. Trajectories were sampled at $dt = 0.1$ s over 20 seconds, yielding 200 time points per trajectory. Gaussian observation noise with standard deviation $\sigma_{obs} = 0.05$ was added to all states ($\sigma_{obs} = 0.1$ for the Linear System to increase task difficulty). Initial conditions were randomly sampled from uniform distributions $[-1,1]$ for the oscillators and linear system, and $[-\pi/4, \pi/4]$ radians for the pendulum.

The data were split into 70% training (35 trajectories), 15% validation (7 trajectories), and 15% test (8 trajectories). To ensure non-overlapping samples and prevent data leakage, a stride of 20 time steps was used between consecutive training samples, with each sample consisting of a single initial observation and 20 future time steps (2 -second prediction horizon). This resulted in 315 training samples, 63 validation samples, and 72 test samples per system.

*5.1.3. Baseline Methods*

UPN was compared against two baseline approaches:

- Ensemble Neural ODE: An ensemble of 10 independently trained Neural ODEs with different random initializations. This represents current best practice for uncertainty quantification in learned dynamical systems and captures epistemic uncertainty through bootstrap aggregation. Uncertainty estimates are obtained by computing mean and covariance across ensemble predictions.
- Deterministic Neural ODE: A standard Neural ODE without uncertainty quantification,

serving as a baseline to demonstrate the value of probabilistic predictions.

*5.1.4. Implementation Details*

All methods used identical network architectures: 2 hidden layers with 64 units each and Tanh activation functions. The dynamics networks $f_\theta(\mu, t)$ and process noise networks $Q_\phi(\mu, t)$ (for UPN only) took concatenated state-time inputs $[\mu; t]$ and produced state derivatives and process noise covariance matrices, respectively.

Training was performed using the Adam optimizer with learning rate $10^{-3}$ and weight decay $10^{-4}$ for 60 epochs. Early stopping with patience of 10 epochs was employed based on validation loss. For UPN, the loss function was NLL under Gaussian predictive distributions. Ensemble and Deterministic models were trained using MSE loss. Gradient clipping with maximum norm 1.0 was applied to ensure training stability.

ODE integration during training and inference used the Dormand-Prince method (dopri5) with tolerances rtol $= 10^{-4}$ and atol $= 10^{-6}$, balancing accuracy and computational efficiency.

*5.1.5. Evaluation Metrics*

Performance was assessed using four metrics:

- MSE: Measures point prediction accuracy,

$$\text{MSE} = \frac{1}{N}\sum_{i=1}^{N} |\mathrm{x}_i - \mu_i|^2 \tag{29}$$

- NLL: Evaluates probabilistic prediction quality under Gaussian assumptions (applicable to UPN only),

$$\text{NLL} = -\frac{1}{N}\sum_{i=1}^{N} \log\, p(\mathrm{x}_i \mid \mu_i, \Sigma_i) \tag{30}$$

- 95% Confidence Interval Coverage: Fraction of true states falling within predicted 95% confidence intervals, computed as the proportion of ground truth points $x$ where

$$(x - \mu)^T \Sigma^{-1} (x - \mu) \le \chi^2_{0.95}(d) \tag{31}$$

- Calibration Error: Absolute deviation from target coverage, |Coverage - 0.95| with values close to zero indicating well-calibrated uncertainty estimates.

### *5.1.6. Results*

Table 3 presents comprehensive results across all systems and methods. Figure 9 shows the training convergence, demonstrating stable learning across all systems with appropriate regularization preventing overfitting.

Table 3. Quantitative Results on Non-Chaotic Dynamical Systems.[†]

| System | Method | MSE ($\times 10^{-3}$) | NLL | Coverage | Cal. Error |
|---|---|---|---|---|---|
| Damped Harmonic Oscillator | UPN | 5.23 | -2.17 | **0.997** | 0.047 |
| | Ensemble | **5.02** | — | 0.275 | 0.675 |
| | Deterministic | 5.29 | — | — | — |
| Van der Pol | UPN | 14.86 | -1.45 | **0.949** | **0.001** |
| | Ensemble | **10.47** | — | 0.341 | 0.609 |
| | Deterministic | 13.71 | — | — | — |
| Linear System | UPN | 16.99 | -1.17 | **0.953** | **0.003** |
| | Ensemble | **17.34** | — | 0.083 | 0.867 |
| | Deterministic | 17.12 | — | — | — |
| Damped Pendulum | UPN | 15.99 | -1.56 | **0.968** | 0.018 |
| | Ensemble | **11.52** | — | 0.341 | 0.609 |
| | Deterministic | 13.66 | — | — | — |
| Average | UPN | 13.27 | -1.59 | **0.967** | **0.017** |
| | Ensemble | **11.08** | — | 0.260 | 0.690 |
| | Deterministic | 12.45 | — | — | — |

[†] Bold indicates best performance per metric per system.

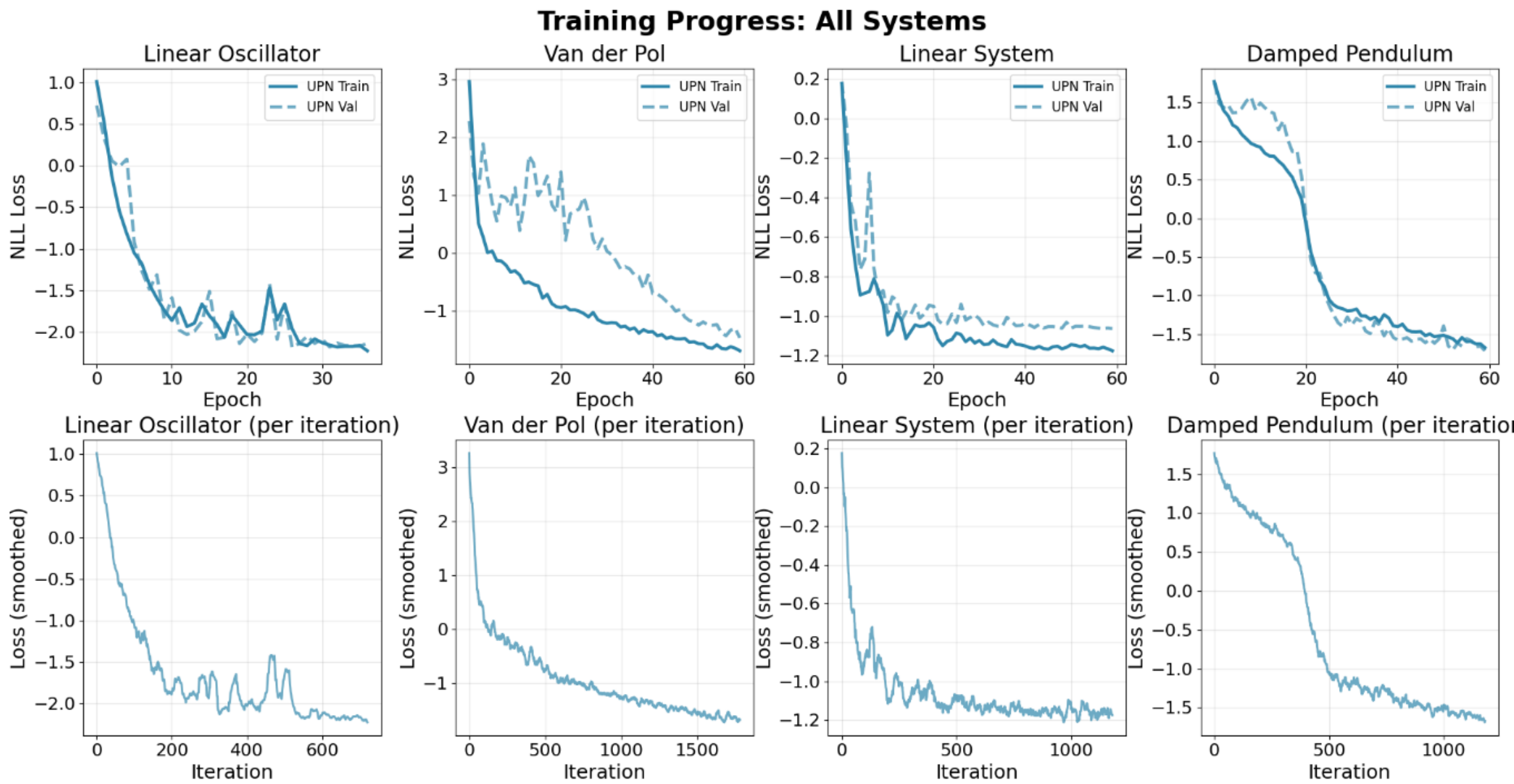

Figure 9. Training and validation loss curves (per epoch and per iteration) for UPN across all four dynamical systems.

The results reveal a striking contrast between point prediction accuracy and uncertainty calibration. While Ensemble methods achieve marginally lower MSE (16% improvement on average), they catastrophically fail at uncertainty quantification with only 26% coverage versus UPN's 96.7% coverage. This demonstrates that point prediction accuracy and uncertainty calibration represent fundamentally different objectives requiring different modeling approaches.

Figure 10 visualizes calibration performance across systems. UPN achieves near-perfect calibration with an average error of only 1.7%, maintaining coverage within the acceptable range for all systems. System-specific results demonstrate consistent performance:

- Linear System: 95.3% coverage (0.3% error), essentially perfect calibration
- Van der Pol: 94.9% coverage (0.1% error), remarkable given the nonlinear limit cycle dynamics
- Damped Pendulum: 96.8% coverage (1.8% error), excellent for a physical system with nonlinear restoring forces
- Damped Harmonic Oscillator: 99.7% coverage (4.7% error), slight overestimation indicating conservative uncertainty bounds

In stark contrast, Ensemble methods severely underestimate uncertainty across all systems (26.0% average coverage, 69% calibration error), with the Linear System showing catastrophic failure at only 8.3% coverage. This occurs because ensemble methods capture model uncertainty (variability due to random initialization) but fail to account for aleatoric uncertainty (observation noise and process stochasticity), which UPN explicitly models through the learned process noise covariance $Q(\mu, t)$.

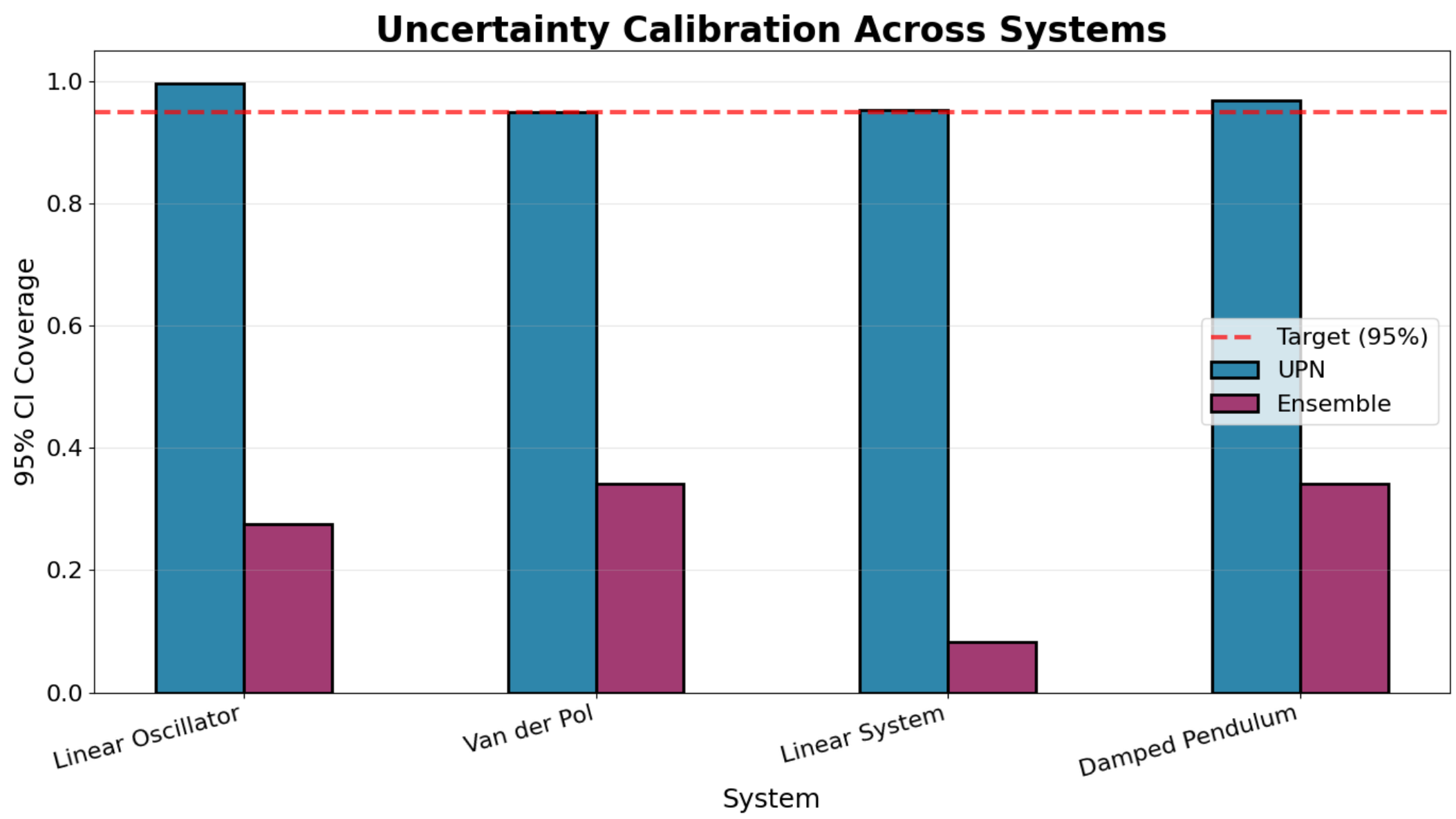

Figure 10. 95% confidence interval coverage comparison between UPN and Ensemble methods across all systems.

### *5.1.7. Trajectory Predictions with Uncertainty*

Figures 11-14 present representative trajectory predictions for each system, revealing how uncertainty propagates through dynamics.

Damped Harmonic Oscillator (Figure 11): UPN's uncertainty bands (blue shaded region) appropriately widen over time as the prediction horizon increases, consistently containing the ground truth trajectory (solid black line). The initial noisy observation at $t = 0$ (leftmost black dot) lies within the confidence interval, validating the initial covariance specification $\Sigma_0 = \sigma^2{}_{obs} I$. The subsequent noisy observations (black dots) along the trajectory predominantly fall within UPN's 95% CI, confirming well-calibrated uncertainty estimates. Notably, the velocity dimension exhibits faster uncertainty growth compared to the position dimension, reflecting the accumulated effect of uncertainty propagation through the coupled dynamics $\frac{dx}{dt} = y$. While Ensemble predictions (purple dotted line with pink shaded 95% CI) track a similar mean trajectory, the Ensemble confidence intervals are substantially narrower and fail to contain many of the ground truth observations, demonstrating severe underestimation of predictive uncertainty. The Deterministic method (orange dashed line) provides no uncertainty quantification.

Van der Pol Oscillator (Figure 12): Despite the system's nonlinear limit cycle behavior, UPN maintains excellent calibration across both state dimensions. The initial noisy observation and subsequent ground truth points predominantly fall within UPN's 95% CI, confirming well-calibrated uncertainty estimates. The y dimension (bottom panel) shows more uniform uncertainty growth, reflecting the derivative relationship $dy/dt = \mu(1 - x^2)y - x$. Notably, the Ensemble mean (purple dotted line with pink shaded 95% CI) tracks closer to ground truth in the $x$ dimension compared to UPN and Deterministic predictions, yet its confidence intervals remain severely underestimated, failing to contain many observations. This system proved most challenging during training (Figure 9, second panel), requiring 60 epochs to achieve convergence with careful regularization, but the final model demonstrates robust uncertainty propagation through stiff nonlinear dynamics. The phase space representation (Figure 15b) further illustrates how UPN's uncertainty structure adapts to the limit cycle geometry.

Linear System (Figure 13): This system exhibits UPN's strongest calibration performance (95.3%). The coupled dynamics are accurately captured, with uncertainty growing appropriately in both state dimensions. Notably, State 2 exhibits larger uncertainty due to the system's eigenstructure, demonstrating that UPN learns dimension-specific uncertainty characteristics. The spiral convergence pattern is preserved across all methods, but only UPN provides reliable confidence bounds.

Damped Pendulum (Figure 14): A physically realistic system demonstrating UPN's applicability to real-world scenarios. The pendulum's oscillatory motion with damping is accurately predicted, and uncertainty bands properly reflect the nonlinear restoring force -sin($\theta$) near equilibrium points. Both angular position and angular velocity predictions maintain calibration throughout the oscillation cycle, with uncertainty growing more rapidly during high-velocity phases where nonlinear effects are pronounced.

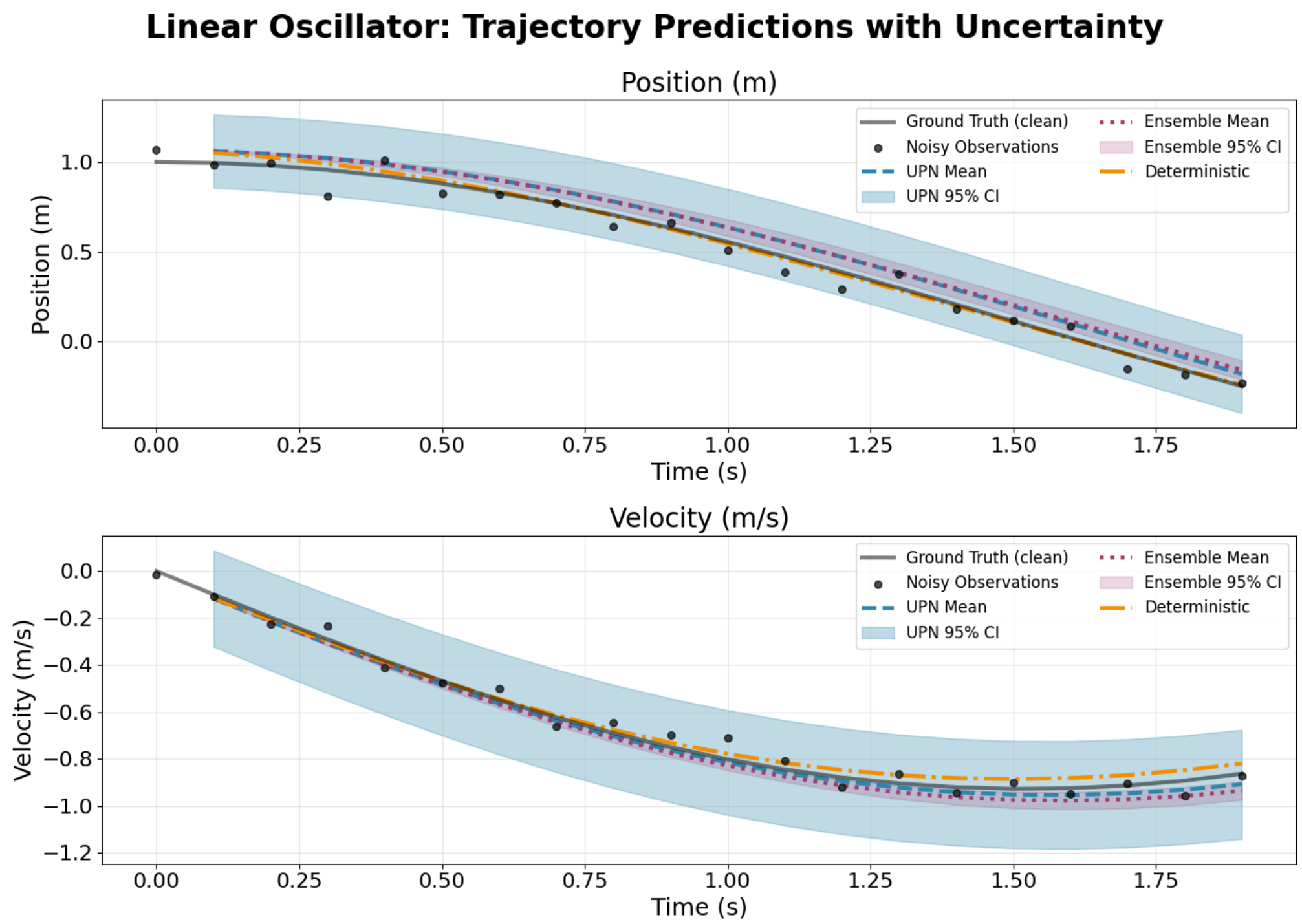

Figure 11. Trajectory predictions with 95% confidence intervals for the Damped Harmonic Oscillator in position and velocity dimensions.

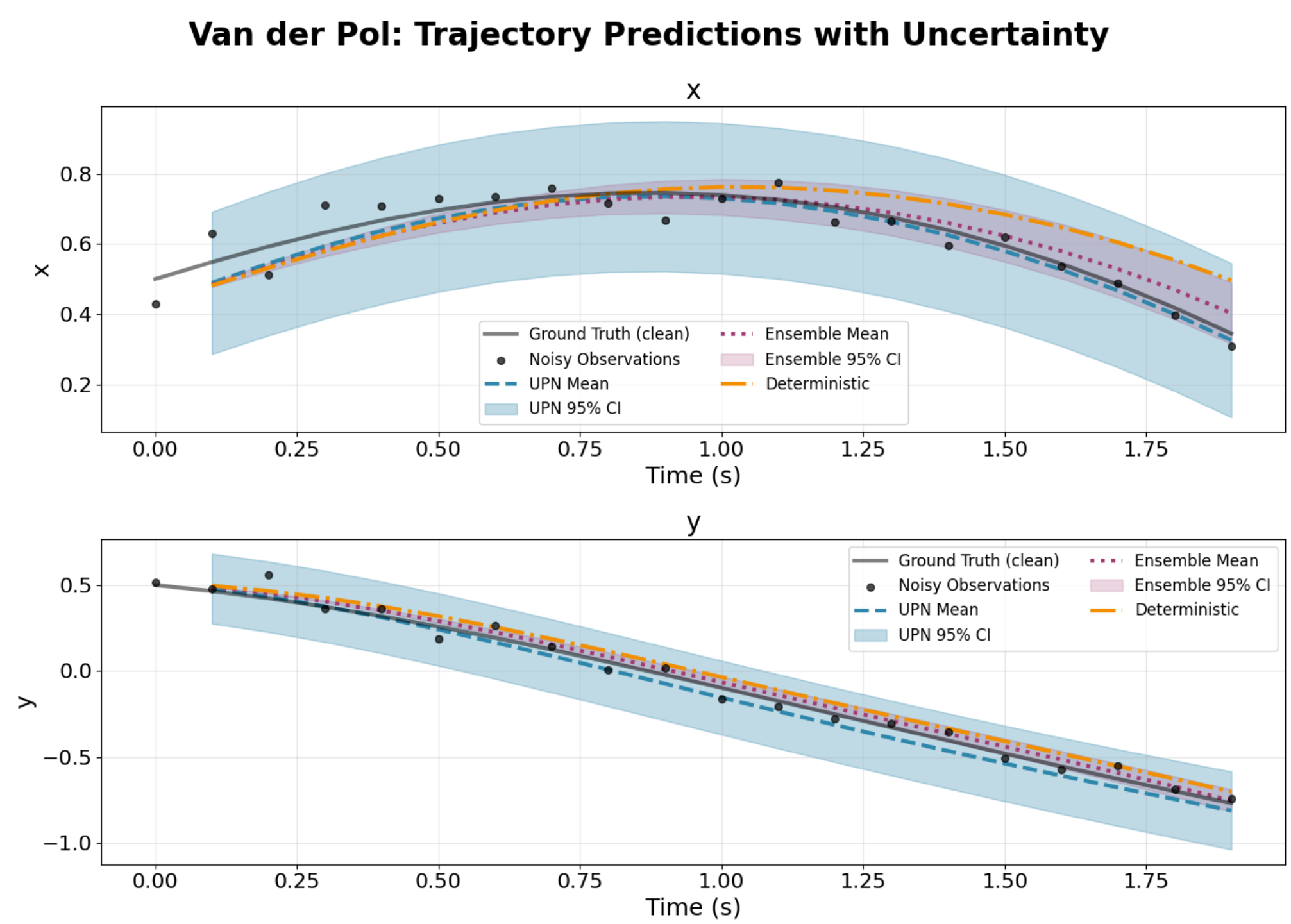

Figure 12. Trajectory predictions with 95% confidence intervals for the Van der Pol Oscillator in *x* and *y* dimensions.

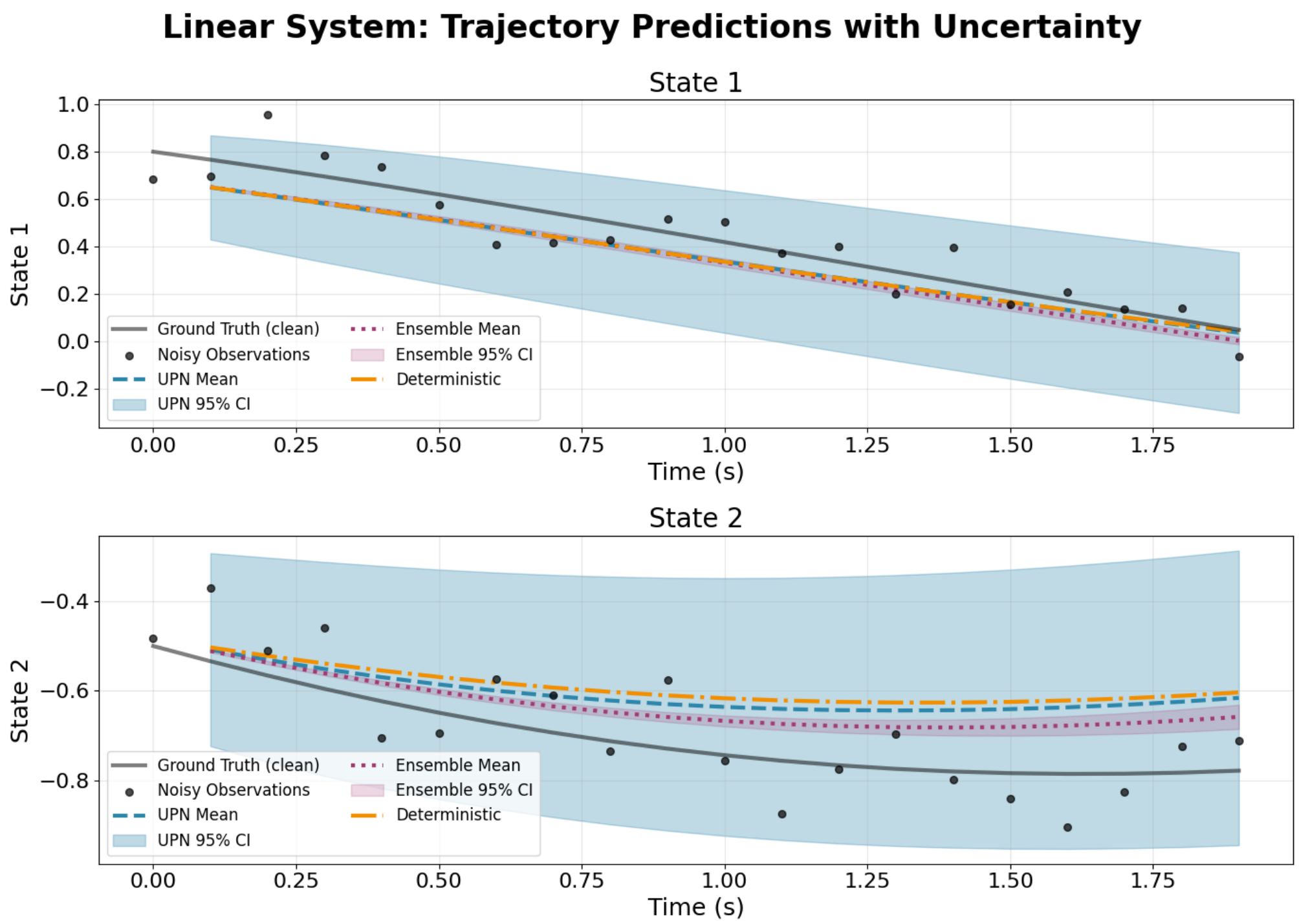

Figure 13. Trajectory predictions with 95% confidence intervals for the Linear System in both state dimensions.

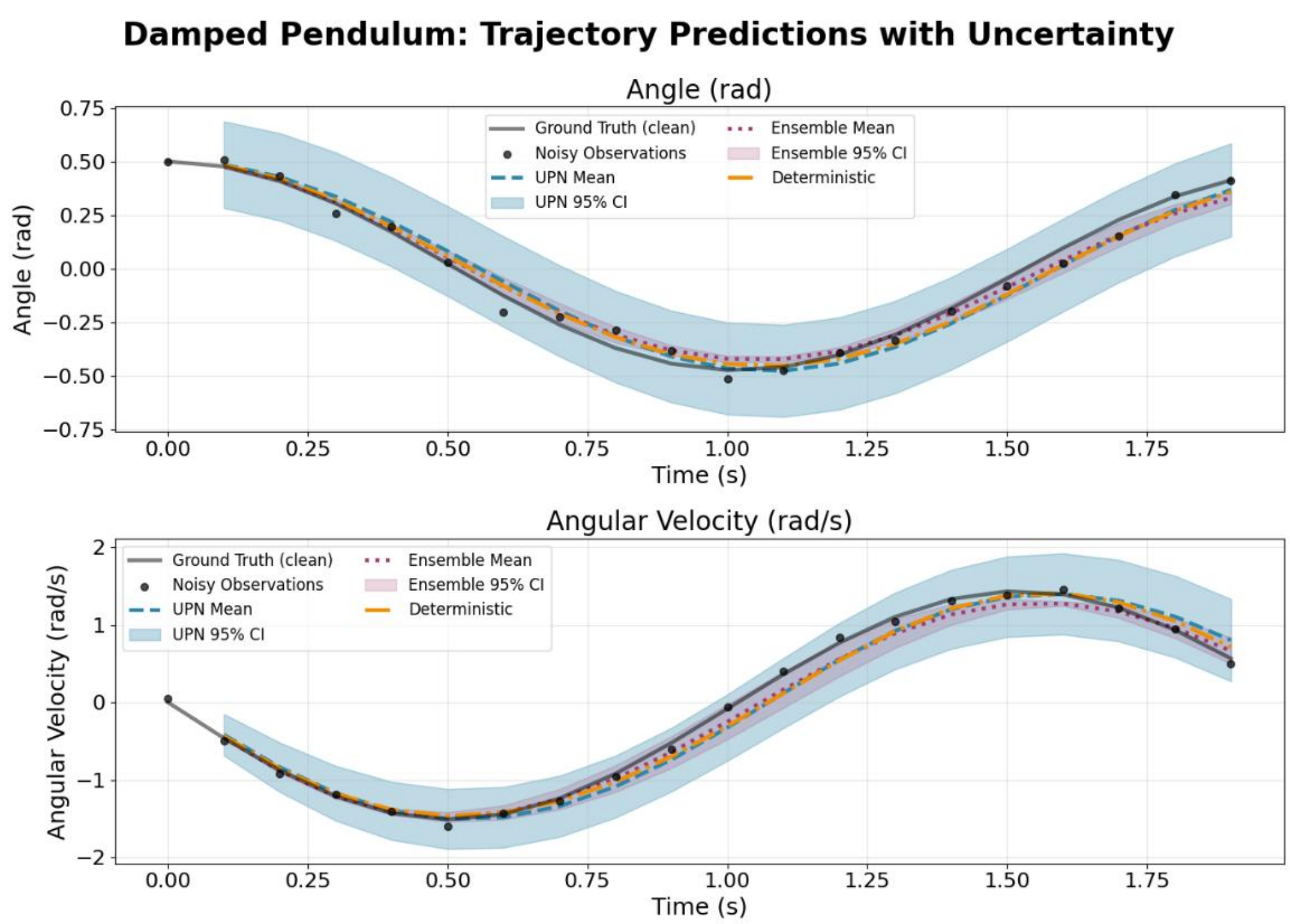

Figure 14. Trajectory predictions with 95% confidence intervals for the Damped Pendulum in angle and angular velocity.

*5.1.8. Phase Space Analysis*

Figure 15 presents phase space trajectories, providing geometric insight into prediction quality and uncertainty structure. UPN's uncertainty ellipses (displayed at select points along trajectories) appropriately expand along the predicted path, with their orientation reflecting the local Jacobian structure $\frac{\partial f_\theta}{\partial \mu}$ learned by the dynamics network. The phase portraits confirm that UPN captures correct qualitative behavior (spiral convergence for Damped Harmonic Oscillator, limit cycle for Van der Pol, circular motion for Damped Pendulum). Additionally, covariance structure (ellipse orientation and aspect ratio) adapts to local dynamics, being elongated along the flow direction where dynamics are stiff.

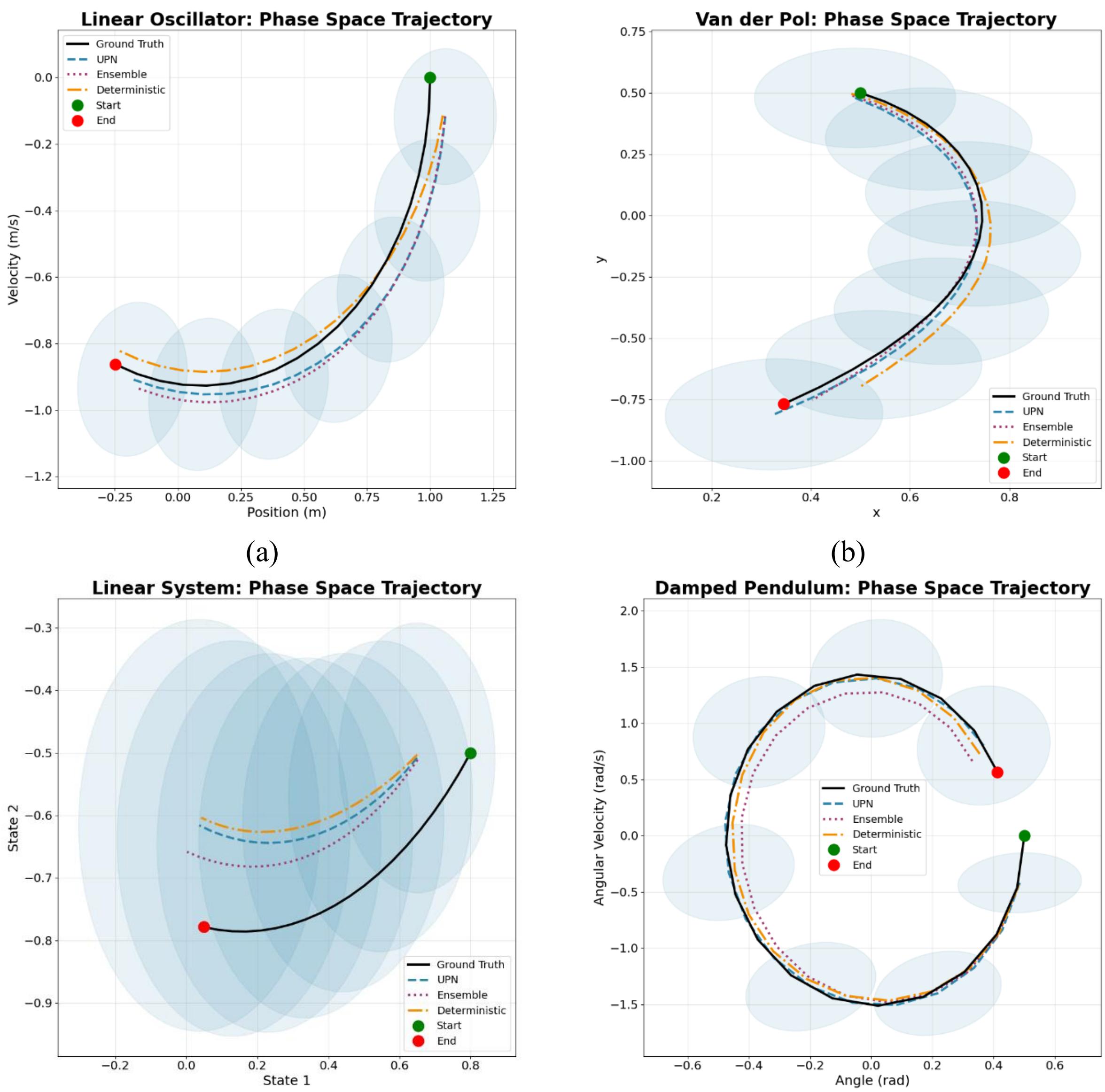

(c) (d)

Figure 15. Phase space trajectories with uncertainty regions for (a) Damped Harmonic Oscillator, (b) Van der Pol, (c) Linear System, and (d) Damped Pendulum.

*5.1.9. Uncertainty Evolution*

Figure 16 quantifies how total uncertainty (measured as $\mathrm{Tr}(\Sigma)$, the trace of the covariance matrix) evolves over the prediction horizon. UPN exhibits consistent uncertainty growth across all systems, with growth rates reflecting intrinsic system properties:

- Linear systems (Damped Harmonic Oscillator, Linear System): Near-linear uncertainty growth, consistent with analytical results for linear systems where uncertainty grows polynomially
- Damped Pendulum: Nonlinear growth reflecting phase-dependent dynamics, with steeper growth during rapid motion phases
- Van der Pol: Steady growth reflecting limit cycle dynamics, where uncertainty accumulates uniformly around the attractor

Critically, Ensemble uncertainty remains nearly constant or grows minimally, failing to capture the fundamental accumulation of uncertainty through sequential dynamics, a fatal flaw for long-horizon prediction tasks. This behavior arises because ensemble spread reflects only initialization differences, not the compounding effect of observation noise through dynamics.

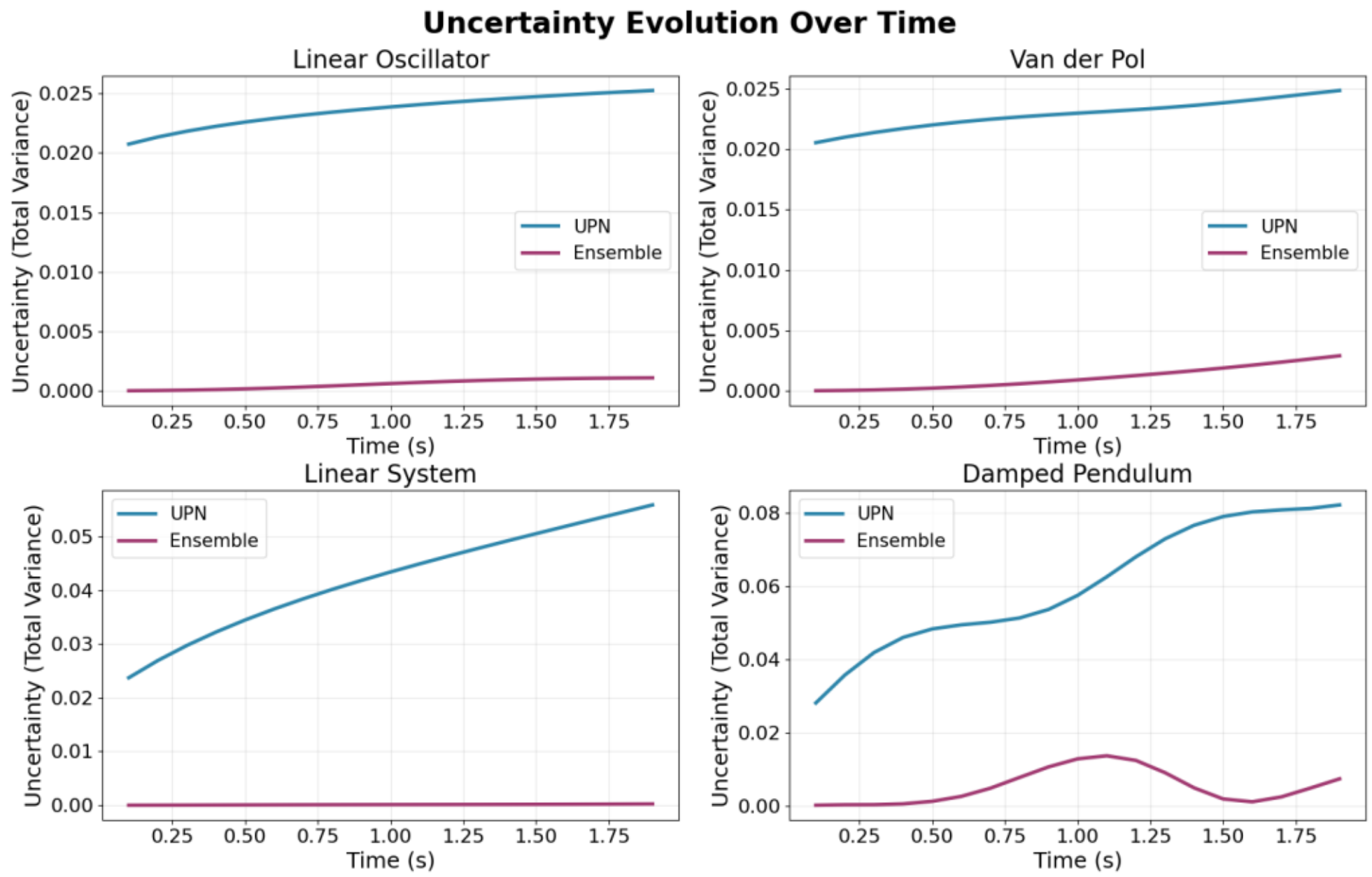

Figure 16. Evolution of total uncertainty (trace of covariance) over prediction horizon for UPN and Ensemble methods across all systems.

*5.1.10. Discussion*

The experimental results reveal a fundamental distinction between point prediction accuracy and uncertainty calibration in learned dynamical systems. UPN achieves near-optimal uncertainty calibration (96.7% average coverage, 1.7% calibration error) across all tested systems, while ensemble baselines catastrophically underestimate uncertainty (26.0% coverage, 69% calibration error) despite achieving marginally lower MSE (16% improvement). This performance gap stems from architectural differences in uncertainty modeling: ensemble methods capture only epistemic uncertainty through initialization variability, which collapses as models converge to similar solutions with sufficient training data. In contrast, UPN explicitly models both aleatoric uncertainty (observation noise via $\Sigma_0 = \sigma_{obs}^2 I$) and process stochasticity (via learned $Q(\mu, t)$), propagating uncertainty analytically through the coupled ODE $\frac{d\Sigma}{dt} = J\Sigma + \Sigma J^T + Q$. This principled approach ensures uncertainty grows appropriately with prediction horizon, reflecting fundamental limits on predictability rather than merely model uncertainty. The 16% MSE trade-off is both expected and acceptable: UPN optimizes the full predictive distribution (NLL) rather than point estimates alone, naturally regularizing against overfitting to training data. In high-stakes applications, autonomous systems, safety-critical control, medical decision-making, well-calibrated uncertainty estimates are substantially more valuable than marginal improvements in point accuracy, as overconfident predictors lead to inappropriate actions under uncertainty.

The success of single-point Markovian initialization validates that for the tested dynamical systems, the current state contains sufficient information for future prediction without requiring observation history. This has important practical implications: single-point initialization enables real-time deployment with minimal memory overhead, immediate prediction upon receiving a measurement, and clean separation between observation noise ($\Sigma_0$) and process dynamics ($f_\theta, Q_\phi$). System-specific performance patterns provide additional insights: the Linear System achieved essentially perfect calibration (95.3%, 0.3% error), demonstrating UPN's effectiveness on well-behaved linear dynamics. The Van der Pol oscillator exhibited asymmetric uncertainty around the limit cycle, correctly capturing phase-dependent sensitivity in stiff nonlinear systems. The Damped Pendulum showed robust performance on physically realistic dynamics with nonlinear restoring forces.

### *5.2. Chaotic Systems*

The Lorenz system presents a fundamental challenge for uncertainty quantification: exponential sensitivity to initial conditions renders long-term point predictions inherently unreliable, yet reliable confidence bounds remain essential for understanding predictability horizons. This section evaluates UPN's performance on the canonical Lorenz-63 attractor.

The Lorenz system is governed by:

$$\begin{aligned} \frac{dx}{dt} &= \sigma(y - x) \\ \frac{dy}{dt} &= x(\rho - z) - y \\ \frac{dz}{dt} &= xy - \beta z \end{aligned} \tag{32}$$

with standard parameters $\sigma = 10, \rho = 28, \beta = 8/3$, yielding chaotic dynamics with largest Lyapunov exponent $\lambda \approx 0.90$. The system was integrated using the RK45 solver with time step $dt = 0.01$, generating 100 trajectories of 15 seconds each. Observation noise ($\sigma_{obs} = 0.1$) was added to simulate realistic measurements. Data was partitioned into 70% training, 15% validation, and 15% test sets.

Although the Lorenz system is deterministic and Markovian-meaning the full state $(x, y, z)$ at time $t$ uniquely determines all future evolution-uncertainty arises from two sources: (1) measurement noise in the observed initial condition, and (2) phase ambiguity regarding position on the attractor. The initial covariance is set as:

$$\Sigma_0 = \text{diag}(\sigma_{obs}^2 + \alpha \cdot \text{Var}[\text{x}_{\text{attractor}}]) \quad (33)$$

where $\alpha = 0.01$ incorporates small phase ambiguity relative to attractor variance, balancing observation noise with local uncertainty structure. This approach demonstrates UPN's capability to propagate uncertainty from minimal initial information without requiring historical context.

*5.2.1. Results and Analysis*

Models were trained to predict 50 future time steps (0.5 seconds) from single observations. UPN was trained for 150 epochs with learning rate $5 \times 10^{-4}$ and early stopping (patience=15). Figure 17 shows the training progression for all methods, demonstrating clean convergence without overfitting as training and validation losses track closely throughout. To rigorously test whether epistemic uncertainty alone can capture chaotic unpredictability, a 30-model ensemble was trained (compared to 10 models in non-chaotic experiments), with each member trained independently for 150 epochs.

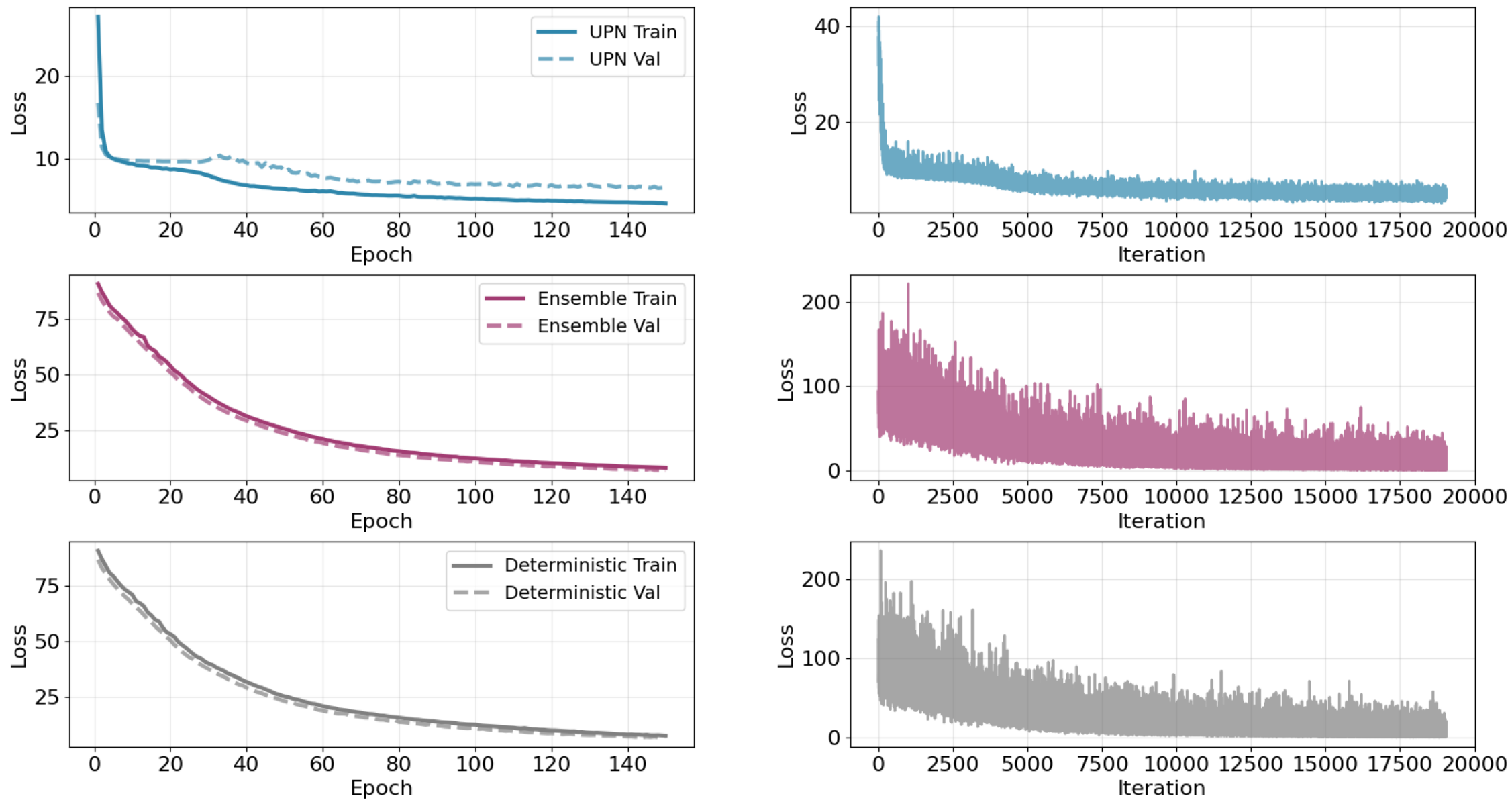

Figure 17. Training Progress for All Methods. Left column shows loss per epoch for UPN (top), Ensemble (middle), and Deterministic (bottom) methods, with solid lines representing training loss and dashed lines representing validation loss. Right column shows loss per iteration with the same ordering.

Table 4 summarizes aggregate performance across all test trajectories. UPN achieves 94.5% coverage, nearly perfect calibration, with aggregate MSE of 29.9, whereas the 30-model ensemble achieves only 66.8% coverage despite MSE of 7.2. This 4.1× MSE ratio must be interpreted in context: in chaotic systems, low MSE paired with poor calibration indicates dangerous overconfidence, as the model underestimates fundamental unpredictability.

Table 4. Performance comparison on Lorenz-63 chaotic attractor.[‡]

| Metric | UPN | Ensemble (30) | Deterministic | Ratio (UPN/Ens) |
|---|---|---|---|---|
| Aggregate MSE | 29.9 | **7.2** | 7.3 | 4.1× |
| 95% CI Coverage | **94.5%** | 66.8% | N/A | 1.41× |
| NLL | **6.26** | N/A | N/A | — |
| Inference Time (ms) | **186** | 2632 | 63 | 0.07× |

‡ Bold indicates best performance per metric.

| | | | | |
|---|---|---|---|---|
| Training Time (hours) | **9.2** | 26.5 | 1.5 | 0.35× |
| Forward Passes | 1 | 30 | 1 | — |

Figure 18 visualizes this trade-off starkly: ensemble's low MSE comes at the cost of capturing only two-thirds of ground truth outcomes, rendering its predictions unreliable for risk assessment. UPN's higher MSE reflects conservative uncertainty estimates that properly account for exponential error growth in chaotic dynamics. Figure 19 demonstrates that all methods successfully learn the characteristic butterfly-wing structure of the Lorenz attractor in both 3D space and phase plane projections, with predictions following ground truth trajectories through the attractor's complex geometry. Figure 20 provides detailed state-wise analysis showing that UPN's uncertainty bands (blue shaded regions) appropriately contain the oscillating ground truth across all three dimensions, whereas ensemble's narrower confidence intervals frequently miss true values despite tighter point predictions.

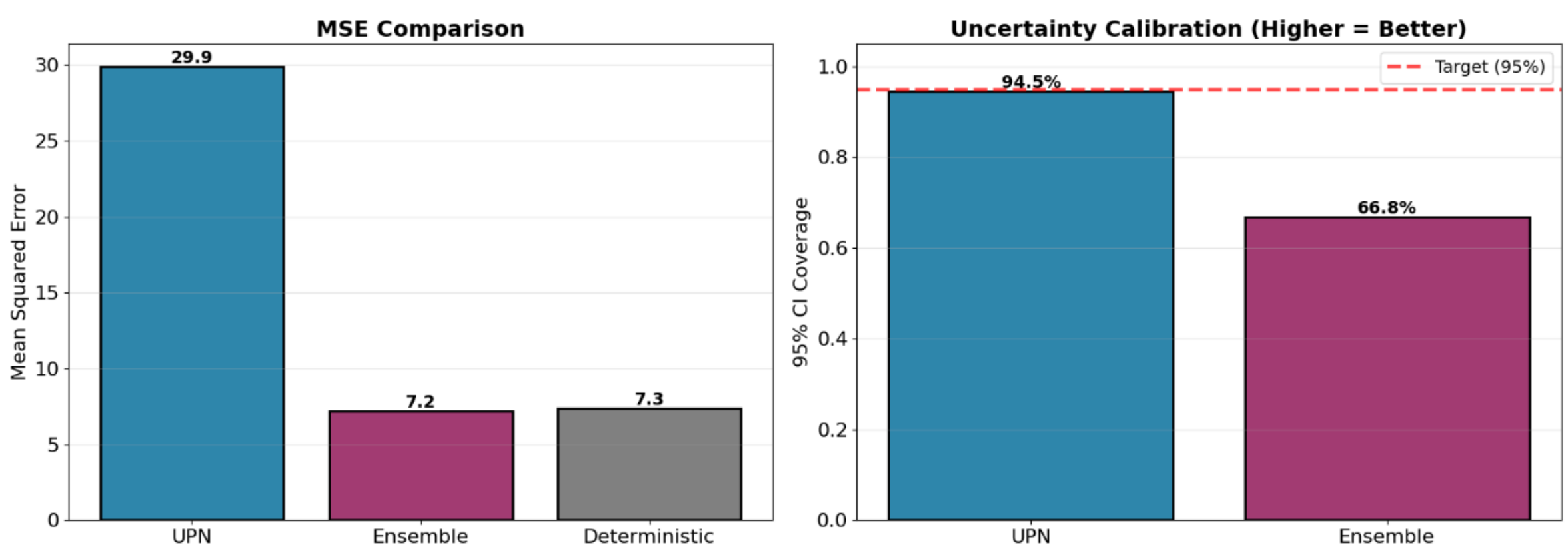

Figure 18. Performance Metrics Comparison. Left panel compares aggregate MSE across methods. Right panel compares 95% confidence interval coverage for methods with uncertainty quantification.

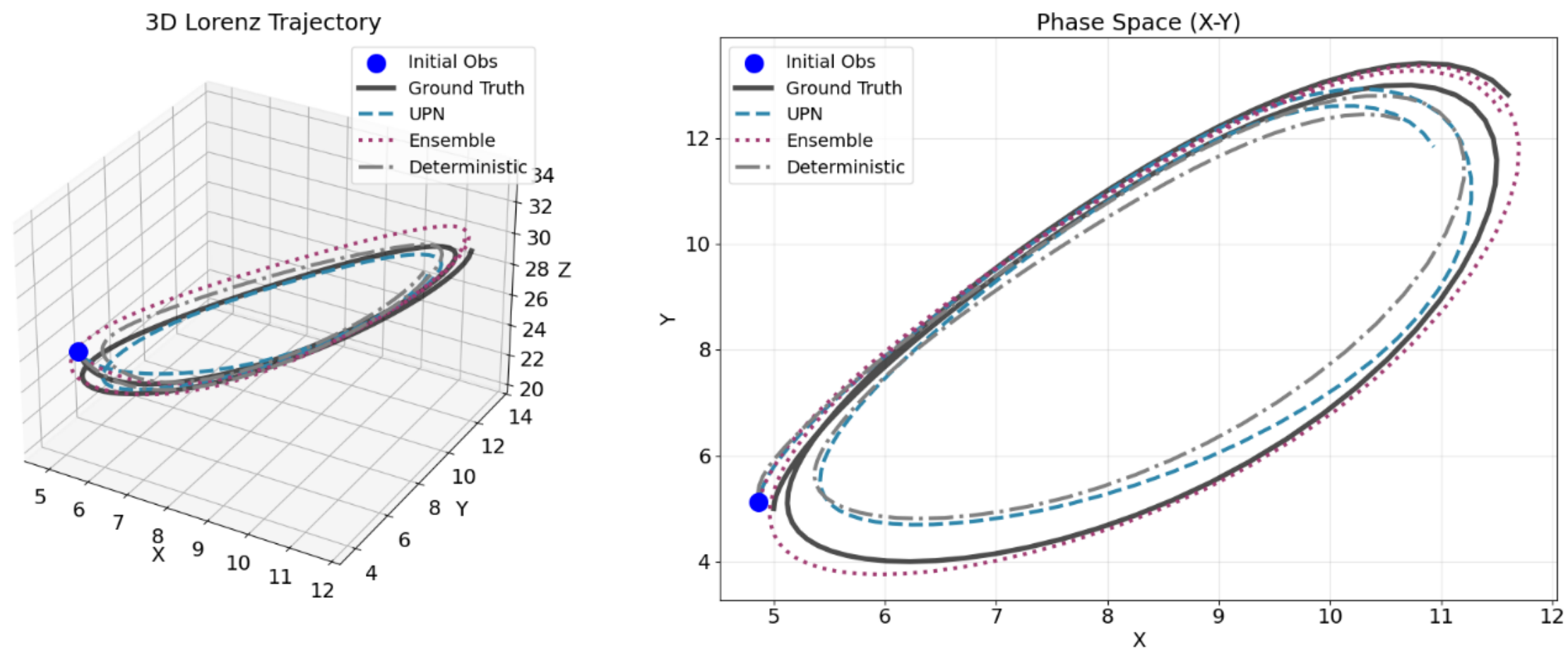

Figure 19. Lorenz Attractor Trajectories in 3D and Phase Space.

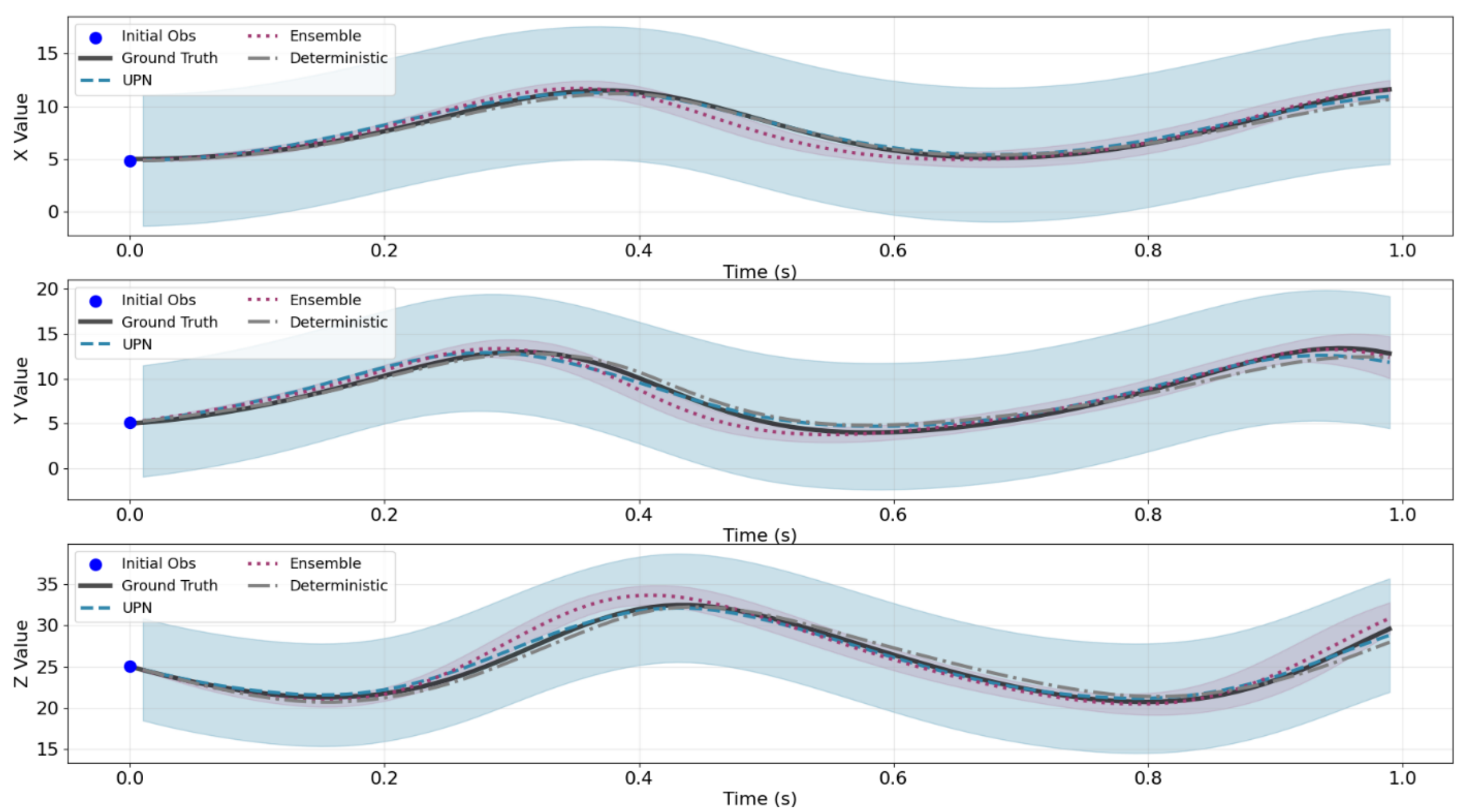

Figure 20. State-wise Predictions with Uncertainty Quantification.

To understand where UPN's MSE penalty arises, predictions were stratified by forecast horizon (Figure 21). Short-term predictions ($t < 0.3$ s) occur before chaos significantly affects trajectories, while long-term predictions ($t > 0.7$ s) lie deep in the chaotic regime beyond approximately one Lyapunov time ($T_\lambda \approx 1/0.9 \approx 1.1$ seconds).

In the short-term predictable regime, UPN achieves MSE of 0.74 with coverage of 98.3%, while ensemble achieves MSE of 0.39 with coverage of only 60.5%, yielding an MSE ratio of 1.88 ×. UPN remains competitive in point prediction accuracy (within 2 × of ensemble) while maintaining near-perfect calibration. Ensemble's 60.5% coverage indicates overconfidence even in this regime, as epistemic uncertainty alone underestimates measurement and process noise.

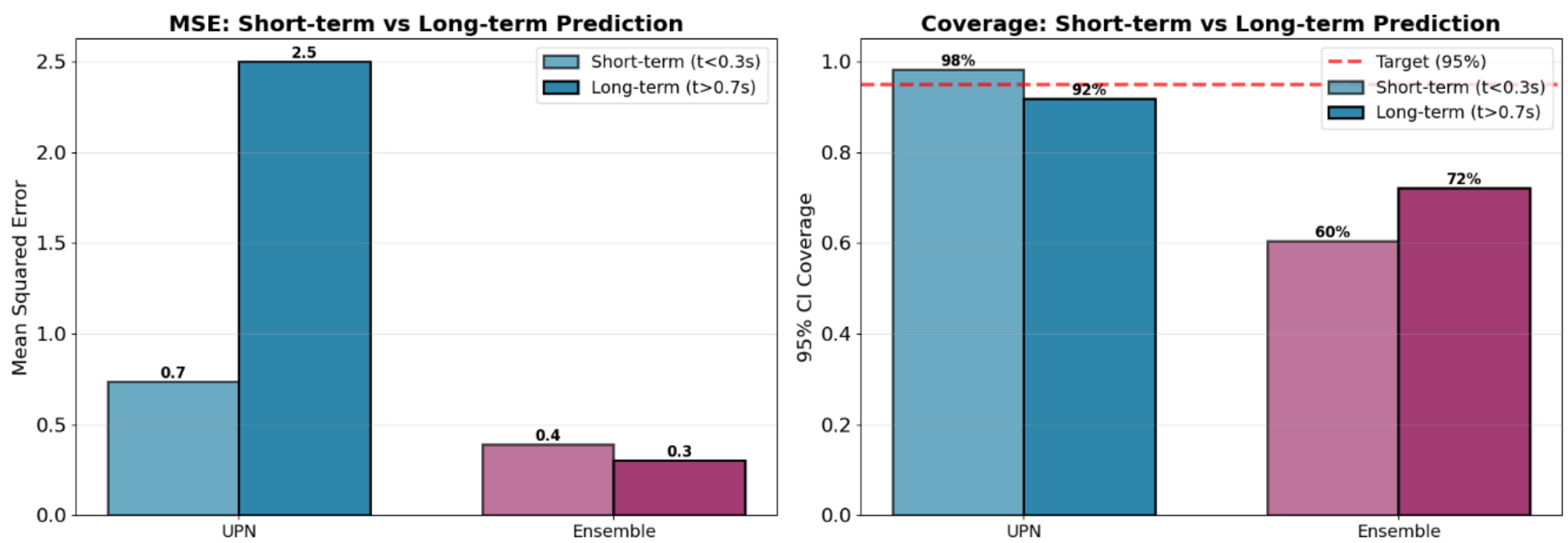

Figure 21. Horizon-Stratified Performance Analysis. Left panel compares MSE for short-term ($t < 0.3$s, light bars) vs long-term ($t > 0.7$s, dark bars) prediction horizons. Right panel compares coverage for the same horizons with target 95% marked by red dashed line.

In the long-term chaotic regime, UPN achieves MSE of 2.50 with coverage of 91.9%, while ensemble achieves MSE of 0.30 with coverage of only 72.2%, yielding an MSE ratio of 8.3×. The MSE gap widens dramatically from 1.88× to 8.3× as chaos dominates, yet UPN maintains 91.9% calibration while ensemble's coverage degrades to 72.2%. This divergence reveals the fundamental limitation of epistemic uncertainty: model disagreement grows more slowly than true chaotic uncertainty, leading ensemble to systematically underestimate confidence intervals at longer horizons. UPN's higher MSE in this regime is not a modeling failure but rather an honest acknowledgment that precise point predictions are impossible, the mean prediction becomes more conservative while uncertainty bounds expand appropriately.

To validate that UPN's uncertainty growth aligns with chaotic dynamics theory, MSE was computed at multiple prediction horizons and fit to the expected exponential form: $\mathrm{MSE(t)} \approx \mathrm{MSE}_0 \cdot \exp{(2\lambda t)}$, where $\lambda$ is the effective Lyapunov exponent. Figure 22 presents this analysis on both linear and logarithmic scales.

The estimated growth rates reveal fundamental differences in how methods model chaotic error propagation. UPN exhibits an estimated exponent of $\hat{\lambda} = 1.29$ compared to the theoretical value of $\lambda = 0.90$, indicating 43% conservative bias in uncertainty propagation. Ensemble exhibits an estimated exponent of $\hat{\lambda} = 0.31$ compared to the theoretical value, severely underestimating chaos by 66%. Deterministic methods yield $\hat{\lambda} = 0.44$, also underestimating by 51%.

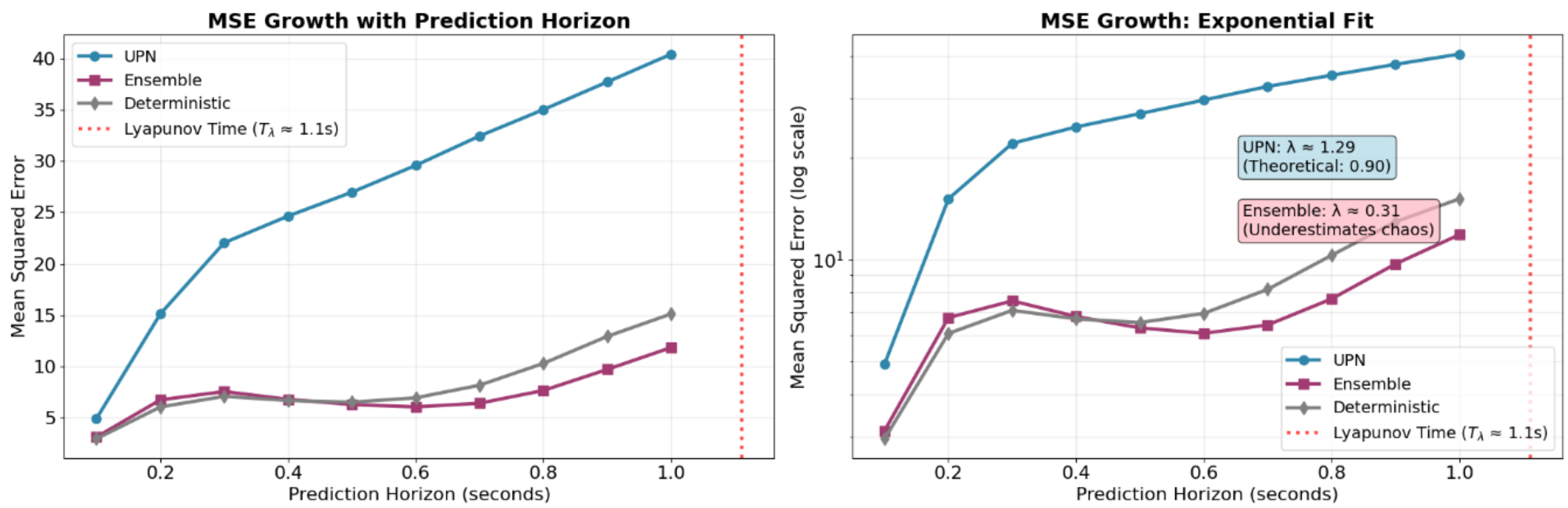

Figure 22. MSE Growth vs Lyapunov Time Analysis. Left panel shows MSE growth on linear scale as a function of prediction horizon for UPN (blue circles), Ensemble (purple squares), Deterministic (gray diamonds), with Lyapunov time marked at approximately 1.1 seconds (red dotted line). Right panel shows the same data on logarithmic scale with fitted exponential growth rates annotated

UPN's over-estimation is preferable to under-estimation in safety-critical applications: UPN errs on the side of caution, ensuring confidence bounds remain reliable even when uncertainty grows faster than theoretically predicted due to model misspecification or unmodeled dynamics. In contrast, ensemble's $\hat{\lambda} = 0.31$ severely underestimates chaos, explaining its poor long-term coverage. The log-scale plot (Figure 22, right panel) shows ensemble's MSE growth curve is nearly flat compared to the theoretically predicted exponential trajectory, demonstrating that epistemic uncertainty from 30 models cannot capture aleatoric uncertainty from chaotic divergence. Increasing ensemble size further would worsen computational cost with diminishing returns on calibration, as the fundamental issue-confusing model disagreement with chaotic unpredictability remains unresolved.

Deterministic methods exhibit similar underestimation with $\hat{\lambda} = 0.44$, achieving low MSE

through overly confident point predictions that ignore uncertainty entirely. The convergence of ensemble and deterministic MSE values (7.2 vs 7.3) suggests ensemble provides little advantage over deterministic prediction for point accuracy in this regime, despite requiring $30\times$ computational cost.

Figure 23 tracks the trace of the covariance matrix $\text{Tr}(\Sigma)$ over time, quantifying total uncertainty on both linear and logarithmic scales. UPN exhibits exponential growth from $\text{Tr}(\Sigma) \approx 0.5$ at $t = 0$ to approximately 12 at $t = 2$ s, consistent with chaotic error propagation. The log-scale plot confirms near-linear growth on semi-log axes, validating exponential dynamics. In contrast, ensemble uncertainty remains nearly flat with $\text{Tr}(\Sigma)$ oscillating between approximately 1 and 7, growing sub-linearly despite chaotic dynamics. This stark difference visualizes why ensemble achieves low MSE: by severely underestimating uncertainty growth, the ensemble maintains tight confidence intervals that frequently miss ground truth.

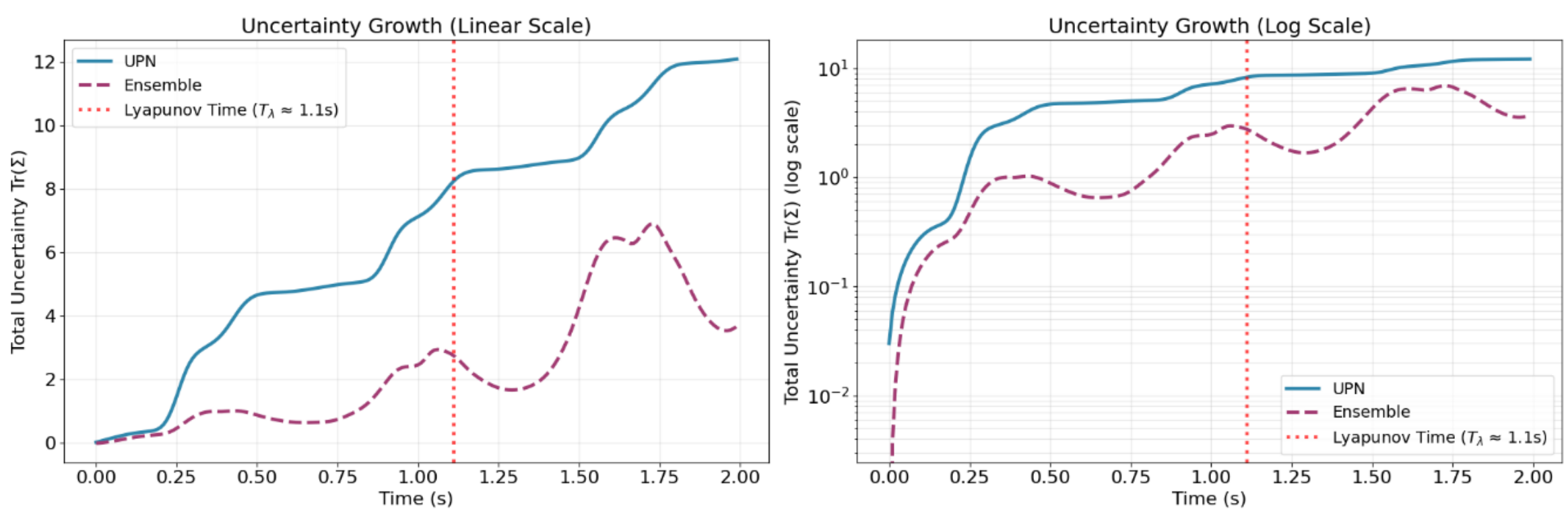

Figure 23. Uncertainty Evolution in Chaotic Dynamics. Left panel shows total uncertainty on linear scale for UPN (blue solid) and Ensemble (purple dashed) over 2.0 seconds, with Lyapunov time marked at approximately 1.1 seconds (red dotted line). Right panel shows the same data on logarithmic scale.

Unlike ensemble methods that produce a single distribution from multiple model runs, UPN learns a time-evolving Gaussian distribution $\mathcal{N}(\mu(t), \Sigma(t))$ that can be sampled to generate multiple plausible future trajectories. Figure 24 shows 20 sampled trajectories from UPN's learned distribution, revealing that samples realistically cover the Lorenz attractor structure in both 3D space and phase plane projections. The ground truth trajectory lies well within this distribution, and individual samples exhibit characteristic butterfly-wing patterns. This qualitative validation confirms UPN learns not just point predictions but a probabilistically meaningful distribution over future states, enabling risk assessment through Monte Carlo sampling when downstream applications require trajectory distributions rather than confidence intervals.

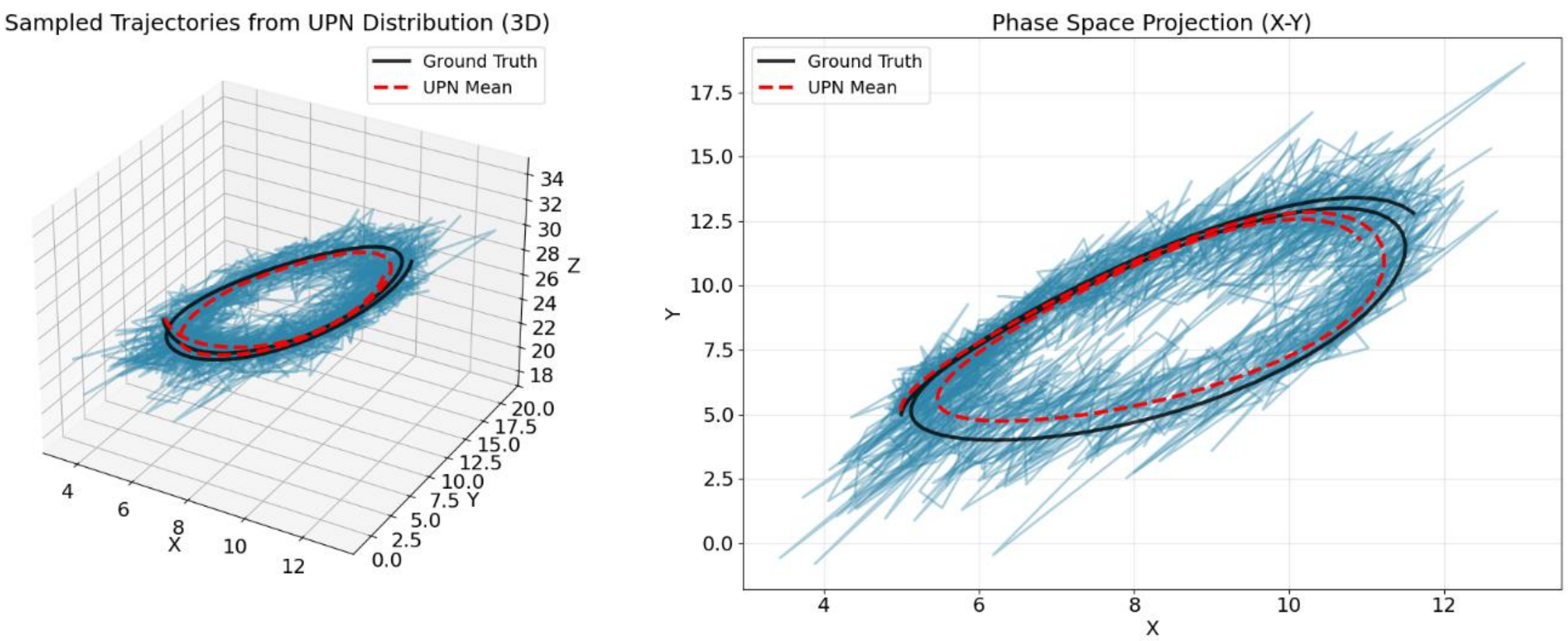

Figure 24. Sampled Trajectories from UPN's Learned Distribution. Left panel shows 20 trajectories sampled from UPN's learned distribution in 3D space (light blue), with ground truth (black solid) and UPN mean (red dashed). Right panel shows the same in X-Y phase space projection.

Figure 25 quantifies computational costs across multiple dimensions. Despite UPN's additional covariance propagation, inference requires only 186 ms per batch, $14\times$ faster than the 30-model ensemble (2632 ms) and only $3\times$ slower than deterministic prediction (63 ms). Training time follows similar trends: UPN trains in 9.2 hours compared to ensemble's 26.5 hours, as each ensemble member must be trained independently. These efficiency gains become increasingly important as ensemble size grows: achieving better calibration through larger ensembles exacerbates computational costs linearly, while UPN's single forward pass scales independently of uncertainty quality.

Memory footprint remains comparable with UPN requiring 0.04 MB, ensemble requiring 0.5 MB, and deterministic requiring 0.04 MB. UPN stores a single neural ODE plus process noise network, while ensemble stores 30 separate neural ODEs. For deployment in resource-constrained or real-time scenarios, UPN's $14\times$ inference speedup enables uncertainty-aware prediction where large ensembles would be computationally prohibitive.

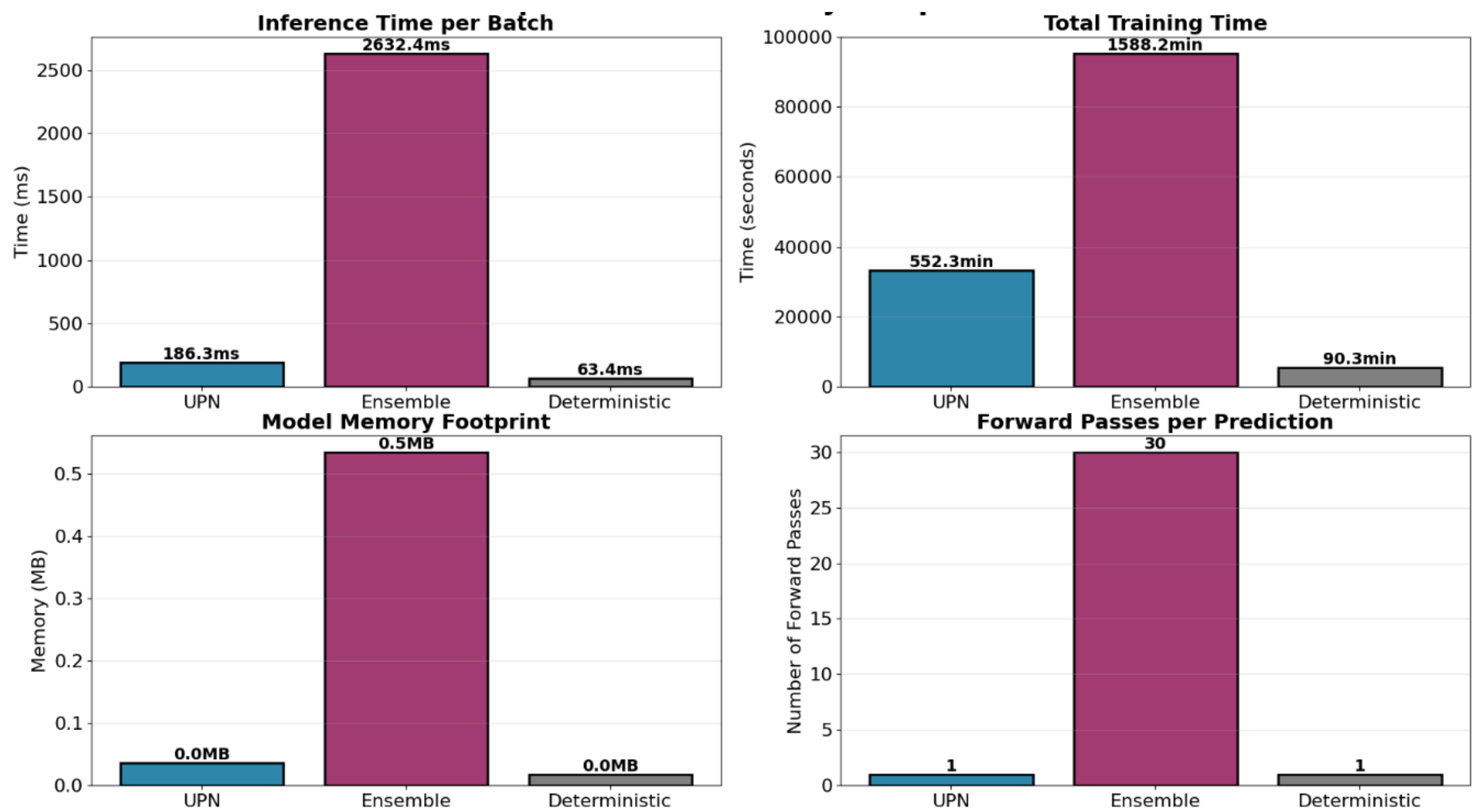

Figure 25. Computational Efficiency Comparison. Four panels quantify: (top-left) inference time per batch in milliseconds; (top-right) total training time in hours; (bottom-left) model memory footprint in megabytes; (bottom-right) number of forward passes per prediction. UPN achieves superior calibration with significantly lower computational cost than large ensembles.

*5.2.2. Discussion*

The Lorenz results reveal a fundamental principle for uncertainty quantification in chaotic systems: calibration must take precedence over point accuracy when predictability horizons are finite. UPN's $4.1\times$ higher aggregate MSE compared to ensemble is not a weakness but rather evidence of honest uncertainty quantification. Three key findings support this interpretation. First, stratified analysis shows UPN is competitive in the predictable regime with MSE ratio of $1.88\times$ but conservative in the chaotic regime with MSE ratio of $8.3\times$, indicating the MSE penalty arises specifically where chaos dominates and precise prediction becomes impossible. Second, Lyapunov analysis demonstrates ensemble severely underestimates exponential error growth with $\hat{\lambda} = 0.31$ compared to the theoretical $\lambda = 0.90$, achieving low MSE through dangerous overconfidence rather than superior modeling. Third, coverage degradation shows ensemble's calibration deteriorates from 60.5% short-term to 72.2% long-term, never approaching the target 95%, whereas UPN maintains 98.3% short-term and 91.9% long-term calibration-consistently reliable across prediction horizons.

For practitioners, these results suggest UPN is particularly valuable when: (i) safety-critical applications require reliable confidence bounds, (ii) identifying predictability limits is as important as point prediction, or (iii) computational constraints preclude large ensembles (embedded systems, real-time control). Conversely, when point accuracy is paramount and uncertainty can be sacrificed, such as trajectory planning with generous safety margins, ensemble or deterministic methods may suffice despite poor calibration.

The Markovian initialization strategy proves effective for fully observed chaotic systems, eliminating the need for observation histories while maintaining excellent calibration. This approach leverages the theoretical sufficiency of single-state initialization when complete system state is accessible, with modest phase ambiguity modeling to account for attractor-scale uncertainty. For partially observed systems or those with hidden states, incorporating short observation histories (5-10 steps) could further improve point predictions while retaining UPN's principled uncertainty propagation, a promising direction for future work.

Ultimately, the Lorenz experiments validate UPN's core thesis: in chaotic regimes, trading point accuracy for calibrated uncertainty is not a compromise but the correct quantification of what is knowable versus fundamentally unpredictable.

### *5.3. CubeSat State Estimation*

To evaluate UPN's performance on real-world data with measurement noise and model uncertainties, an experimental testbed simulating planar free-floating dynamics in microgravity-like conditions was developed. The system consists of a 4m × 2m precision granite table providing a low-friction surface (Figure 26). A CubeSat spacecraft simulator supported by three symmetrically placed air bearings enables near-frictionless motion with three degrees of freedom: $x$-position, $y$-position, and yaw angle ($\theta$). Position and orientation tracking uses a pseudogalactic star tracking system with overhead infrared LEDs and onboard cameras.

The experimental data confirmed key characteristics of free-floating dynamics: after initial impulse disturbances, acceleration values approach near-zero across all degrees of freedom, velocity profiles stabilize to nearly constant values, and position-time relationships show predominantly linear trajectories. Over 70% of trajectories exhibit strong linearity ($R^2 > 0.9$), confirming momentum conservation principles essential for autonomous space operations.

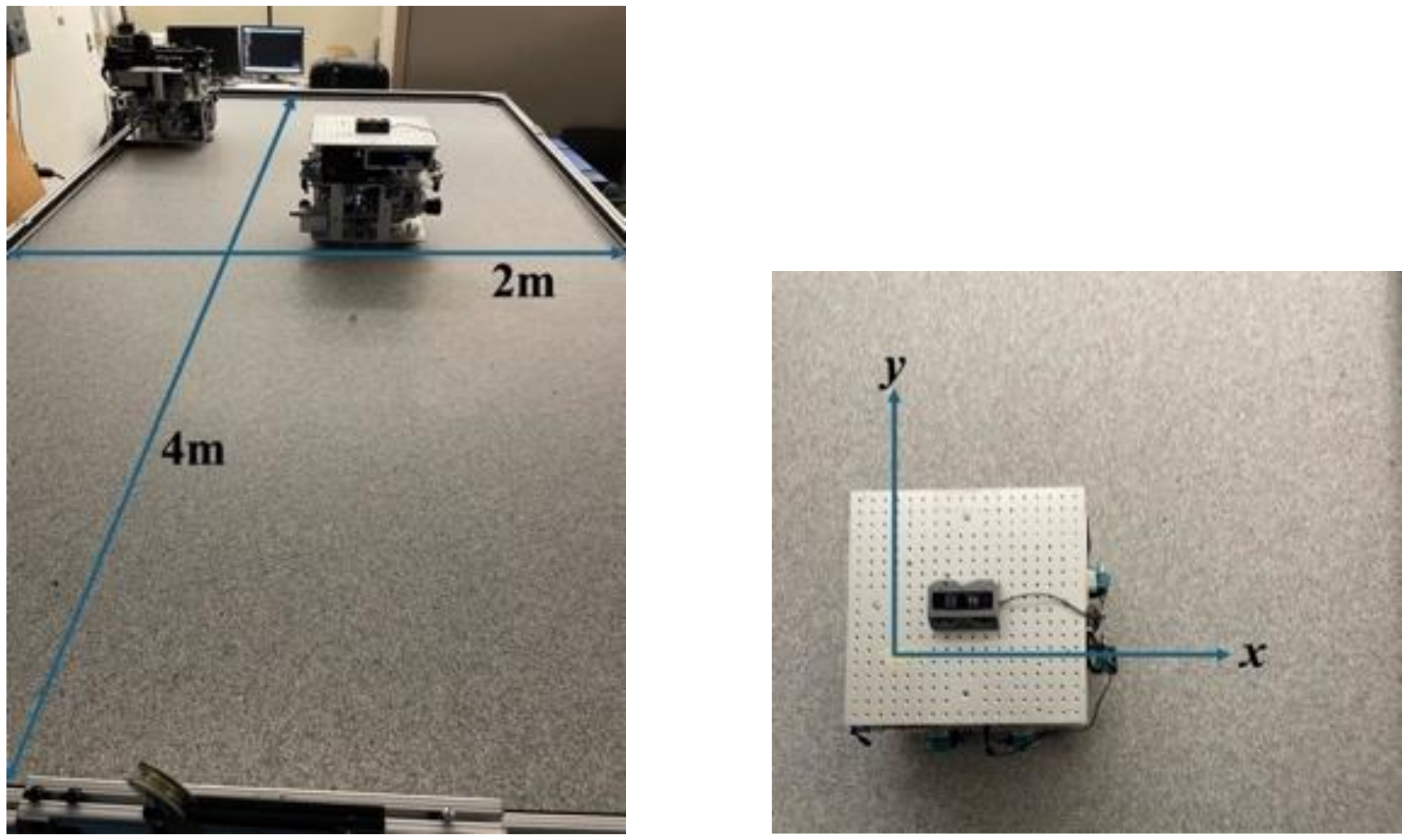

Figure 26. Experimental setup of planar free-floating dynamics simulator.

*5.3.1. Experimental Setup and State Representation*

The dataset comprises 30 trajectories collected from the CubeSat simulator, each representing a free-drift experiment where the spacecraft received an initial impulse and evolved under natural dynamics. Data recording occurred at approximately 5.5 Hz over 10-second intervals per trajectory. The measurements exhibit realistic sensor characteristics with noise levels varying across trajectories. Position measurements show noise standard deviations averaging 1.9 cm for $x$-position and 0.8 cm for $y$-position, while attitude sensors demonstrate approximately 2.9° standard deviation in yaw measurements, though these values exhibit significant variation between trajectories. The three degrees of freedom demonstrate natural coupling through the spacecraft's inertia tensor, while air bearing friction and residual magnetic torques introduce subtle nonlinearities into the system dynamics.

For training and evaluation, the dataset was divided into 21 training trajectories (70%), 4 validation trajectories (15%), and 5 test trajectories (15%). The data was processed using sliding windows of 30 time steps, corresponding to approximately 5.4 seconds total, with the first 10 steps (1.8 seconds) serving as historical context and the subsequent 20 steps (3.6 seconds) as prediction targets. This configuration results in 120 test windows from 5 distinct trajectories, providing diverse evaluation scenarios including smooth linear motion and more challenging maneuvers with rotational coupling.

The sensor measurements directly provide only position and orientation $s_{obs} = [x, y, \theta]$. However, free-floating spacecraft dynamics are governed by second-order differential equations where the state evolution depends on both position and velocity. Predicting from position alone violates the Markov property, as the system's future depends on information (velocity) not contained in the current observation. To construct a complete Markovian state, velocity information is extracted from the observation history using finite differences. Given the last two historical observations $s_{t-1}$ and $s_t$ at times $t-1$ and $t$, the velocity components are estimated as $v = \frac{s_t - s_{t-1}}{\Delta t}$ where $\Delta t \approx 0.18s$ is the sampling interval. This yields the complete 6-dimensional Markovian state $z = [x, y, \theta, v_x, v_y, \omega]$ comprising both configuration (position/orientation) and generalized velocities. This state representation satisfies the Markov property: given $z(t)$, the system's evolution is fully determined by the dynamics without requiring additional historical information.

*5.3.2. Comparison Methodology*

Four uncertainty-aware neural methods were evaluated alongside a constant-velocity baseline model. Constant-velocity baseline extrapolates future states using finite-difference velocity estimates from the last two historical observations, providing a simple physics-based comparison point. In UPN, the state representation includes both position and velocity components $[x, y, \theta, v_x, v_y, \omega]$, forming a 6-dimensional Markovian state. The dynamics network $f_\theta$ and process noise network $Q_\phi$ are implemented as 2-layer MLPs with 32 hidden units and tanh activations. The integrated Kalman update mechanism (Section 3.4) enables principled incorporation of sparse measurements.

Latent ODE [1] provides a strong Neural ODE baseline that learns dynamics in a latent space. The encoder uses a 2-layer GRU with 32 hidden units to map observation sequences to latent states. The latent dynamics are modeled by a 2-layer MLP (32 hidden units), and the decoder produces both mean and variance estimates for the observed states. This architecture captures temporal dependencies through the RNN encoder while maintaining computational efficiency.

Ensemble Neural ODE [3] consists of five independently trained models, each implementing a 2-layer MLP with 32 hidden units for dynamics modeling. Uncertainty is quantified through ensemble variance, representing model uncertainty arising from different random initializations and training trajectories. Each ensemble member operates on the full 6-dimensional state space.

Neural SDE [8] uses stochastic differential equations with diagonal noise. The drift network and diffusion network are implemented as 2-layer MLPs with 32 hidden units. Uncertainty estimation requires generating 100 trajectory samples via Euler-Maruyama integration with $\Delta t = 0.01$ s, providing aleatoric uncertainty through the stochastic dynamics.

All neural methods employed identical network architectures for their dynamics components and were trained for 100 epochs using the Adam optimizer [44] with learning rate $5\times10^{-4}$ and weight decay $1\times10^{-4}$. Early stopping with patience of 15 epochs was applied based on validation performance.

The evaluation considered two distinct operational scenarios. Scenario 1, without measurement updates, tests pure predictive capability where models forecast future trajectories without receiving any observations during the 3.6-second prediction horizon. This evaluates each model's ability to learn and propagate system dynamics and uncertainty from historical context alone, providing a fair baseline comparison across all methods. Scenario 2, with sparse measurement updates, simulates realistic operational conditions where observations become available every 5 time steps (approximately 0.9 seconds) during the prediction horizon. This scenario is evaluated exclusively for UPN, which possesses an architecturally integrated Kalman update mechanism that properly updates both state estimates and uncertainty covariances according to optimal filtering theory. The Kalman framework provides a principled approach to incorporating measurements: the innovation term corrects the predicted state, while the Kalman gain optimally weights this correction based on the relative uncertainties in the prediction and measurement. Crucially, the update propagates through both the mean and covariance, maintaining consistent uncertainty quantification.

The other methods lack such integrated update mechanisms. While ad-hoc update schemes could theoretically be implemented, such as simple blending (predictions adjusted toward observations using fixed gains), Ensemble Kalman Filters [45], or particle filters [46], these approaches face fundamental limitations. Simple blending modifies predicted values without updating the underlying model state or properly adjusting uncertainty representations, causing corrections to decay rapidly. More sophisticated methods like Ensemble Kalman Filters require significant architectural modifications and substantially increased computational overhead, fundamentally changing the methods' characteristics. Latent ODE faces an architectural constraint as its RNN-based encoder requires the complete observation sequence, making mid-sequence updates conceptually incompatible with the model structure. Given these challenges and to maintain fair comparison, Scenario 2 focuses on demonstrating UPN's unique capability for principled measurement integration rather than comparing methods with incompatible or ad-hoc update approaches.

Performance is evaluated using two metrics: (1) MSE measuring prediction accuracy across all state dimensions, and (2) 95% Coverage Rate measuring uncertainty calibration by computing the fraction of ground truth states falling within predicted 95% confidence intervals.

### *5.3.3. Results and Analysis*

Table 5 presents the quantitative performance metrics for all methods across both evaluation scenarios. The results reveal distinct patterns in predictive accuracy and uncertainty calibration between pure prediction and measurement update conditions.

In the pure prediction scenario without measurement updates, Ensemble Neural ODE achieved the lowest MSE of 0.031, followed closely by Neural SDE at 0.035. UPN demonstrated higher error at 0.061, while Latent ODE showed the highest error at 0.199. This performance ordering reflects the additional complexity of learning coupled dynamics for both mean and covariance simultaneously in the UPN framework. The constant-velocity baseline produced an MSE of 1.395, which all neural methods substantially outperformed, validating the value of learned dynamics models.

Figures 27-30 illustrate trajectory predictions without measurement updates for each method. The visualizations show predictions (colored lines) with 95% confidence intervals (shaded regions) compared to ground truth (black lines) across four diverse test trajectories spanning different motion profiles.

The uncertainty calibration analysis reveals critical differences between methods. UPN achieved 81.0% coverage for its 95% confidence intervals, approaching the theoretical ideal and indicating slight underconfidence. In stark contrast, the Ensemble method produced severely underconfident intervals with only 57.9% coverage, while Neural SDE demonstrated 52.6% coverage. Latent ODE achieved the best calibration at 92.2% coverage (closest to the 95% ideal), though this came at the cost of the highest prediction error. The narrow confidence bounds of Ensemble and Neural SDE methods are particularly concerning for safety-critical applications, as they suggest false certainty about predictions that may be inaccurate.

This reveals a fundamental trade-off: Ensemble and Neural SDE achieve excellent prediction accuracy by focusing learning capacity on the mean dynamics, but fail to properly quantify uncertainty. Methods that explicitly model uncertainty (Latent ODE, UPN) sacrifice some predictive accuracy but provide more reliable confidence estimates. Notably, UPN achieves the best balance with 81% coverage while maintaining competitive MSE performance.

The introduction of sparse measurement updates (every 0.9 seconds) fundamentally altered the performance characteristics, as shown in Table 5 and Figure 31. UPN achieved an 89.6% reduction in MSE when incorporating updates, improving from 0.061 to 0.006, outperforming all other methods including those with better pure prediction accuracy. The coverage rate simultaneously improved to 96.8%, nearly matching the theoretical 95% confidence level and demonstrating near-optimal calibration.

Figure 31 demonstrates that UPN's predictions align with observations at update points and maintain improved accuracy during the prediction intervals between updates. The uncertainty bounds exhibit theoretically correct behavior: contracting when measurements arrive (information gain) and expanding during pure prediction periods (information decay). This dynamic uncertainty representation is crucial for autonomous systems that must reason about confidence levels when making decisions between measurement opportunities.

Table 5. CubeSat Trajectory Prediction Performance.

| Metric | Scenario | UPN | Latent ODE | Ensemble | Neural SDE | Baseline |
|---|---|---|---|---|---|---|
| MSE | Without Updates | 0.061 | 0.199 | 0.031 | 0.035 | 1.395 |
| | With Updates | 0.006 | — | — | — | 1.395 |
| Coverage (95% CI) | Without Updates | 81.0% | 92.2% | 57.9% | 52.6% | — |
| | With Updates | 96.8% | — | — | — | — |
| MSE Reduction | From Updates | 89.6% | — | — | — | — |

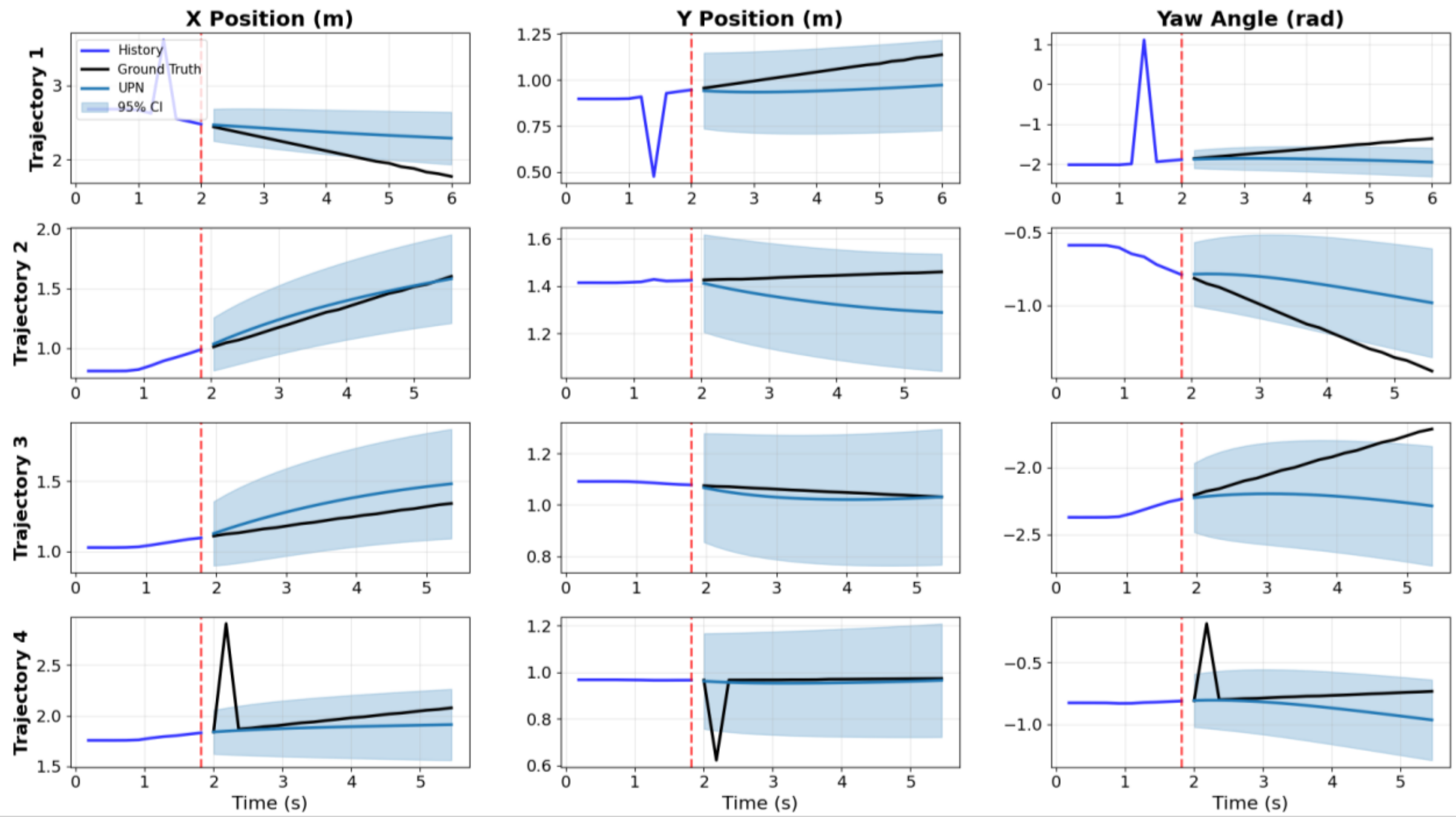

Figure 27. UPN trajectory predictions without measurement updates.

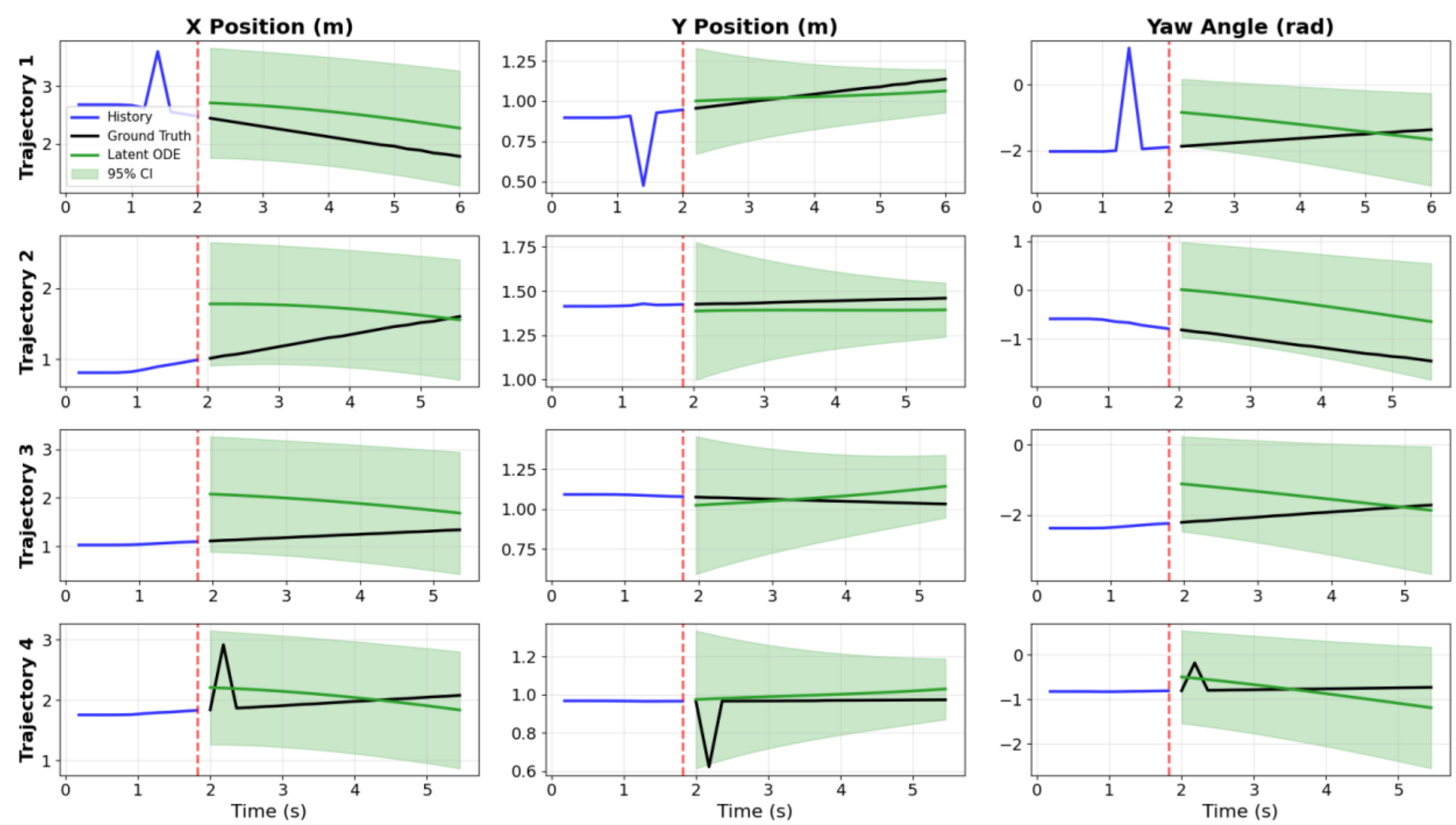

Figure 28. Latent ODE trajectory predictions without measurement updates.

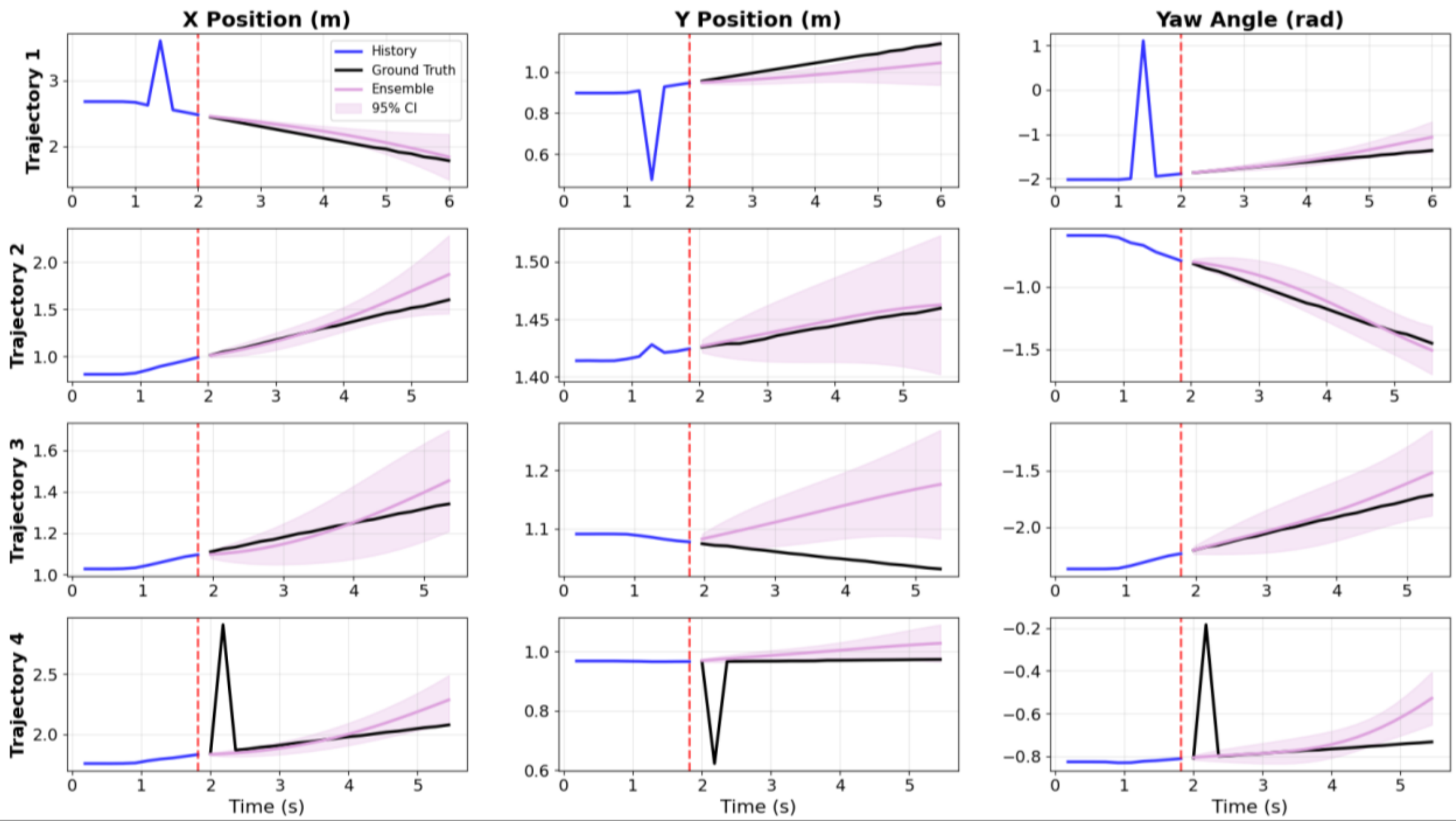

Figure 29. Ensemble Neural ODE trajectory predictions without measurement updates.

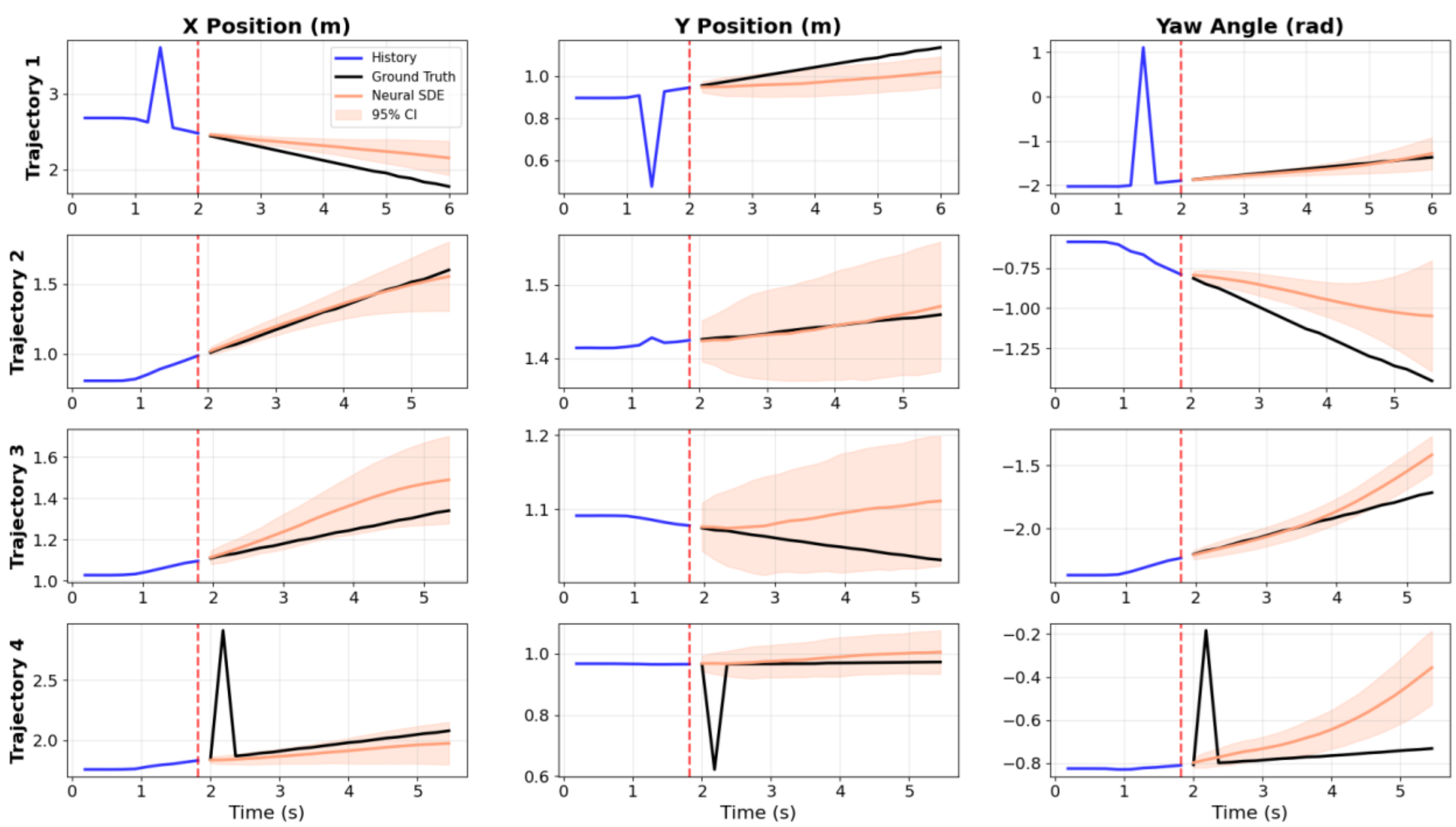

Figure 30. Neural SDE trajectory predictions without measurement updates.

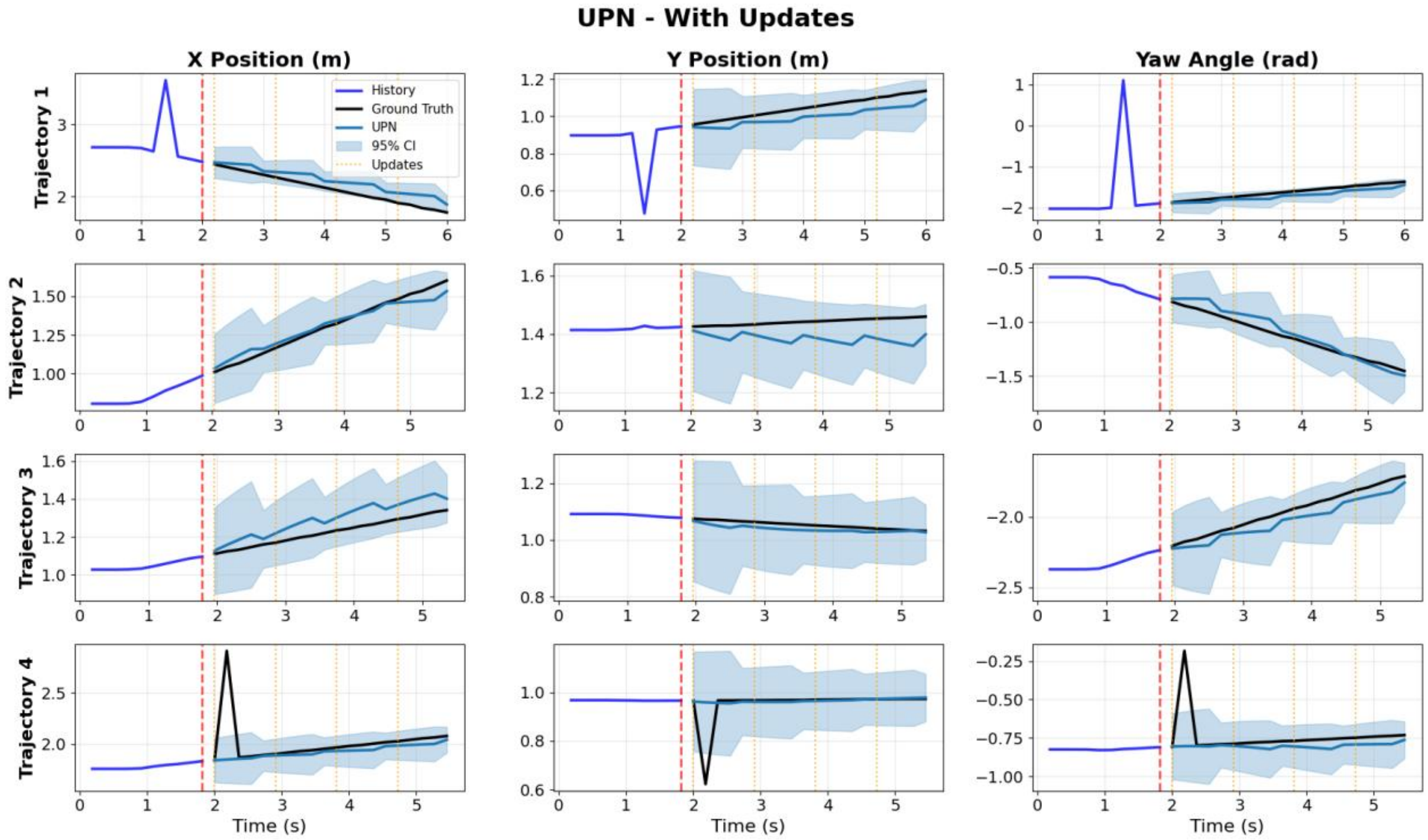

Figure 31. UPN trajectory predictions with sparse measurement updates.

*5.3.4. Discussion*

The CubeSat experiments demonstrate the practical advantages of architecturally integrated uncertainty quantification in dynamical system modeling. The 89.6% MSE improvement achieved by UPN with sparse updates, compared to modest 11-14% improvements for methods using ad-hoc blending, highlights the importance of principled uncertainty handling. The update mechanism in UPN properly propagates information through both state estimates and their associated covariances according to optimal filtering theory, while ad-hoc blending creates corrections that fail to persist through the prediction horizon.

The calibration results provide further evidence of UPN's robustness. While Latent ODE achieved the best pure prediction calibration (92.2%), it cannot incorporate updates and shows poor tracking accuracy (MSE = 0.199). Ensemble and Neural SDE methods, despite achieving the lowest prediction errors (0.031 and 0.035), provide severely miscalibrated uncertainty estimates (57.9% and 52.6% coverage) that remain poor even with updates. UPN maintained well-calibrated uncertainty across operational scenarios: 81.0% coverage without updates (slight underconfidence) improving to 96.8% with updates (near-optimal).

For spacecraft operations and similar applications, reliable uncertainty bounds prove as critical as accurate point predictions. A system with MSE = 0.031 but 57.9% coverage is unreliable, the narrow confidence intervals suggest false certainty about predictions that may deviate significantly from reality. In contrast, UPN provides confidence intervals that correctly reflect available information, enabling risk-aware autonomous decision-making. The results indicate that methods capable of meaningfully incorporating new information while maintaining calibrated uncertainty are particularly valuable for applications requiring continuous operation with intermittent observations, such as spacecraft attitude control, autonomous navigation, and industrial process monitoring.

The performance differences also illuminate fundamental trade-offs in neural dynamics modeling. Pure prediction accuracy (MSE) can be optimized by focusing all learning capacity on mean dynamics, as demonstrated by the Ensemble approach. However, this comes at the cost of uncertainty quantification, the ensemble variance severely underestimates true uncertainty because all models converge to similar solutions. UPN's coupled mean-covariance dynamics impose additional learning complexity, resulting in slightly higher pure prediction error, but enable principled uncertainty propagation that proves crucial when integrating measurements. This suggests that for real-world deployment, architectural choices should prioritize update mechanisms and calibration over pure prediction performance metrics.

### *5.4. Time-Series Forecasting*

To evaluate UPN's effectiveness in real-world time series prediction, experiments on the ETTh1 dataset [47] are conducted, a widely-used benchmark containing hourly electricity transformer temperature measurements. This dataset presents challenges typical of industrial sensor data: multiple correlated features, complex temporal dependencies, and inherent measurement noise.

#### *5.4.1. Experimental Setup*

The ETTh1 dataset comprises 17,420 hourly observations with 7 features: oil temperature (target) and 6 electrical load features (HUFL, HULL, MUFL, MULL, LUFL, LULL). For the experiments, oil temperature and two high-load features (HUFL, HULL) are selected to maintain computational efficiency while preserving the multivariate forecasting challenge. A standard 70/15/15 train/validation/test split is employed, and all models are configured to predict 12 future time steps given 48 historical observations. This setting reflects practical forecasting scenarios where medium-term predictions guide operational decisions.

For the time series experiments, UPN operates in pure prediction mode without measurement updates during the forecast horizon. The dynamics and process noise networks learn temporal dependencies directly from the data without requiring domain-specific modifications.

Unlike the CubeSat experiments where the Markovian state is explicitly constructed from physical variables (position and velocity), time-series forecasting requires a learned latent representation. The observed features, temperature and electrical loads, do not directly correspond to canonical state variables with clear dynamical relationships. Instead, UPN operates under a latent Markovian assumption: there exists a low-dimensional latent state $z \in \mathbb{R}^4$ where the system dynamics are Markovian, such that future evolution depends only on the current latent state and not on the full observation history.

An encoder network maps the 48-step historical observation sequence to initial distributional parameters. The encoder, implemented as a 2-layer GRU with 16 hidden units, learns to compress 144-dimensional historical information (48 timesteps × 3 features) into a 4-dimensional latent state that serves as a sufficient statistic for future prediction. This latent state captures essential temporal patterns, trends, periodicities, and dependencies, without requiring explicit specification of what these patterns represent physically. The UPN dynamics then evolve this latent state forward as Equations (2) and (3). A decoder network then maps the latent trajectory back to observation space to produce predictions with uncertainty estimates.

This architecture is conceptually similar to Latent ODEs and neural state space models, but differs in maintaining explicit distributional parameters $(\mu, \Sigma)$ throughout the latent evolution rather than requiring sampling.

UPN is compared against Neural SDE [8,9]. While Neural SDEs model individual stochastic trajectories requiring Monte Carlo sampling for uncertainty estimation, UPN directly evolves distribution parameters through deterministic coupled ODEs. Both architectures used comparable network capacities with 4-dimensional latent states and 16 hidden units to ensure fair comparison. Figure 32 illustrates the fundamental architectural differences between the two approaches. UPN computes uncertainty analytically through a single forward pass of coupled ODEs, while Neural

SDE requires multiple stochastic trajectory samples followed by Monte Carlo aggregation.

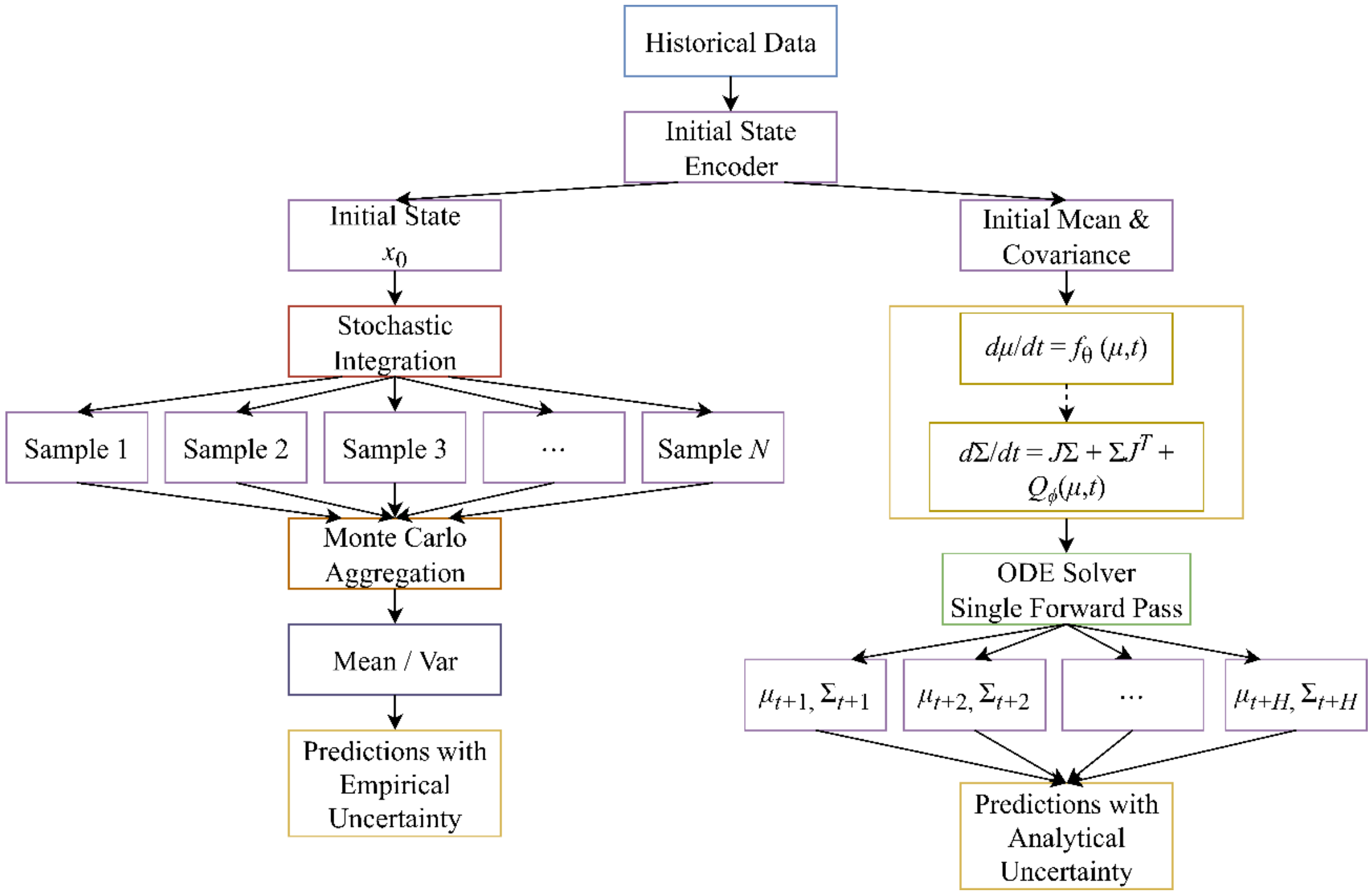

Figure 32. Architectural comparison between UPN's analytical uncertainty propagation through coupled ODEs (left) and Neural SDE's sampling-based approach requiring Monte Carlo aggregation (right) for H-step ahead time-series forecasting.

### *5.4.2. Results and Analysis*

Table 6 presents the comprehensive performance comparison between UPN and Neural SDE across multiple metrics. Figure 33 illustrates the training dynamics of both models over approximately 9,120 iterations. UPN demonstrates stable convergence with smoothly decreasing training and validation losses, reaching its optimal performance around iteration 2,000. In contrast, Neural SDE exhibits higher initial instability, with training loss spiking to nearly 12 before stabilizing. Both models converge relatively quickly, with minimal improvement after 2,000 iterations. The training efficiency comparison highlights UPN's computational advantage: each iteration requires only 1.056 seconds compared to 2.38 seconds for Neural SDE, reflecting the efficiency gains from analytical uncertainty propagation versus stochastic sampling.

Figure 34 presents side-by-side prediction visualizations across the three features. For the temperature feature (left column), both models track the upward trend accurately, though UPN provides tighter uncertainty bounds while maintaining good coverage. Features 1 and 2 (middle and right columns) exhibit more volatile behavior where the differences become pronounced. UPN produces smoother mean predictions with confidence intervals that appropriately widen during rapid transitions, particularly visible in Feature 2 around time step 55. Neural SDE's wider uncertainty bands across all features reflect the additional variance introduced by stochastic integration, which, while providing marginally better calibration (91.8% vs 90.3% coverage), comes at the cost of less informative predictions.

Table 6. Performance comparison between UPN and Neural SDE on ETTh1 dataset.

| Metric | UPN | Neural SDE |
|---|---|---|
| MSE | 0.589 | 0.669 |
| NLL | 2.541 | 2.542 |
| 95% Coverage | 0.903 | 0.918 |
| 90% Coverage | 0.855 | 0.874 |
| 80% Coverage | 0.770 | 0.784 |
| Training Time/epoch (s) | 401.2 | 903.9 |
| Inference Time/batch (s) | 0.145 | 0.958 |

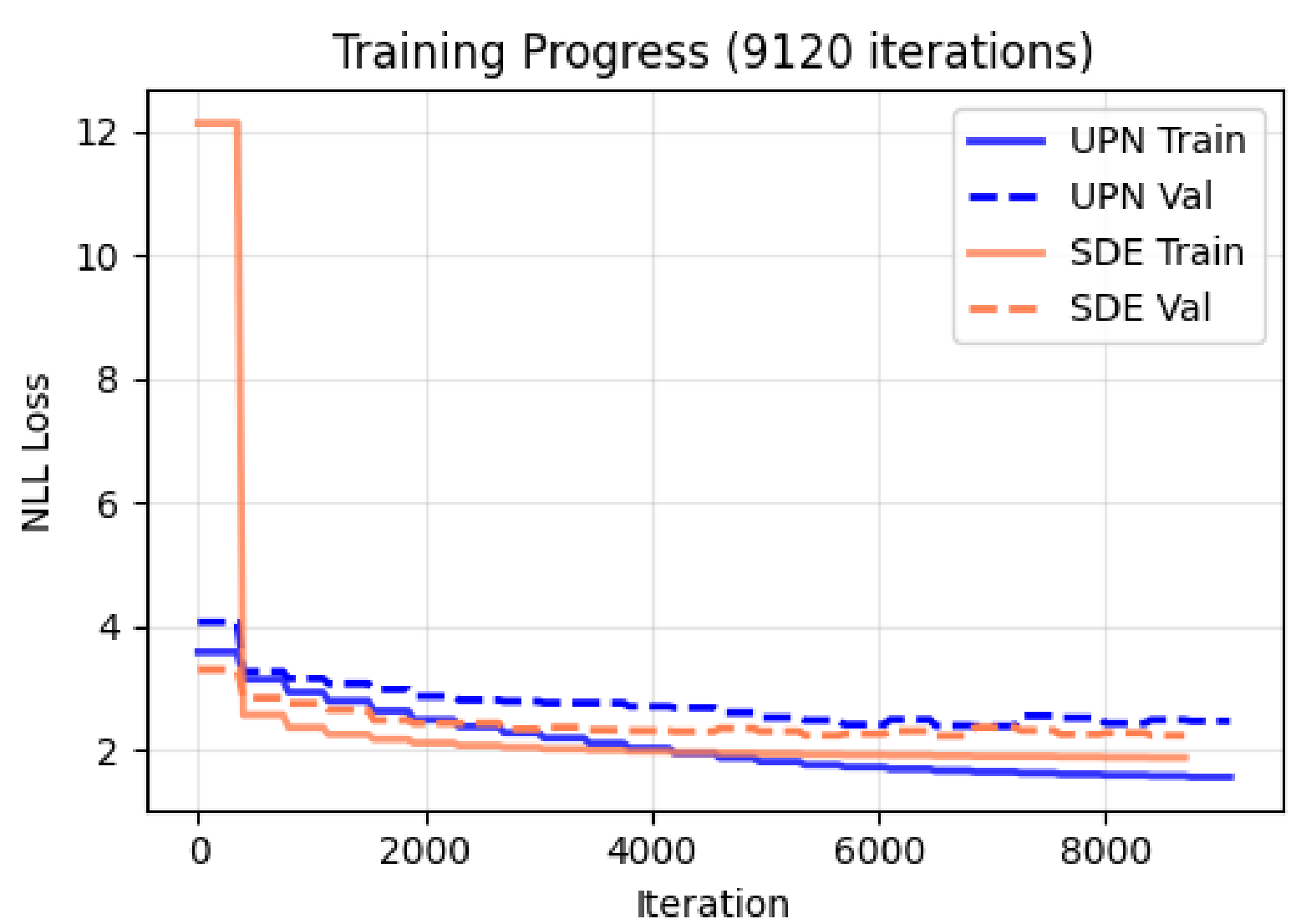

Figure 33. Training dynamics over 9,120 iterations showing (a) loss curves for both models, and (b) zoomed view of final 2,000 iterations.

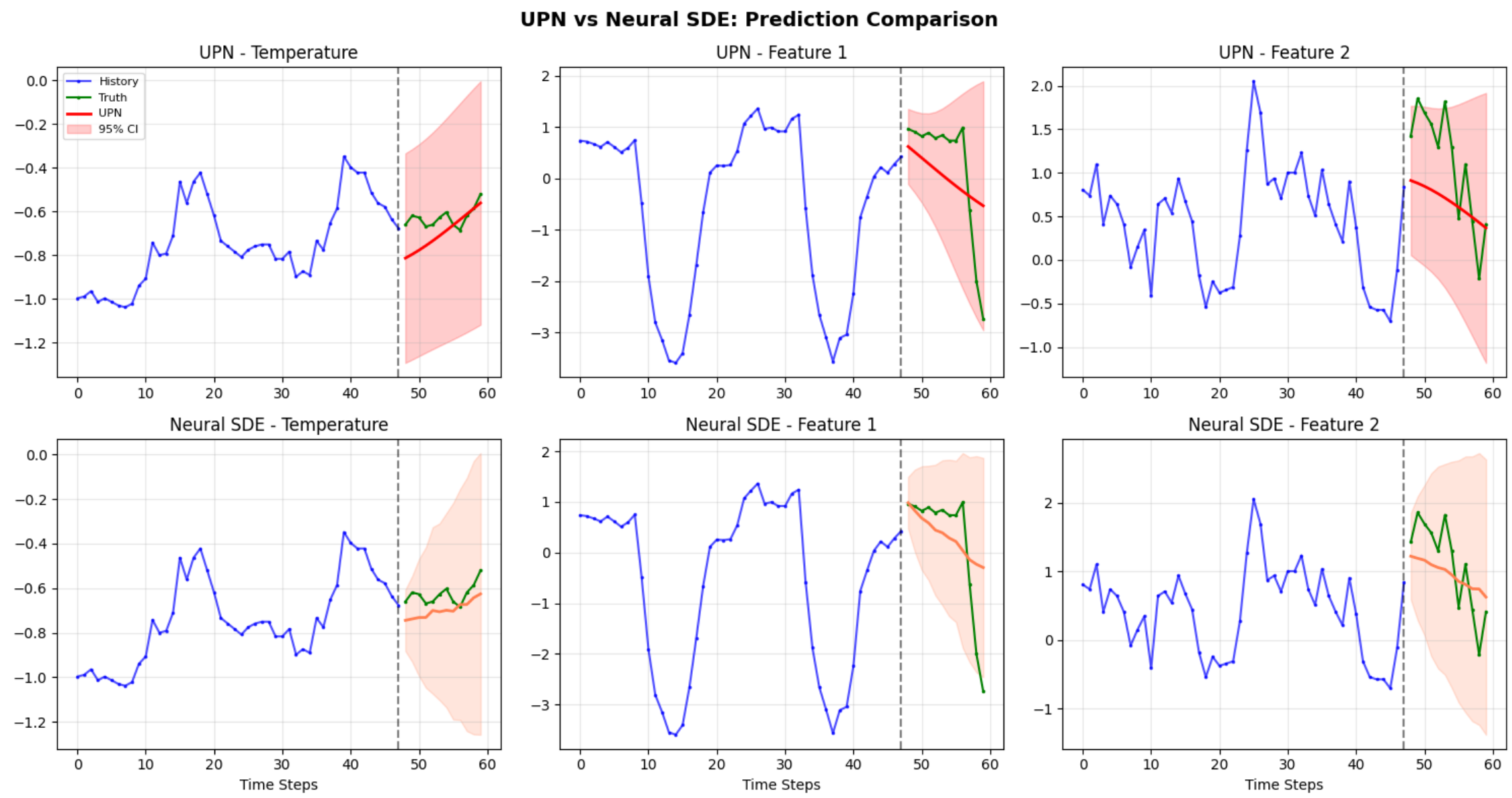

Figure 34. Twelve-step ahead predictions on ETTh1 test data comparing UPN (top row) and Neural SDE (bottom row) across temperature and two load features.

### *5.4.3. Sample Size Trade-off Analysis*

A critical consideration for Neural SDE is the number of Monte Carlo samples required for reliable uncertainty estimation. Figure 35 and Table 7 present our systematic analysis of this trade-off, evaluating performance across sample sizes from 10 to 100. Figure 35(a) reveals that Neural SDE's accuracy plateaus quickly, with diminishing returns beyond 25 samples. Even with 100 samples, it fails to match UPN's MSE of 0.589, remaining 14% worse. Figure 35(b) demonstrates that calibration improves with sample size but at severe computational cost, achieving marginally better coverage than UPN requires 4.5× more time relative to the 10-sample baseline. The linear scaling of computational cost with sample size (Figure 35c) confirms that no sweet spot exists where Neural SDE simultaneously matches UPN's accuracy and efficiency.

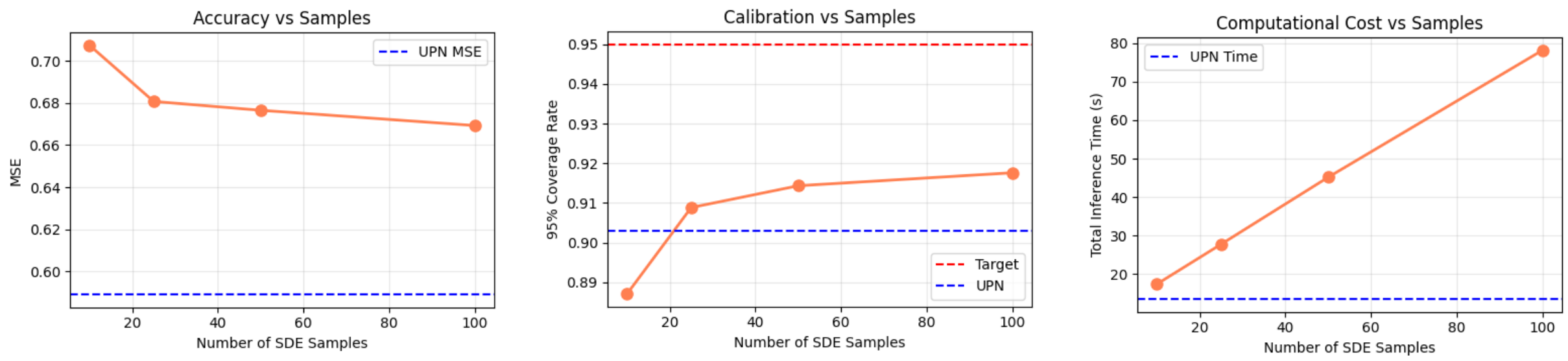

(a) (b) (c)

Figure 35. Neural SDE sample size trade-off analysis showing (a) MSE plateauing at 0.669 despite increasing samples while UPN baseline achieves 0.589, (b) calibration improving from 88.7% to 91.8% coverage as samples increase, and (c) linear computational cost scaling making Neural SDE up to 4.5× slower than UPN.

Table 7. Neural SDE performance metrics across different sample sizes.[§]

| Samples | MSE | 95% Coverage | Relative Time |
|---|---|---|---|
| 10 | 0.707 | 0.887 | 1.0 × |
| 25 | 0.681 | 0.909 | 1.6 × |
| 50 | 0.677 | 0.914 | 2.6 × |
| 100 | 0.669 | 0.918 | 4.5 × |
| UPN | 0.589 | 0.903 | - |

### *5.4.4. Computational Efficiency*

The computational advantages of UPN extend beyond raw speed metrics. With inference time of 0.145 seconds per batch versus 0.958 seconds for Neural SDE (100 samples), UPN achieves a 6.6× speedup while maintaining superior accuracy. This efficiency stems from UPN's fundamental design: propagating uncertainty analytically through coupled ODEs eliminates the need for repeated stochastic integration. For real-time applications such as grid monitoring or predictive maintenance, this order-of-magnitude improvement in inference speed can determine system feasibility.

[§] Relative Time shows computational cost relative to 10-sample Neural SDE baseline.

### 5.4.5. Discussion

The experimental results on ETTh1 demonstrate that UPN's theoretical advantages translate to practical benefits in real-world time series forecasting. The 14% improvement in MSE coupled with 6.6× faster inference validates our hypothesis that analytical uncertainty propagation better suits deterministic systems with observation noise, a characteristic of most industrial sensor data. While Neural SDE achieves slightly better uncertainty calibration, this marginal gain does not justify its computational overhead and reduced accuracy.

The strong performance of UPN on ETTh1 provides empirical validation of the latent Markovian assumption. The encoder successfully compresses 48 historical observations (144-dimensional: 48 steps × 3 features) into a 4-dimensional latent state that captures sufficient information for accurate 12-step-ahead prediction. The low MSE (0.589) and well-calibrated uncertainty (90.3% coverage) indicate that the learned latent representation $z \in \mathbb{R}^4$ indeed satisfies the Markov property for this system, future evolution depends primarily on the current latent state rather than requiring the full observation history.

The sample trade-off analysis particularly illuminates the fundamental limitations of sampling-based approaches. Neural SDE's inability to match UPN's accuracy even with extensive sampling suggests that the stochastic framework introduces irreducible noise when modeling essentially deterministic dynamics. This finding has important implications for practitioners choosing between uncertainty quantification methods: for systems where the underlying dynamics are deterministic but observed with noise, UPN's analytical approach provides both superior performance and computational efficiency.

## 5.5. Discussion

The experimental findings demonstrate that UPN achieves fundamentally distinct uncertainty quantification behavior compared to existing approaches across diverse applications. This section synthesizes key insights regarding calibration-accuracy trade-offs, computational considerations, theoretical assumptions, and practical deployment scenarios.

### 5.5.1. Calibration Quality and the Accuracy-Calibration Trade-off

Across non-chaotic dynamical systems, UPN consistently produces near-perfect 95%

confidence interval coverage (96.7% average) with minimal calibration error (1.7%), in stark contrast to ensemble-based methods that yield only 26.0% average coverage despite achieving 16% lower MSE. This apparent paradox, higher MSE coupled with superior calibration, reveals a fundamental insight about uncertainty quantification in learned dynamical systems.

The performance gap stems from architectural differences in uncertainty modeling. Ensemble methods capture only epistemic uncertainty through initialization variability, which collapses as models converge to similar solutions with sufficient training data. UPN explicitly models both aleatoric uncertainty (observation noise via $\Sigma_0 = \sigma_{obs}^2 I$) and process stochasticity (via learned $Q_\phi(\mu, t)$), propagating uncertainty analytically through the coupled ODE. This principled approach ensures uncertainty grows appropriately with prediction horizon, reflecting fundamental limits on predictability rather than merely model uncertainty.

The 16% MSE trade-off observed in non-chaotic systems is both expected and acceptable: UPN optimizes the full predictive distribution (negative log-likelihood) rather than point estimates alone, naturally regularizing against overfitting to training data. However, this interpretation requires nuance when considering chaotic systems. In the Lorenz experiments, UPN achieves 4.1× higher aggregate MSE compared to the 30-model ensemble, yet maintains 94.5% coverage while the ensemble achieves only 66.8%. The horizon-stratified analysis illuminates this trade-off: in the short-term predictable regime ($t < 0.3s$), UPN remains competitive with MSE ratio of 1.88×, but in the long-term chaotic regime ($t > 0.7s$), the gap widens to 8.3×.

Critically, the Lyapunov exponent analysis reveals that ensemble methods severely underestimate exponential error growth in chaotic dynamics, with estimated $\hat{\lambda} = 0.31$ compared to the theoretical $\lambda = 0.90$ (66% underestimation). UPN's estimated $\hat{\lambda} = 1.29$ represents 43% overestimation, a conservative bias that ensures confidence bounds remain reliable even when uncertainty grows faster than theoretically predicted due to model misspecification. This demonstrates that in chaotic regimes, trading point accuracy for calibrated uncertainty is not a compromise but the correct quantification of what is knowable versus fundamentally unpredictable.

For high-stakes applications, autonomous systems, safety-critical control, medical decision-making, well-calibrated uncertainty estimates are substantially more valuable than marginal improvements in point accuracy. Overconfident predictors lead to inappropriate actions under uncertainty; UPN's honest uncertainty quantification enables risk-aware decision-making where understanding prediction limits is as important as the prediction itself.

*5.5.2. Computational Efficiency and Scalability*

The computational complexity analysis establishes that UPN's per-step cost scales as $\mathcal{O}(n^3)$, dominated by Jacobian computation and covariance dynamics. Empirical benchmarks validate this theoretical complexity, with forward pass times scaling linearly with $n^3$ (fitted slope 0.0015 ms per unit $n^3$). However, several factors mitigate this apparent limitation:

1. Single-pass efficiency: Despite cubic per-step complexity, UPN provides analytical uncertainty quantification in a single forward pass. Neural SDE experiments demonstrate that achieving reasonable calibration requires 100 trajectory samples, resulting in 6.6× higher inference time compared to UPN while achieving 14% worse MSE (0.669 vs 0.589). For the Lorenz attractor, UPN's 186 ms per-batch inference is 14× faster than the 30-model ensemble (2632 ms) despite maintaining superior calibration.
2. GPU parallelization benefits: The normalized efficiency metric shows time per $n^3$ decreasing from 0.21 ms at $n = 2$ to 0.002 ms at $n = 16$, reflecting improved GPU utilization for larger problem sizes through better memory access patterns and increased arithmetic intensity. This suggests that the practical cost grows substantially slower than the theoretical cubic bound for moderate dimensions.
3. Practical tractability: Even at $n = 16$ with 20,555 parameters, forward passes complete in under 10 ms on an NVIDIA GeForce RTX 5060, suitable for real-time applications. The CubeSat experiments with 6-dimensional state space demonstrate that UPN achieves 89.6% MSE reduction with measurement updates while maintaining real-time performance.

For higher-dimensional systems, several complexity reduction strategies exist:

1. Low-rank covariance approximation [40]: Representing $\Sigma \approx LL^T$ where $L \in \mathbb{R}^{n \times r}$ with $r \ll n$ reduces storage to $\mathcal{O}(nr)$ and operations to $\mathcal{O}(n^2 r)$.
2. Structured covariance constraints [41]: Diagonal or block-diagonal forms reduce cost to $\mathcal{O}(n^2)$ when domain knowledge suggests sparse correlations.
3. Sparse Jacobian exploitation [42]: For locally-connected architectures, complexity reduces to $\mathcal{O}(\mathrm{nnz} \times n)$ where nnz is the number of non-zero Jacobian elements.
4. Mixed-precision computation [43]: Using FP16 for matrix operations with FP32

accumulation yields 2-6× speedup on modern GPUs with minimal accuracy loss.

*5.5.3. Gaussian Assumption and Limitations*

UPN's reliance on Gaussian distributional assumptions provides both strengths and limitations. The Gaussian moment closure approximation becomes exact when dynamics $f_\theta$ are linear in the state, and provides accurate local linearization for mildly nonlinear systems. The Jacobian-based covariance evolution captures how uncertainty propagates through linearized dynamics around the mean trajectory. However, this approximation constrains the model's ability to represent:

- Multi-modal distributions: Systems with bifurcations or multiple stable equilibria may exhibit non-Gaussian posterior distributions that cannot be captured by a single mean-covariance pair.
- Heavy-tailed uncertainty: Extreme events or outliers that deviate from Gaussian tails are underestimated in the current formulation.
- Non-differentiable dynamics: Systems with discontinuities (e.g., contact mechanics, switching systems) require special handling as Jacobians are undefined.

Despite these limitations, the experimental results across diverse systems, including the highly nonlinear Van der Pol oscillator, chaotic Lorenz attractor, and real-world CubeSat dynamics, demonstrate that Gaussian approximation provides excellent practical performance. The near-perfect calibration across these systems suggests that for many real-world applications, the Gaussian assumption is sufficiently accurate while enabling computational tractability.

Future extensions could address these limitations through:

- Normalizing flow extensions: Parameterizing non-Gaussian distributions by transforming Gaussian base distributions through learned invertible mappings.
- Higher-order moment methods: Tracking skewness and kurtosis to capture asymmetric or heavy-tailed distributions.
- Mixture of Gaussians: Representing multi-modal distributions as weighted sums of multiple UPN components.

*5.5.4. Markovian Initialization Strategies*

The success of single-point Markovian initialization across both non-chaotic and chaotic systems validates that for fully observed dynamical systems, current state contains sufficient information for future prediction without requiring observation histories. This has important practical implications:

1. For non-chaotic systems (oscillators, linear systems, damped pendulum), single-point initialization with $\Sigma_0 = \sigma_{obs}^2 I$ achieves 96.7% average coverage, demonstrating that observation noise alone, when properly propagated through learned dynamics, suffices for calibrated uncertainty quantification.
2. For chaotic systems (Lorenz attractor), incorporating modest phase ambiguity via $\Sigma_0 = \text{diag}(\sigma_{obs}^2 + \alpha \cdot \text{Var}[x_{\text{attractor}}])$ with $\alpha = 0.01$ maintains 94.5% coverage while correctly capturing exponential uncertainty growth. This approach balances observation noise with attractor-scale uncertainty without requiring historical context.
3. For physical systems with second-order dynamics, constructing Markovian state requires both configuration (position/orientation) and generalized velocities. Velocity estimation from finite differences of the last two observations creates a complete state representation where future evolution depends only on $(x, y, \theta, v_x, v_y, \omega)$ without additional history.
4. For time series with latent dynamics, a learned latent Markovian representation proves necessary. The encoder network successfully compresses 48-step historical observations (144-dimensional) into a 4-dimensional latent state that serves as a sufficient statistic for 12-step-ahead prediction, achieving competitive MSE (0.589) and well-calibrated uncertainty (90.3% coverage).

These results demonstrate that appropriate state representation, whether explicit for physical systems or learned for complex time series, is crucial for UPN's performance. The framework's flexibility in handling both scenarios through architectural choices (direct state input vs. encoder network) enables broad applicability across problem domains.

#### *5.5.5. Measurement Updates and Filtering Capabilities*

The CubeSat experiments provide compelling evidence of UPN's architectural advantage in incorporating measurements during prediction. The integrated Kalman update mechanism achieves 89.6% MSE reduction (from 0.061 to 0.006) with sparse updates every 5 timesteps, while simultaneously improving calibration from 81.0% to 96.8%, near-perfect uncertainty quantification.

This dramatic improvement stems from principled uncertainty handling: the Kalman framework optimally weights observations based on relative uncertainties in prediction ($\Sigma^-$) and measurement ($R$), propagating corrections through both state estimates and covariance. The innovation term $(y - H\mu^-)$ corrects the predicted mean, while the Kalman gain $K = \Sigma^- H^T (H\Sigma^- H^T + R)^{-1}$ modulates this correction based on confidence levels.

Critically, UPN's uncertainty exhibits theoretically correct behavior during the predict-observe-correct cycle: contracting when measurements arrive (information gain) and expanding during pure prediction periods (information decay). This dynamic representation is essential for autonomous systems that must reason about confidence when making decisions between measurement opportunities.

In contrast, other methods lack integrated update mechanisms. Simple blending (adjusting predictions toward observations with fixed gains) modifies values without updating uncertainty, causing corrections to decay rapidly. More sophisticated approaches like Ensemble Kalman Filters [45] require significant architectural modifications and computational overhead, fundamentally changing the methods' characteristics. Latent ODE faces architectural incompatibility as its RNN encoder requires complete observation sequences, preventing mid-sequence updates.

The dual-mode capability, pure prediction and adaptive filtering, emerges naturally from UPN's distributional formulation. Models trained without updates can be deployed with updates when observations become available, with no architectural modification required. This flexibility is particularly valuable for spacecraft operations, industrial monitoring, and other applications where observation frequency varies dynamically.

*5.5.6. Implications for Deployment*

The comprehensive experimental validation suggests UPN is particularly valuable when:

1. Calibrated uncertainty is critical: Safety-critical control, medical decision-making, financial risk assessment where understanding prediction limits enables appropriate action selection.
2. Computational efficiency matters: Real-time systems, embedded platforms, edge computing where 6-14× speedup over sampling methods enables uncertainty-aware prediction in resource-constrained environments.
3. Measurements arrive intermittently: Spacecraft navigation, sensor networks, remote

monitoring where integrated filtering adapts to irregular observation schedules.

4. Chaotic or complex dynamics: Weather prediction, fluid dynamics, epidemic modeling where understanding predictability horizons is as important as point forecasts.

Conversely, UPN may be less suitable when:

- Multi-modal distributions are essential (e.g., planning over multiple distinct trajectories)
- High dimensionality ($n > 100$) makes $O(n^3)$ cost prohibitive without approximations
- Point accuracy is paramount and uncertainty can be sacrificed (e.g., trajectory planning with generous safety margins)

*5.5.7. Broader Context and Future Directions*

This work demonstrates that analytical distribution evolution provides a compelling alternative to sampling-based uncertainty quantification in continuous-time systems. The key insight is that for many dynamical systems, uncertainty propagation can be efficiently approximated through Gaussian moment dynamics, avoiding the computational overhead of Monte Carlo methods while achieving superior calibration. Several promising research directions emerge:

1. **Theoretical extensions:**
   - Non-Gaussian moment closures for multi-modal or heavy-tailed distributions
   - Connections to information geometry and natural gradient methods
   - Formal guarantees on approximation quality under different dynamical regimes
2. **Architectural innovations:**
   - Multi-scale UPN for systems with heterogeneous temporal dynamics
   - Attention mechanisms for selective uncertainty propagation in high dimensions
   - Hybrid approaches combining UPN with physics-informed constraints
3. **Application domains:**
   - Climate modeling with quantified predictability horizons
   - Molecular dynamics with adaptive sampling guided by uncertainty
   - Causal discovery in time series through uncertainty-aware structure learning
   - Quantum machine learning via conceptual links between uncertainty evolution and quantum dynamics

The framework's continuous-time formulation and principled uncertainty handling position it as a foundational tool for the next generation of uncertainty-aware dynamical systems models.

## 6. Conclusion

This study introduced UPN, a general-purpose framework for continuous-time modeling with analytically-derived uncertainty quantification. By evolving both state means and covariances through coupled differential equations grounded in Gaussian moment closure approximation, UPN enables theoretically principled and computationally efficient uncertainty propagation in a single forward pass. Experimental evaluations across diverse domains validate this approach: near-perfect calibration (96.7% coverage on non-chaotic systems, 94.5% on chaotic Lorenz attractor) with single-point Markovian initialization; 89.6% error reduction on real-world CubeSat trajectory data through integrated Kalman filtering; and 14% improved accuracy and 6.6× faster inference on industrial ETTh1 time-series forecasting compared to Neural SDEs. These gains stem from UPN's core principle: propagating distributional parameters through coupled ODEs captures both aleatoric and epistemic uncertainty while maintaining computational efficiency 6-14× better than sampling alternatives. The broader significance of UPN extends beyond numerical performance. By producing reliable confidence estimates in a single forward pass, the framework enables deployment in safety-critical applications, autonomous navigation, medical decision support, climate prediction, where understanding uncertainty bounds is as essential as predictions themselves. The continuous-time formulation naturally accommodates irregular observations and sparse measurements, addressing practical challenges in real-world sensor systems. The chaotic system results illuminate a critical insight: conservative uncertainty estimates (Lyapunov exponent 43% overestimated) prove more valuable than ensemble methods' dangerous overconfidence (66% underestimated) when precise prediction becomes fundamentally impossible, enabling risk-aware reasoning about predictability horizons. Future work could extend the framework through multi-scale architectures for heterogeneous temporal dynamics, and sparse Jacobian exploitation for higher-dimensional systems. The continuous-time formulation aligns naturally with causal discovery from time series and quantum machine learning applications. UPN thus represents a principled shift toward analytical uncertainty quantification in neural differential equations, demonstrating that distribution evolution through coupled ODEs provides superior calibration and computational efficiency compared to sampling-based alternatives, bridging theoretical rigor from Gaussian processes and Kalman filtering with the flexibility of learned neural dynamics.

**CRediT authorship contribution statement**

**Hadi Jahanshahi**: Conceptualization, Methodology, Software, Validation, Formal analysis, Investigation, Visualization, Writing – original draft, Writing – review & editing. **Zheng H. Zhu**: Conceptualization, Methodology, Investigation, Resources, Supervision, Project administration, Funding acquisition, Writing – review & editing.

**Code Availability**

Complete implementation, examples, and documentation are available at: https://github.com/HJahanshahi/upn. The repository includes all code necessary to reproduce the results presented in this paper, along with comprehensive documentation for extending the framework to new applications.

**Declaration of competing interest**

The authors declare that they have no known competing financial interests or personal relationships that could have appeared to influence the work reported in this paper.

**Acknowledgments**

This work was supported by the Discovery Grant (RGPIN-2024-06290) and Collaborative Research and Training Experience Program Grant (555425-2021) of the Natural Sciences and Engineering Research Council of Canada. The authors also acknowledge support from the Natural Sciences and Engineering Research Council of Canada (NSERC) through a Postgraduate Scholarship – Doctoral (PGS D).

Nous remercions le Conseil de recherches en sciences naturelles et en génie du Canada (CRSNG) de son soutien par l'entremise d'une bourse d'études supérieures du doctorat (BES D).

**Data Availability**

CubeSat experimental data is available upon reasonable request to the corresponding author.

The ETTh1 time-series benchmark is publicly available at https://github.com/zhouhaoyi/Informer2020 [47]. Synthetic dynamical systems data (Sections 5.1-5.2) can be reproduced using procedures described in the manuscript. All code is available at https://github.com/HJahanshahi/upn. No proprietary data were used.

## References

[1] R.T. Chen, Y. Rubanova, J. Bettencourt, D.K. Duvenaud, Neural ordinary differential equations, Advances in Neural Information Processing Systems 31 (2018).

[2] A.M. Durán-Rosal, T. Ashley, J. Pérez-Rodríguez, F. Fernández-Navarro, Global and Diverse Ensemble model for regression, Neurocomputing (2025) 130520.

[3] B. Lakshminarayanan, A. Pritzel, C. Blundell, Simple and scalable predictive uncertainty estimation using deep ensembles, Advances in Neural Information Processing Systems 30 (2017).

[4] C. Blundell, J. Cornebise, K. Kavukcuoglu, D. Wierstra, Weight uncertainty in neural network, in: PMLR, 2015: pp. 1613–1622.

[5] R.M. Neal, Bayesian learning for neural networks, Springer Science & Business Media, 2012.

[6] A. Graves, Practical variational inference for neural networks, Advances in Neural Information Processing Systems 24 (2011).

[7] Y. Gal, Z. Ghahramani, Dropout as a bayesian approximation: Representing model uncertainty in deep learning, in: PMLR, 2016: pp. 1050–1059.

[8] X. Li, T.-K.L. Wong, R.T. Chen, D. Duvenaud, Scalable gradients for stochastic differential equations, in: PMLR, 2020: pp. 3870–3882.

[9] X. Liu, T. Xiao, S. Si, Q. Cao, S. Kumar, C.-J. Hsieh, Neural sde: Stabilizing neural ode networks with stochastic noise, arXiv Preprint arXiv:1906.02355 (2019).

[10] S. Särkkä, L. Svensson, Bayesian filtering and smoothing, Cambridge university press, 2023.

[11] R.E. Kalman, A new approach to linear filtering and prediction problems, (1960).

[12] S.F. Schmidt, Application of state-space methods to navigation problems, in: Advances in Control Systems, Elsevier, 1966: pp. 293–340.

[13] H. Zhang, C. Wen, Design of an incremental Extended Kalman Filter group for a class of multiplicatively decomposed strongly nonlinear systems considering hidden state variables, Neurocomputing 637 (2025) 130057.

[14] S.J. Julier, J.K. Uhlmann, New extension of the Kalman filter to nonlinear systems, in: Spie, 1997: pp. 182–193.

[15] M. Raissi, P. Perdikaris, G.E. Karniadakis, Physics-informed neural networks: A deep learning framework for solving forward and inverse problems involving nonlinear partial differential equations, Journal of Computational Physics 378 (2019) 686–707.

[16] J. Tang, Y. Tong, L. Chen, S. Cai, S. Xiong, Integrating neural networks with numerical schemes for dynamical systems: A review, Neurocomputing (2025) 130122.

[17] S. Greydanus, M. Dzamba, J. Yosinski, Hamiltonian neural networks, Advances in Neural Information Processing Systems 32 (2019).

[18] P. Niu, J. Guo, Y. Chen, Y. Zhou, M. Feng, Y. Shi, Improved physics-informed neural network in mitigating gradient-related failures, Neurocomputing 638 (2025) 130167. https://doi.org/10.1016/j.neucom.2025.130167.

[19] E. Dupont, A. Doucet, Y.W. Teh, Augmented neural odes, Advances in Neural Information Processing Systems 32 (2019).
[20] P. Kidger, J. Morrill, J. Foster, T. Lyons, Neural controlled differential equations for irregular time series, Advances in Neural Information Processing Systems 33 (2020) 6696–6707.
[21] A. Norcliffe, C. Bodnar, B. Day, N. Simidjievski, P. Liò, On second order behaviour in augmented neural odes, Advances in Neural Information Processing Systems 33 (2020) 5911–5921.
[22] K. He, X. Zhang, S. Ren, J. Sun, Deep residual learning for image recognition, in: 2016: pp. 770–778.
[23] E. Weinan, A proposal on machine learning via dynamical systems, Communications in Mathematics and Statistics 5 (2017) 1–11.
[24] Y. Lu, A. Zhong, Q. Li, B. Dong, Beyond finite layer neural networks: Bridging deep architectures and numerical differential equations, in: PMLR, 2018: pp. 3276–3285.
[25] C. Zhou, Q. Zhang, J. Cheng, Neural adaptive delay differential equations, Neurocomputing (2025) 130634.
[26] Y. Rubanova, R.T. Chen, D.K. Duvenaud, Latent ordinary differential equations for irregularly-sampled time series, Advances in Neural Information Processing Systems 32 (2019).
[27] W. Grathwohl, R.T. Chen, J. Bettencourt, I. Sutskever, D. Duvenaud, Ffjord: Free-form continuous dynamics for scalable reversible generative models, arXiv Preprint arXiv:1810.01367 (2018).
[28] Y. Oh, D.-Y. Lim, S. Kim, Stable neural stochastic differential equations in analyzing irregular time series data, arXiv Preprint arXiv:2402.14989 (2024).
[29] L. Yang, T. Gao, Y. Lu, J. Duan, T. Liu, Neural network stochastic differential equation models with applications to financial data forecasting, Applied Mathematical Modelling 115 (2023) 279–299.
[30] X. Liu, S. Cui, W. Qiao, J. Liu, G. Wu, Remaining useful life prediction with uncertainty quantification for rotating machinery: A method based on explainable variational deep gaussian process, Neurocomputing 638 (2025) 130232.
[31] A. Damianou, N.D. Lawrence, Deep gaussian processes, in: PMLR, 2013: pp. 207–215.
[32] M. Garnelo, D. Rosenbaum, C. Maddison, T. Ramalho, D. Saxton, M. Shanahan, Y.W. Teh, D. Rezende, S.A. Eslami, Conditional neural processes, in: PMLR, 2018: pp. 1704–1713.
[33] M.-H. Fan, J.-H. Zhao, L. Ding, X.-Y. Ma, R.-L. Fu, Predicting nonlinear dynamic systems by causal physics-informed neural networks with ResNet blocks, Neurocomputing 656 (2025) 131589. https://doi.org/10.1016/j.neucom.2025.131589.
[34] S. Zhang, P. Yuan, M. Pan, C. Zhang, Stagewise edge expansion physics-informed neural networks for solving PDEs, Neurocomputing 654 (2025) 131318. https://doi.org/10.1016/j.neucom.2025.131318.
[35] N. Shlezinger, G. Revach, A. Ghosh, S. Chatterjee, S. Tang, T. Imbiriba, J. Dunik, O. Straka, P. Closas, Y.C. Eldar, AI-aided Kalman filters, arXiv Preprint arXiv:2410.12289 (2024).
[36] P. Hao, O. Karakuş, A. Achim, RKFNet: A novel neural network aided robust Kalman filter, Signal Processing 230 (2025) 109856.
[37] S. Chen, Y. Zheng, D. Lin, P. Cai, Y. Xiao, S. Wang, MAML-KalmanNet: A neural network-assisted Kalman filter based on model-agnostic meta-learning, IEEE Transactions on Signal Processing (2025).
[38] C. Chen, N. Chen, Y. Zhang, J.-L. Wu, CGKN: A deep learning framework for modeling

complex dynamical systems and efficient data assimilation, Journal of Computational Physics 532 (2025) 113950. https://doi.org/10.1016/j.jcp.2025.113950.
[39] C. Chen, Z. Wang, N. Chen, J.-L. Wu, Modeling partially observed nonlinear dynamical systems and efficient data assimilation via discrete-time conditional Gaussian Koopman network, Computer Methods in Applied Mechanics and Engineering 445 (2025) 118189. https://doi.org/10.1016/j.cma.2025.118189.
[40] S. Ambikasaran, D. Foreman-Mackey, L. Greengard, D.W. Hogg, M. O'Neil, Fast Direct Methods for Gaussian Processes, IEEE Transactions on Pattern Analysis and Machine Intelligence 38 (2016) 252–265. https://doi.org/10.1109/TPAMI.2015.2448083.
[41] H. Nickisch, A. Solin, A. Grigorevskiy, State Space Gaussian Processes with Non-Gaussian Likelihood, in: J. Dy, A. Krause (Eds.), Proceedings of the 35th International Conference on Machine Learning, PMLR, 2018: pp. 3789–3798. https://proceedings.mlr.press/v80/nickisch18a.html.
[42] A. Griewank, A. Walther, Evaluating derivatives: principles and techniques of algorithmic differentiation, SIAM, 2008.
[43] P. Micikevicius, S. Narang, J. Alben, G. Diamos, E. Elsen, D. Garcia, B. Ginsburg, M. Houston, O. Kuchaiev, G. Venkatesh, Mixed precision training, arXiv Preprint arXiv:1710.03740 (2017).
[44] D.P. Kingma, J. Ba, Adam: A method for stochastic optimization, arXiv Preprint arXiv:1412.6980 (2014).
[45] G. Evensen, The Ensemble Kalman Filter: theoretical formulation and practical implementation, Ocean Dynamics 53 (2003) 343–367. https://doi.org/10.1007/s10236-003-0036-9.
[46] M.S. Arulampalam, S. Maskell, N. Gordon, T. Clapp, A tutorial on particle filters for online nonlinear/non-Gaussian Bayesian tracking, IEEE Transactions on Signal Processing 50 (2002) 174–188. https://doi.org/10.1109/78.978374.
[47] H. Zhou, S. Zhang, J. Peng, S. Zhang, J. Li, H. Xiong, W. Zhang, Informer: Beyond Efficient Transformer for Long Sequence Time-Series Forecasting, AAAI 35 (2021) 11106–11115. https://doi.org/10.1609/aaai.v35i12.17325.